\documentclass[final,12pt]{clear2023} 

\usepackage{enumitem}       
\usepackage{tikz}
\usepackage{booktabs}
\usepackage{xcolor}
\usepackage{pifont}
\usepackage{caption}
\usepackage{titlesec}
\usepackage{array}

\usepackage[capitalize]{cleveref}
\crefname{section}{Sec.}{Secs.}
\Crefname{section}{Section}{Sections}
\Crefname{table}{Table}{Tables}
\crefname{table}{Tab.}{Tabs.}

\crefformat{section}{\S#2#1#3} 
\crefformat{subsection}{\S#2#1#3}
\crefformat{subsubsection}{\S#2#1#3}

\newcommand{\causaltriplet}{{\tt Causal} {\tt Triplet}}

\newcommand{\cmark}{\ding{51}}
\newcommand{\xmark}{\ding{55}}

\newcommand{\update}[1]{#1}

\usetikzlibrary{bayesnet}

\usepackage{notation}

\title[Causal Triplet]{Causal Triplet: An Open Challenge for\\Intervention-centric Causal Representation Learning}
\usepackage{times}

\clearauthor{%
 \Name{Yuejiang Liu}\textsuperscript{$1,2,$}{\footnote{Most work done during an internship at Amazon.}} \Email{yuejiang.liu@epfl.ch}\\
 \Name{Alexandre Alahi}\textsuperscript{$2$} \Email{alexandre.alahi@epfl.ch}\\
 \Name{Chris Russell}\textsuperscript{$1$} \Email{cmruss@amazon.de}\\
 \Name{Max Horn}\textsuperscript{$1$} \Email{hornmax@amazon.de}\\
 \Name{Dominik Zietlow}\textsuperscript{$1$} \Email{zietld@amazon.de}\\
 \Name{Bernhard Schölkopf}\textsuperscript{$1,3$} \Email{bs@tuebingen.mpg.de}\\
 \Name{Francesco Locatello}\textsuperscript{$1$} \Email{locatelf@amazon.de}\\
 \addr 1. Amazon, T\"ubingen, Germany\\
 \addr 2. École Polytechnique Fédérale de Lausanne (EPFL), Switzerland\\
 \addr 3. Max Planck Institute for Intelligent Systems, T\"ubingen, Germany
 \vspace{-10pt}
}

\begin{document}

\maketitle


\begin{abstract}
Recent years have seen a surge of interest in learning high-level causal representations from low-level image pairs under interventions.
Yet, existing efforts are largely limited to simple synthetic settings that are far away from real-world problems.
In this paper, we present {\tt Causal Triplet}, a causal representation learning benchmark featuring not only visually more complex scenes, but also two crucial desiderata commonly overlooked in previous works:
(i) an actionable counterfactual setting, where only certain object-level variables allow for counterfactual observations whereas others do not;
(ii) an interventional downstream task with an emphasis on out-of-distribution robustness from the independent causal mechanisms principle.
Through extensive experiments, we find that models built with the knowledge of disentangled or object-centric representations significantly outperform their distributed counterparts.
However, recent causal representation learning methods still struggle to identify such latent structures, indicating substantial challenges and opportunities for future work.
Our code and datasets will be available at \url{https://sites.google.com/view/causaltriplet}.

\end{abstract}

\vspace{2pt}




\section{Introduction}
\label{sec:intro}
Causal representation learning, which strives to discover and represent high-level causal variables from low-level sensory observations, is a critical component for combining modern representation learning with classical causal modeling~\citep{scholkopfCausalRepresentationLearning2021a}.
Yet, this is a highly challenging and even ill-posed problem in the unsupervised setting, due to the limits of i.i.d. observational data and ambiguities in levels of abstractions.
A promising remedy is to exploit independent causal mechanisms~\citep{scholkopfCausalAnticausalLearning2012,parascandoloLearningIndependentCausal2018a} and sparse mechanism shifts~\citep{scholkopfCausalRepresentationLearning2021a} between pairs of observations under interventions.
Significant progress has been made in both theoretical underpinnings and practical algorithms~\citep{locatelloWeaklySupervisedDisentanglementCompromises2020a,vonkugelgenSelfSupervisedLearningData2021,lachapelleDisentanglementMechanismSparsity2022,lippeCITRISCausalIdentifiability2022,brehmerWeaklySupervisedCausal2022}.
However, existing efforts are still confined to simple synthetic data sets that are far from practical problems.
In this work, we aim to bridge this gap by revisiting recent hypotheses and methods in more realistic settings. Beside much richer visual complexity of high-dimensional data with heavy occlusions, camera motion, ego motion, and visual scenes with multiple objects, we incorporate two key properties.

\textit{Desideratum 1:} We take the perspective of embodied agents acting in the world~\citep{cohenGroundedTheoryCausation2022,deitkeRetrospectivesEmbodiedAI2022}, blurring the line between interventions and counterfactuals. In fact, we notice that assuming that only one/few variables change as a result of a sparse shift that maintains the noise realization of all other variables is a counterfactual and not a proper intervention. Arguably, {\em perfect counterfactuals} are difficult to observe, and not all variables may allow observable counterfactuals, invalidating the identifiability results of existing methods~\citep{locatelloWeaklySupervisedDisentanglementCompromises2020a,vonkugelgenSelfSupervisedLearningData2021,klindtNonlinearDisentanglementNatural2021,brehmerWeaklySupervisedCausal2022}. Instead, we propose a practical alternative, where only specific variables corresponding to physical and manipulable objects allow for counterfactual observations while others do not (\textit{e.g.}, camera view, self-occlusions, global scene properties that can change over time). We call this ``\textit{actionable counterfactuals}'' -- a setting that can naturally emerge in various situations, such as embodied agents learning to discover causal models through active interventions and human-object interactions observed from a first-person camera view. 


\textit{Desideratum 2:} We argue that the notion of causal representation learning should also move beyond identifying underlying causal factors, and more explicitly account for object affordances that support downstream tasks involving interventions and reasoning.
We thus propose a new task modeling high-level actions between image pairs, with a particular emphasis on out-of-distribution robustness from the independent causal mechanisms principle (\textit{i.e.}, if $X\rightarrow Y$, then $P(Y|X)$ remains valid under interventions on $X$).
This task necessitates the discovery of not only causal variables that can be independently manipulated but also causal mechanisms behind actions that govern abstract transformations of object states.
Solving it can be an important step towards \textit{Intervention-Centric Causal Representation Learning}, where interventions are associated with (and represented alongside) objects, changing an object's default dynamics whenever they take place.

In light of these two desiderata, we introduce a new benchmark for causal representation learning, named \causaltriplet. 
It features (i) pairs of images with high visual complexity and variability, (ii) real-world actions inducing sparse changes in the underlying structures of natural scenes.
Through extensive experiments, we show that intervention models built with causally structured representations (\textit{e.g.}, disentangled and object-centric) can substantially outperform distributed representation counterparts.
However, recent approaches still have difficulty in correctly discovering the latent structures. In particular, we observe that learning independent causal mechanisms is highly challenging in the presence of spurious correlations between actions and object attributes, introducing shortcuts such as recognizing an object based on the action performed or vice-versa that do not generalize to unseen compositions. 
Overall, we hope the proposed benchmark will foster advancement of intervention-centric causal representation learning towards real-world contexts.

\section{Related Work}
\label{sec:related}
Early efforts towards causal representation learning revolve around the notion of {\em independence}.
One line of works attempts to learn disentangled representations~\citep{chenInfoGANInterpretableRepresentation2016,higginsBetaVAELearningBasic2017,rolinekVariationalAutoencodersPursue2019,locatelloChallengingCommonAssumptions2019,goyalRecurrentIndependentMechanisms2021}, where all causal variables are assumed {\em statistically independent}.
Unfortunately, causal structures of real-world observations are often non-trivial, introducing strong correlations between latent variables and undermining the foundation of disentanglement~\citep{traubleDisentangledRepresentationsLearned2021}.
Another prominent branch lies in object-centric representation learning~\citep{greffNeuralExpectationMaximization2017,burgessMONetUnsupervisedScene2019,greffMultiObjectRepresentationLearning2019,locatelloObjectCentricLearningSlot2020a}, which seeks to decompose visual scenes into a set of individual objects that can be {\em independently manipulated}~\citep{yangLearningManipulateIndividual2020}, can {\em move independently}~\citep{atzmonCausalViewCompositional2020}, or {\em independently re-appear} across samples~\citep{yangDyStaBUnsupervisedObject2021}.
However, it remains unclear when and to what extent objects can be viewed as causal variables.

More recently, there has been a growing interest in learning causal representations from interventions~\citep{ahujaInterventionalCausalRepresentation2022}.
In particular, several recent works exploit the {\em sparsity} of mechanism shifts between paired observations~\citep{locatelloWeaklySupervisedDisentanglementCompromises2020a,zimmermannContrastiveLearningInverts2021,vonkugelgenSelfSupervisedLearningData2021,ahujaWeaklySupervisedRepresentation2022,brehmerWeaklySupervisedCausal2022} \update{or in a temporarily intervened sequence}~\citep{klindtNonlinearDisentanglementNatural2021,lippeICITRISCausalRepresentation2022,lippeInterventionDesignCausal2022,lippeCITRISCausalIdentifiability2022,lachapelleDisentanglementMechanismSparsity2022} to identify the underlying causal variables.
Nevertheless, these works are still restricted to simple toy datasets and impractical intervention conditions. 
In contrast, we propose a new benchmark with high visual complexity (\textit{e.g.}, camera view, self-occlusions, non-static global scene properties) and realistic interventions in the form of actionable counterfactuals from the embodied agent perspective.





\section{Benchmark Design}
\label{sec:benchmark}
In this section, we present \causaltriplet, a new benchmark for causal representation learning in visually complex settings.
We will first describe our designed task through the lens of causality, and subsequently, discuss the key properties of the collected datasets.
\subsection{Benchmark Task}
\label{subsec:task}

\begin{table}[t]
    \centering
    \small
    \resizebox{\textwidth}{!}{%
    \begin{tabular}{cccccc}
    \toprule
                            & Causal3DIdent & Causal Pinball & Interventional Pong & Causal World & Causal Triplet \\
                            & \citep{vonkugelgenSelfSupervisedLearningData2021} & \citep{lippeICITRISCausalRepresentation2022} & \citep{lippeCITRISCausalIdentifiability2022} &  \citep{ahmedCausalWorldRoboticManipulation2021} & (ours) \\ \midrule
    Scale Variability       & \xmark & \xmark & \xmark & \xmark & \cmark \\ 
    Shape Variability       & \xmark & \xmark & \xmark & \xmark & \cmark \\ 
    Illumination Variability & \xmark & \xmark & \xmark & \xmark & \cmark \\ 
    Camera Motion           & \xmark & \xmark & \xmark & \xmark & \cmark \\ 
    Complex Texture         & \xmark & \xmark & \xmark & \xmark & \cmark  \\ 
    Object Occlusion        & \xmark & \xmark & \xmark & \cmark &  \cmark \\ 
    Object Number           & one & few & few & few & many \\ 
    Downstream Task         & factor & factor & factor & action & action
    \\ \bottomrule
    \end{tabular}
    }
    \caption{Benchmark comparison for causal representation learning.}
    \vspace{-5pt}
    \label{tab:benchmark}
\end{table}

Modeling paired observations under known or unknown interventions has been a common setting for causal representation learning~\citep{locatelloWeaklySupervisedDisentanglementCompromises2020a,vonkugelgenSelfSupervisedLearningData2021,lachapelleDisentanglementMechanismSparsity2022,brehmerWeaklySupervisedCausal2022}.
Yet, modeling interventions themselves remains largely unexplored.
In fact, learning representations of high-level actions is deeply rooted in a variety of practical problems, from perceiving human actions~\citep{fathiUnderstandingEgocentricActivities2011} to building interventional world models~\citep{haWorldModels2018,leiVariationalCausalDynamics2022}.
To bridge this gap, we extend the paired setup to an intervention modeling task, with an emphasis on out-of-distribution robustness.

\begin{figure}[t]
    \vspace{-5pt}
    \centering
    \resizebox{0.85\linewidth}{!}{
    \definecolor{instantcolor}{HTML}{117733} %
\definecolor{edgecolor}{HTML}{444444}
\definecolor{actioncolor}{HTML}{DB5050}

\begin{tikzpicture}
    \centering

    \node (z_1_1) [latent, thick] {$z^1_1$};
    \node (z_1_2) [latent, right=0.2cm of z_1_1, thick] {$z^2_1$};
	\node (z_1_d) [const, right=0.2cm of z_1_2, thick] {$\cdots$} ; %
    \node (z_1_k) [latent, right=0.2cm of z_1_d, thick] {$z^K_1$};

    \plate[inner sep=0.3em,yshift=0.25em,dashed,color=instantcolor, thick] {o1}{(z_1_1) (z_1_2) (z_1_k)}{};
    \node[below right=-12.5pt and -55pt of o1, thick] {$o_1$};

	\node (o_d_1) [const, right=0.4cm of o1] {$\cdots$} ; %

    \node (z_2_1) [latent, right=0.4cm of o_d_1, thick] {$z^1_n$};
	\node (z_2_d1) [const, right=0.2cm of z_2_1, thick] {$\cdots$}; %
    \node (z_2_2) [latent, right=0.2cm of z_2_d1, thick] {$z_a$};
	\node (z_2_d2) [const, right=0.2cm of z_2_2, thick] {$\cdots$}; %
    \node (z_2_k) [latent, right=0.2cm of z_2_d2, thick] {$z^K_n$};

    \plate[inner sep=0.3em,yshift=0.25em,dashed,color=instantcolor, thick] {o2}{(z_2_1) (z_2_2) (z_2_k)}{};
    \node[below right=-12.5pt and -55pt of o2, thick] {$o_n$};
    

	\node (o_d_2) [const, right=0.4cm of o2] {$\cdots$} ; %

    \node (z_3_1) [latent, right=0.4cm of o_d_2, thick] {$z^1_N$};
    \node (z_3_2) [latent, right=0.2cm of z_3_1, thick] {$z^2_N$};
	\node (z_3_d) [const, right=0.2cm of z_3_2, thick] {$\cdots$}; %
    \node (z_3_k) [latent, right=0.2cm of z_3_d, thick] {$z^K_N$};
    
    \plate[inner sep=0.3em,yshift=0.25em,dashed,color=instantcolor, thick] {o3}{(z_3_1) (z_3_2) (z_3_k)}{};
    \node[below right=-12.5pt and -55pt of o3, thick] {$o_N$};

    \node (x) [obs, below=1.5cm of o2, xshift=-4em, thick] {$x$};

    \edge[color=edgecolor,thick]{o1}{x};
    \edge[color=edgecolor,thick]{o2}{x};
    \edge[color=edgecolor,thick]{o3}{x};
    
    \node (c) [latent, above=1.5cm of o1, xshift=3em, thick] {$c$};
    
    \edge[color=edgecolor,thick]{c}{o1};
    \edge[color=edgecolor,thick]{c}{o2};
    \edge[color=edgecolor,thick]{c}{o3};

    \node (a) [obs, above=of o3, xshift=4em, thick] {$a$};

    \edge[color=edgecolor,thick]{c}{a};

    \node (z_g) [latent, left=0.8cm of z_1_1, thick] {$z_s$};
    \edge[color=edgecolor,thick]{c}{z_g};

    \node (z_tilde) [latent, right=0.8cm of o3, thick, draw=actioncolor] {$\Tilde{z}_a$};
    \edge[color=edgecolor,thick]{a}{z_tilde};
    \path[->, color=edgecolor,thick] (z_2_2) edge[bend left=30] node[yshift=.5em] {} (z_tilde);


    \node (x_tilde) [obs, below=1.5cm of o3, xshift=-4.5em, thick] {$\Tilde{x}$};

    \edge[color=edgecolor,thick]{z_tilde}{x_tilde};
    \path[->, color=edgecolor,thick] (z_g) edge[bend right=27.5] node[yshift=.5em] {} (x_tilde);
    \edge[color=edgecolor,thick]{o1}{x_tilde};
    \edge[color=edgecolor,thick]{o2}{x_tilde};
    \edge[color=edgecolor,thick]{o3}{x_tilde};
    
    \path[->, color=edgecolor,thick] (z_g) edge[bend right=12.5] node[yshift=.5em] {} (x);

    \vspace{-4pt}
\end{tikzpicture}
    }
    \vspace{-12pt}
    \caption{
    Causal graph for a pair of scene observations $(x,\Tilde{x})$ before and after an action $a$.
    The data generating process of each raw observation is described by a set of latent factors, including global scene-level factors $z_s$ and local object-level factors $z_n^k$, which are statistically dependent due to unobserved confounders $c$.
    The action is assumed to sparsely influence only one (or a few) object-level factor $z_a$ in the scene.
    The other latent factors may stay constant in photo-realistic simulations but vary over time in real-world observations.
    }
    \label{fig:causal_graph}
\end{figure}
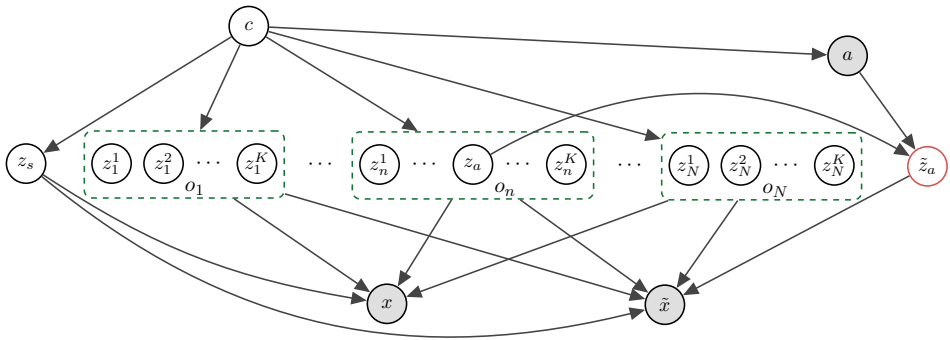

\paragraph{Problem setting.}
Consider the data generating process $\xb = g(\zb)$, where $\xb \in \RR^d$ is a high-dimensional image observation, $\zb \in \RR^l$ is a set of latent factors of variation, and $g: \RR^l \rightarrow \RR^d$ is the underlying generative mechanism.
We assume that the set of latent factors consists of both global scene-level variables $z_s$ and local object-level variables $z_n^k$, where $n$ is the object index and $k$ is the latent index within the object.
Similar to prior work~\citep{lippeInterventionDesignCausal2022,lippeCITRISCausalIdentifiability2022}, we consider an intervention process $\Tilde{\zb} = h(\zb, \ab)$, where action $\ab \in \set{A}$ is a categorical variable, and $h: \RR^l \times \set{A} \rightarrow \RR^l$ is a deterministic transition function.
We assume that each action affects one or a few object-level latent factors, resulting in paired observations $(\xb, \Tilde{\xb})$ before and after intervention.
Additionally, we assume that the latent factors and actions have dependences due to unobserved confounding $\cbb$.
For clarity, we refer to the latent factors affected by the action $\ab$ as $\zb_a$ in the rest of the paper.
\cref{fig:causal_graph} shows the causal graph encompassing our assumptions about the data generating process.

Unlike previous works aimed at identifying latent factors, we focus on downstream reasoning about the categories of high-level actions given image pairs, \textit{i.e.}, modeling the conditional probability $\PP(\ab \given \xb, \Tilde{\xb})$.
Crucially, we assume that the effect of an action $\PP(\Tilde{\zb}_{a} \given {\zb}_{a}, \ab)$ stays invariant, whereas the joint distribution of latent factors may change between training and test data, \textit{i.e.}, $\PP_\tr(\ab,\zb) \neq \PP_\te(\ab,\zb)$. 
Such distribution shifts commonly arise in practice, \textit{e.g.}, the training dataset contains only a subset of object classes or object-action combinations that a learning system encounters during deployment.
Inferring actions in this setting requires discoveries of both causal variables that can be independently manipulated and causal mechanisms behind actions that govern abstract transformations.

\paragraph{Paired observations as actionable counterfactuals.}
Paired observations are often assumed to share the same noise realization in prior works~\citep{locatelloWeaklySupervisedDisentanglementCompromises2020a,vonkugelgenSelfSupervisedLearningData2021, brehmerWeaklySupervisedCausal2022}.
We remark that this amounts to a perfect counterfactual, which is largely lacking in nature~\citep{hollandStatisticsCausalInference1986} and only in part plausible for embodied agents that learn to discover causal models through interactions~\citep{cohenGroundedTheoryCausation2022}, \textit{e.g.}, a robot intentionally keeps almost everything unchanged while manipulating an object in the environment.
Of course, this will allow only a subset of all possible counterfactuals, restricting to those that can be realized by the actions of the embodied agent.
We call these ``\textit{actionable counterfactuals}''.
Unfortunately, even this view would be too restrictive.
When it comes to real-world observations, many factors (\textit{e.g.}, camera view, object occlusions) may vary from time to time.
In our benchmark, we assume that the agent can only perform actionable counterfactuals, where most objects are not manipulated but the global properties of the visual scene may change.
This is a step towards \textit{Intervention-Centric Causal Representation Learning}, where interventions are associated with (and represented alongside) objects.

\paragraph{Sparse mechanism shifts between image representations.} 
Given the causal graph in \cref{fig:causal_graph}, the joint distribution of the latent factors can be factorized into a number of causal conditionals,
\begin{equation*}
    \PP(\zb, \cbb) = \PP(\cbb) \PP(\zb_s \given \cbb) \prod_{n=1}^N \PP(\zb_n^1, \zb_n^2, \cdots, \zb_n^K \given \cbb),
\end{equation*}
where only one/few conditionals are intervened between paired observations~\citep{scholkopfCausalRepresentationLearning2021a}.
Intuitively, an action does not manipulate all objects in a scene or all properties of the manipulated object at the same time.
Instead, it may sparsely affect a small subset of latent components in a causally structured scene representation \citep{locatelloWeaklySupervisedDisentanglementCompromises2020a, lachapelleDisentanglementMechanismSparsity2022}.

\paragraph{Independent causal mechanisms for action representations.}
Modeling the effect of an action essentially amounts to learning the causal mechanism $p(\Tilde{\zb}_a \given \zb_a, \ab)$.
By the principle of Independent Causal Mechanism \citep{scholkopfCausalAnticausalLearning2012,petersElementsCausalInference2017}, the conditional distribution of each variable given its causes does not inform or influence the other conditional distributions.
In other words, $\PP(\Tilde{\zb}_a \given \zb_a, \ab)$ should remain invariant when other mechanisms,
such as $\PP(\ab \given \cbb)$ and $\PP(\zb \given \cbb)$, change.
We hypothesize that this principle is also essential in tackling practical problems like $\PP(\ab \given \xb, \Tilde{\xb})$ that is anti-causal.
In the presence of distribution shifts resulting from changes of unobserved confounders $\cbb$, the action representation that takes into account all the latent factors may break due to the non-stability of the statistical dependencies between the action class and unintervened latent factors. In contrast, the action representation only focused on the transition between the intervened latent factors $\PP(\ab \given \zb_a, \Tilde{\zb}_a)$ is expected to strongly generalize.

\newcommand{\imgheight}{0.067}

\begin{figure}[t]
\vspace{-5pt}
\begin{minipage}[t]{0.45\linewidth} 
    \centering
    \includegraphics[height=\imgheight\textheight]{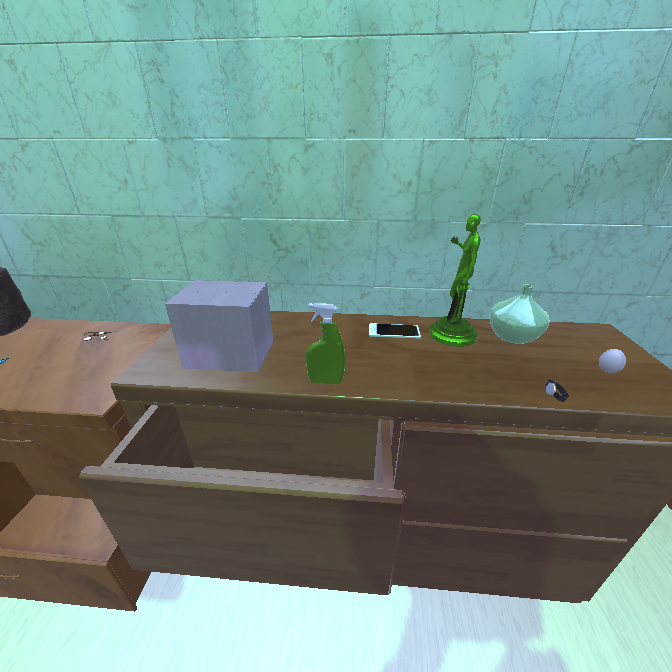}
    \includegraphics[height=\imgheight\textheight]{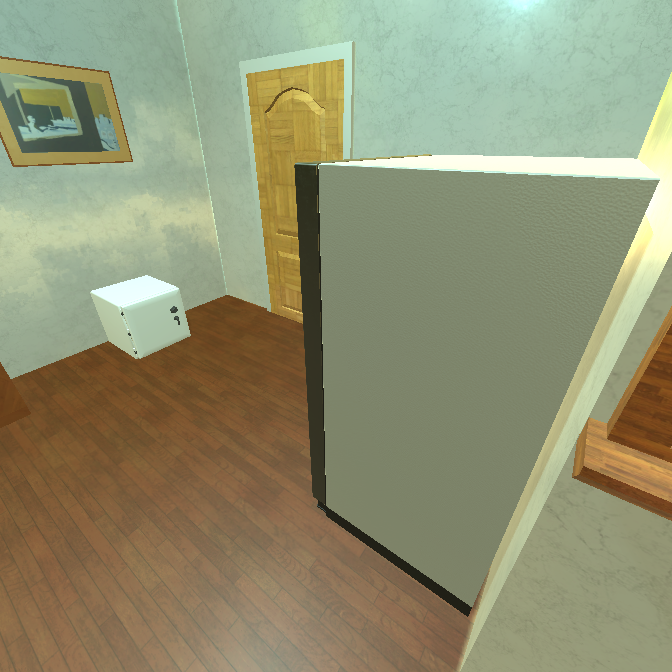}
    \includegraphics[height=\imgheight\textheight]{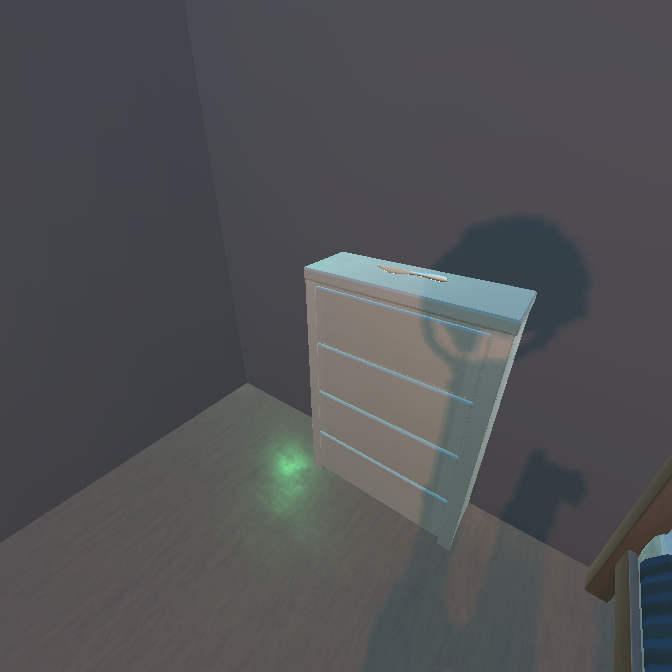}
    \includegraphics[height=\imgheight\textheight]{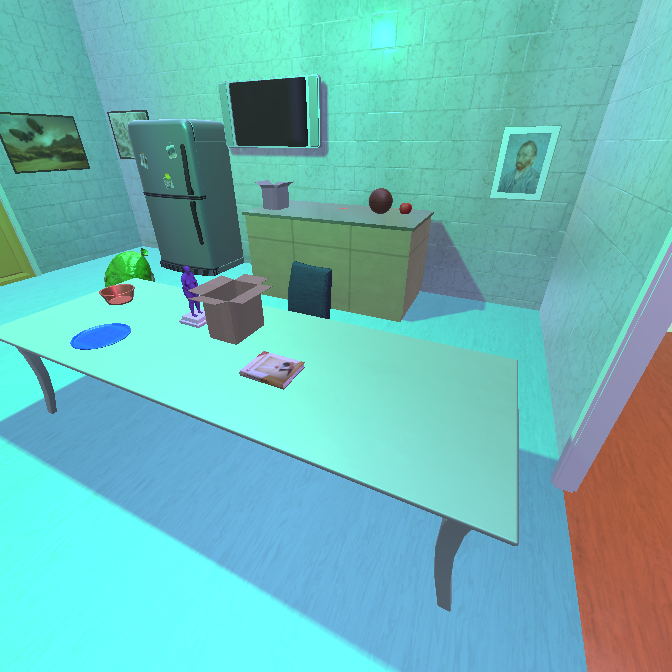}
    \\
    \includegraphics[height=\imgheight\textheight]{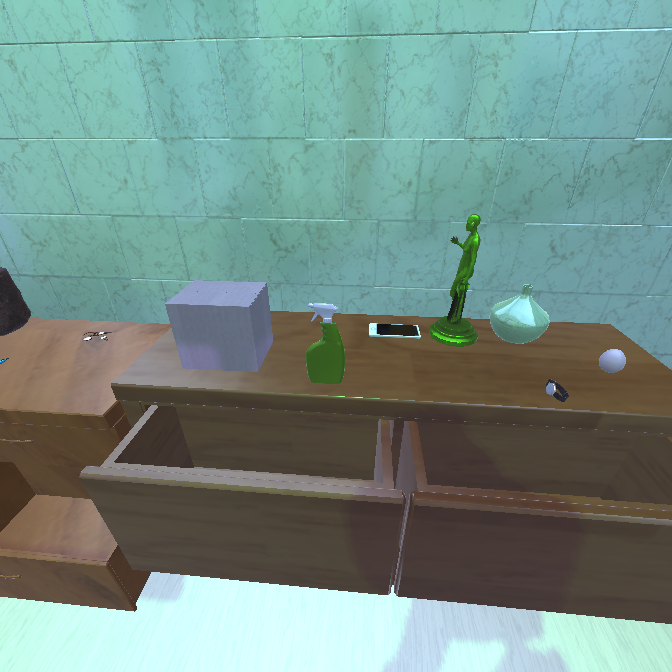}
    \includegraphics[height=\imgheight\textheight]{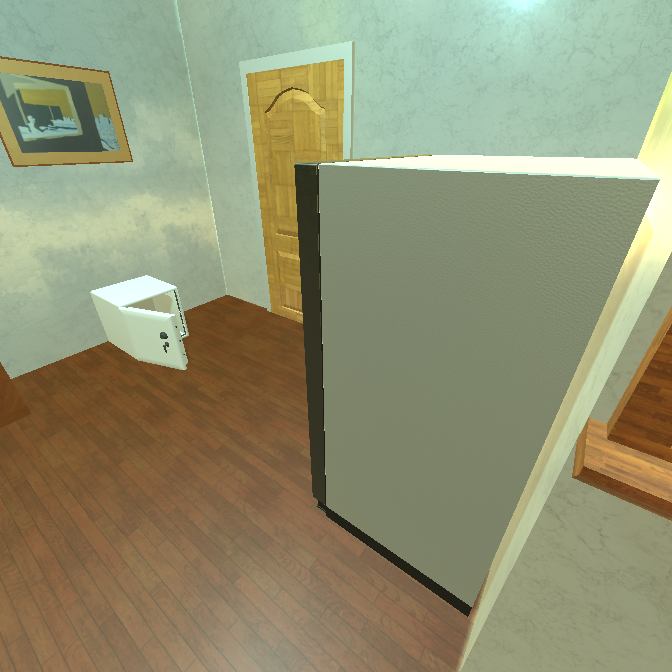}
    \includegraphics[height=\imgheight\textheight]{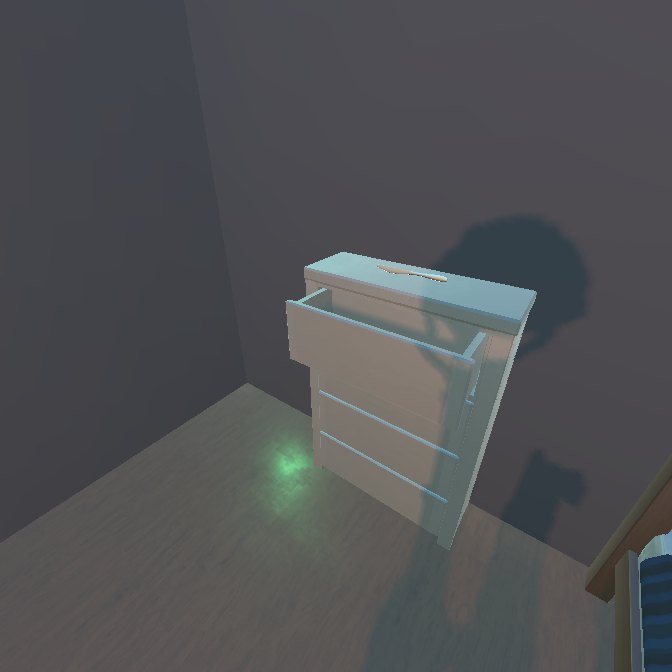}
    \includegraphics[height=\imgheight\textheight]{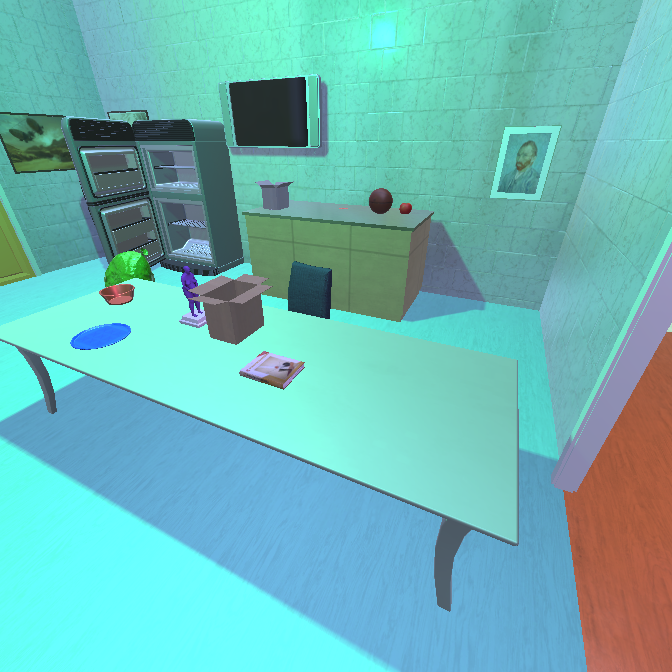}
    \caption{Causal triplet samples collected from photo-realistic simulations.}
    \label{fig:thor}
\end{minipage}
\quad
\begin{minipage}[t]{0.52\linewidth}
    \centering
    \includegraphics[height=\imgheight\textheight]{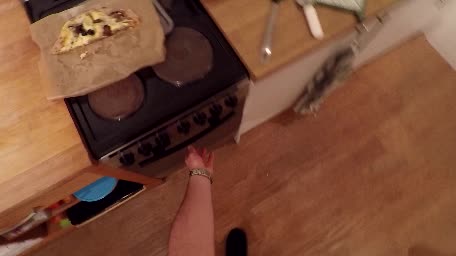}
    \includegraphics[height=\imgheight\textheight]{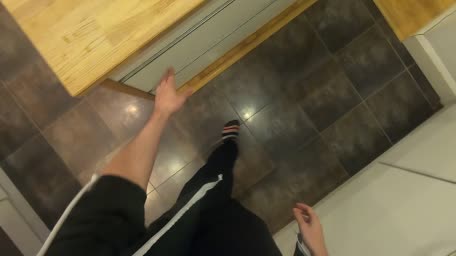}
    \includegraphics[height=\imgheight\textheight]{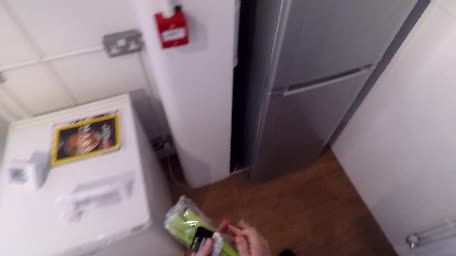}
    \\
    \includegraphics[height=\imgheight\textheight]{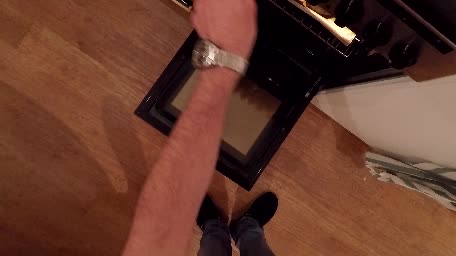}
    \includegraphics[height=\imgheight\textheight]{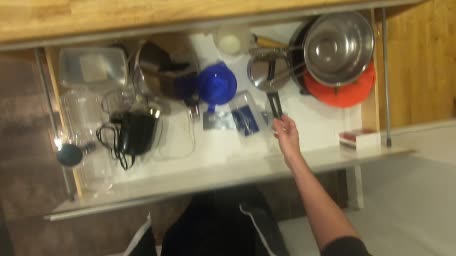}
    \includegraphics[height=\imgheight\textheight]{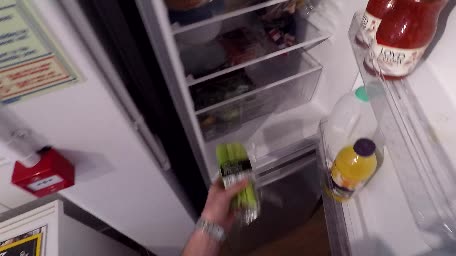}
    \caption{Causal triplet samples collected from real-world observations.}
    \label{fig:moon_quantitative}
\end{minipage}
\end{figure}

\subsection{Benchmark Data}
\label{subsec:data}
Existing datasets for causal representation learning~\citep{vonkugelgenSelfSupervisedLearningData2021,ahmedCausalWorldRoboticManipulation2021,lippeICITRISCausalRepresentation2022,lippeCITRISCausalIdentifiability2022}, as listed in \cref{tab:benchmark}, are heavily simplified in many aspects compared to real-world problems.
To bridge this gap, our benchmark introduces two new datasets, one collected from a photo-realistic simulator of embodied agents and the other repurposed from a real-world video dataset of human-object interactions.
The former one contains $7$ types of actions manipulating $24$ types of objects in $10k$ distinct ProcTHOR indoor environments~\citep{deitkeProcTHORLargeScaleEmbodied2022}, resulting in $100k$ pairs of images. Each original image has a resolution of $672 \times 672$, and is further cropped and/or resize to a resolution of $224 \times 224$ for the experiments in \cref{sec:experiment}.
The latter consists of $2632$ image pairs, collected under a similar setup from the Epic-Kitchens dataset~\citep{damenRescalingEgocentricVision2022}.

Similar to recent empirical studies towards causal representation learning~\citep{vansteenkisteAreDisentangledRepresentations2019,dittadiTransferDisentangledRepresentations2020,monteroRoleDisentanglementGeneralisation2021,zhangACREAbstractCausal2021,liuRobustAdaptiveMotion2022a,dittadiGeneralizationRobustnessImplications2022a}, we explicitly split the collected data into two groups that differ in the joint distribution of action category $\ab \in \set{A}$ and object category $\ob \in \set{O}$.
More specifically, we consider two types of distribution shifts:
\begin{itemize}[nosep]
    \item {\em compositional shifts}: the sets of object class are identical between training and test, \textit{i.e.}, $\set{O}_\tr = \set{O}_\te$, but the sets of object-action composition are disjoint, \textit{i.e.}, $(\set{A}_\tr \times \set{O}_\tr) \ \cap (\set{A}_\te \times \set{O}_\te) = \emptyset$
    \item {\em systematic shifts}: the training and test sets of object class are disjoint, \textit{i.e.}, $\set{O}_\tr \cap \set{O}_\te = \emptyset$
\end{itemize}
We partition each dataset into three parts: $~60\%$ for training, $~20\%$ for in-distribution (ID) testing, and $~20\%$ for out-of-distribution (OOD) testing.
More dataset details are summarized in \cref{sec:appendix}.


\section{Experiments and Results}
\label{sec:experiment}

\begin{figure}[t]
    \vspace{-5pt}
    \centering
    \small
    \subfigure[unseen compositions]{
        \includegraphics[width=0.48\textwidth]{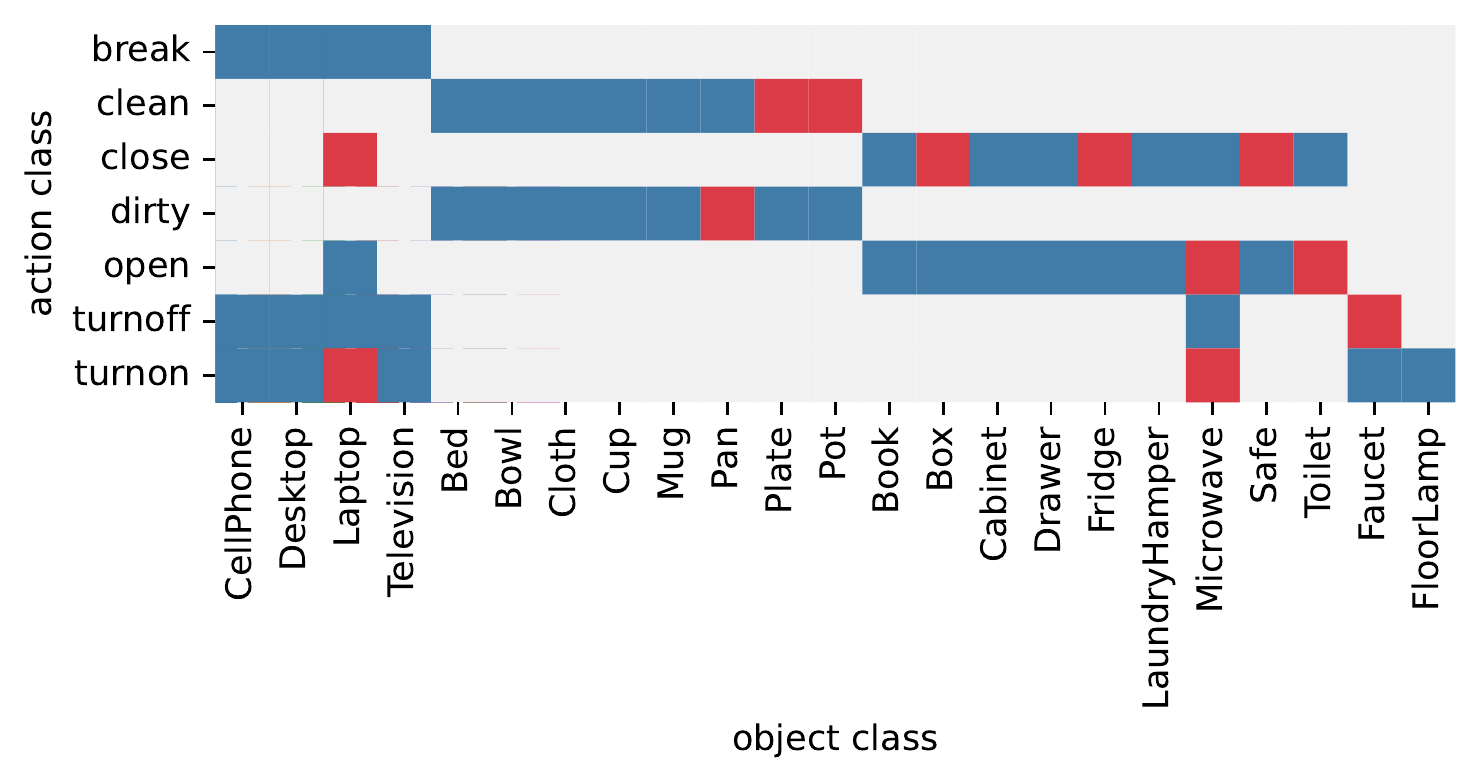}\label{fig:split_comp}
    }
    \subfigure[unseen objects]{
        \includegraphics[width=0.48\textwidth]{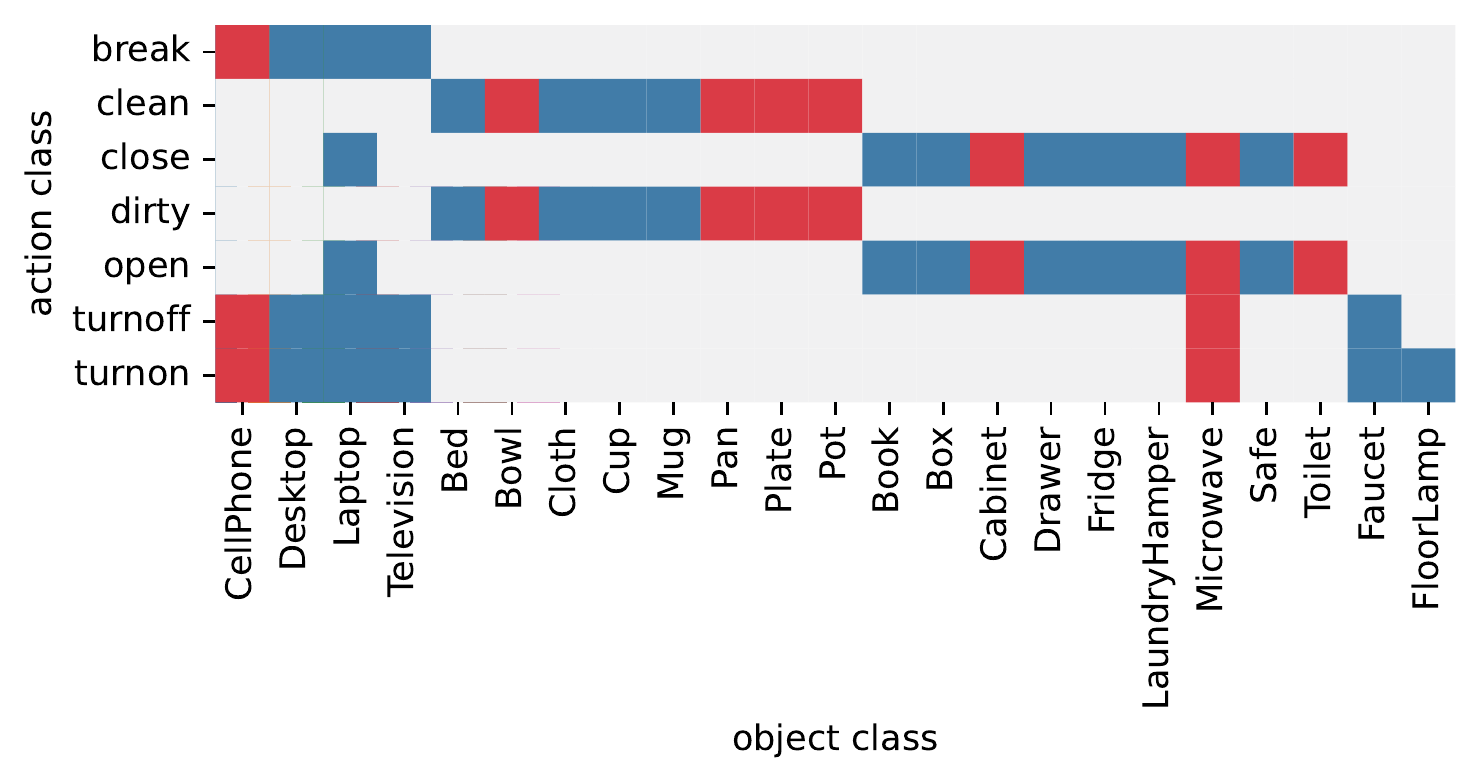}\label{fig:split_noun}
    }
    \caption{Data splits in \causaltriplet. We split the training (blue) and test (red) data into two disjoint groups with different (a) object-action combinations or (b) object classes.}
\end{figure}

\begin{figure}[t]
\centering
\begin{minipage}[t]{0.475\linewidth} \vspace{0pt}
    \centering
    \includegraphics[width=0.95\textwidth]{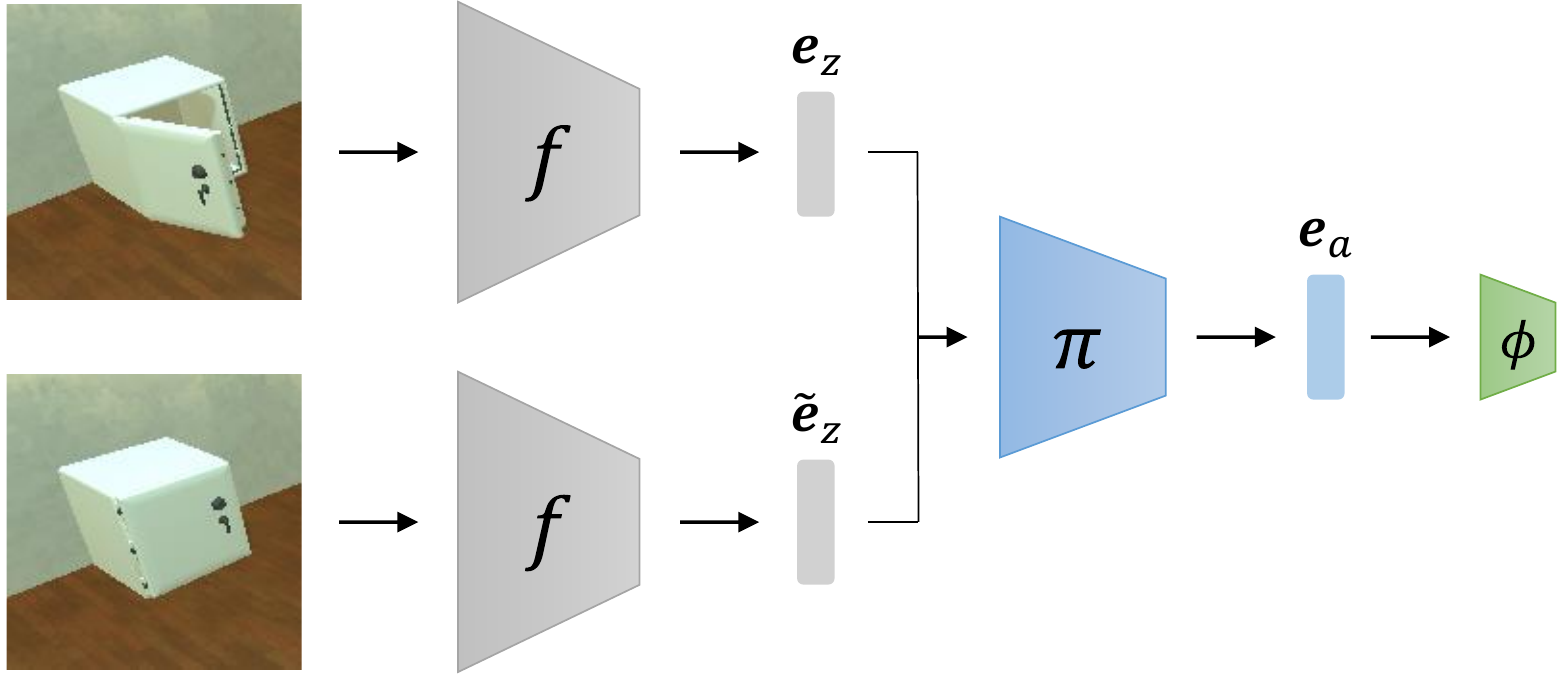}
    \caption{Overview of our benchmark model for reasoning about the action class from a pair of single-object images.
    We consider different regularizers for learning robust image representation $\eb_z$ and action representation $\eb_a$.
    }
    \label{fig:overview-single}
\end{minipage}
\quad
\begin{minipage}[t]{0.475\linewidth}\vspace{5.5pt}
    \centering
    \resizebox{1.0\textwidth}{!}{%
    \begin{tabular}{cccccc}
    \toprule
                            & \multicolumn{2}{c}{compositional shift} & \multicolumn{2}{c}{systematic shift} \\ \cmidrule(l{5pt}r{5pt}){2-3} \cmidrule(l{5pt}r{5pt}){4-5}
                            & ID              & OOD             & ID            & OOD            \\ \midrule
    vanilla                 & $0.96 \pm 0.01$ & $0.36 \pm 0.13$ & $0.94 \pm 0.02$ & $0.48 \pm 0.08$  \\  \midrule
    ICM                     & $0.95 \pm 0.01$ & $0.41 \pm 0.15$ & $0.94 \pm 0.02$ & $0.50 \pm 0.09$ \\ \midrule
    SMS                     & $0.96 \pm 0.01$ & $0.47 \pm 0.18$ & $0.94 \pm 0.02$ & $0.54 \pm 0.07$ \\  \bottomrule
    \end{tabular}
    }
    \captionof{table}{Results of different methods in single-object simulated images. The action representation trained with the independence regularizer (ICM) and the image representation trained with the sparsity regularizer (SMS) result in significantly higher OOD accuracies.}
    \label{tab:single}
\end{minipage}
\end{figure}


The main goal of our experiments is to examine the potential and limitations of recent hypotheses and methods for causal representation learning on \causaltriplet.
In particular, we seek to understand the following two questions:
\begin{itemize}[nosep]
    \item Are causally structured representations (\textit{e.g.}, disentangled, object-centric) helpful for reasoning about interventions between paired observations?
    \item How effective are recent models at identifying the latent structures in the proposed datasets?
\end{itemize}
To this end, we consider a variety of visual representations, including
\begin{itemize}[nosep]
\item modern distributed representations, \textit{e.g.}, ResNet~\citep{heDeepResidualLearning2016}, CLIP~\citep{radfordLearningTransferableVisual2021a};
\item oracle structured representations, \textit{e.g.}, expert knowledge of disentanglement / object-centric;
\item learned structured representations, \textit{e.g.}, Slot Attention~\citep{locatelloObjectCentricLearningSlot2020a}, GroupViT~\citep{xuGroupViTSemanticSegmentation2022a}.
\end{itemize}
We systematically examine their performance in four different settings with growing complexities:
\begin{itemize}[nosep]
\item from compositional shifts (\cref{subsec:comp}) to systematic shifts (\cref{subsec:syst});
\item from single-object images (\cref{subsec:syst}) to multi-object scenes (\cref{subsec:thor});
\item from photo-realistic simulations (\cref{subsec:thor}) to real-world observations (\cref{subsec:epic}).
\end{itemize}
To model the effect of high-level actions between paired images, we consider neural networks made up of three modules: (i) an image encoder $f(\cdot)$ that extracts an abstract representation of each input image, \textit{i.e.}, $\eb_z = f(\xb)$, (ii) an action encoder $\pi(\cdot)$ that reasons about the relation between paired image embeddings, \textit{i.e.}, $\eb_a = \pi(\eb_z, \Tilde{\eb}_z)$, (iii) a classification head $\phi(\cdot)$ that infers the probabilities for each action class from the action embedding.

\subsection{Simulated Single-object Images}

\subsubsection{Compositional Distribution Shifts}
\label{subsec:comp}

\paragraph{Setup.}
We first consider intervention modeling in single-object images collected from photo-realistic simulations.
As shown in \cref{fig:split_comp}, the training and test data are split into two disjoint groups that differ in object-action combinations.
We build the image encoder $f(\cdot)$ with a standard ResNet-18, followed by a two-layer MLP projection head that outputs a 64-dimensional feature vector.
To model the action between an image pair, we concatenate the feature vectors obtained from both images, and feed them into a downstream two-layer MLP $\pi(\cdot)$ to extract a 64-dimensional action representation.
We finally pass the action representation through a linear classifier $\phi(\cdot)$ that outputs the probabilities for each action class.
By default, the model is trained to minimize a standard classification loss $\loss_a$, \textit{i.e.}, cross-entropy between the predicted class probabilities $\hat\ab$ and the action label $\ab$.

\paragraph{Vanilla baseline.}
We start with a vanilla baseline trained in a standard supervised manner.
As shown in \cref{tab:single}, the vanilla baseline is highly accurate ($\sim96\%$) on the ID test set, but generalizes poorly ($\sim36\%$) to unseen compositions.
The large gap is unsurprising though, given that the action label is highly correlated with the object label in the training set, and this correlation varies drastically between the ID and OOD settings.
In order to understand the degree to which the vanilla baseline exploits such spurious correlations, we conduct a post-hoc analysis: freeze the trained model and train another classification head $\psi (\cdot)$ to predict the object label.
The object classifier built on top of the frozen action embedding turns out to be quite accurate ($\sim87\%$) on the test set, revealing the high reliance of the learned action embedding on the non-causal object features.

\begin{figure}[t]
    \centering
    \small
    \subfigure[ID]{
        \includegraphics[width=0.34\textwidth]{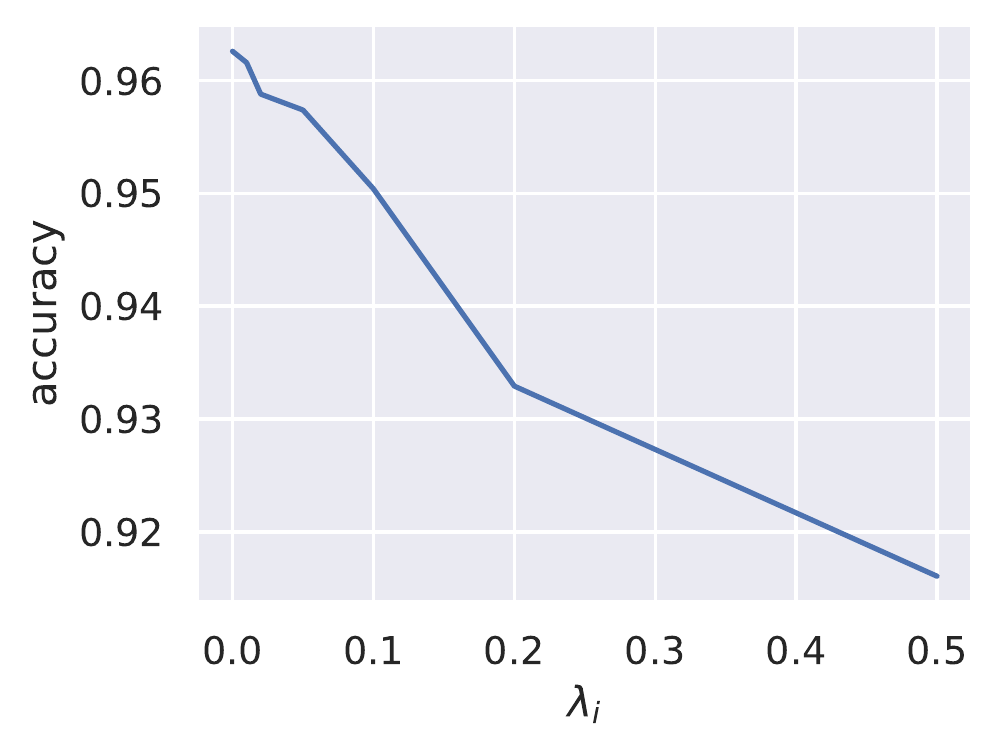}\label{fig:iid_critic}
    }
    \quad
    \subfigure[OOD]{
        \includegraphics[width=0.34\textwidth]{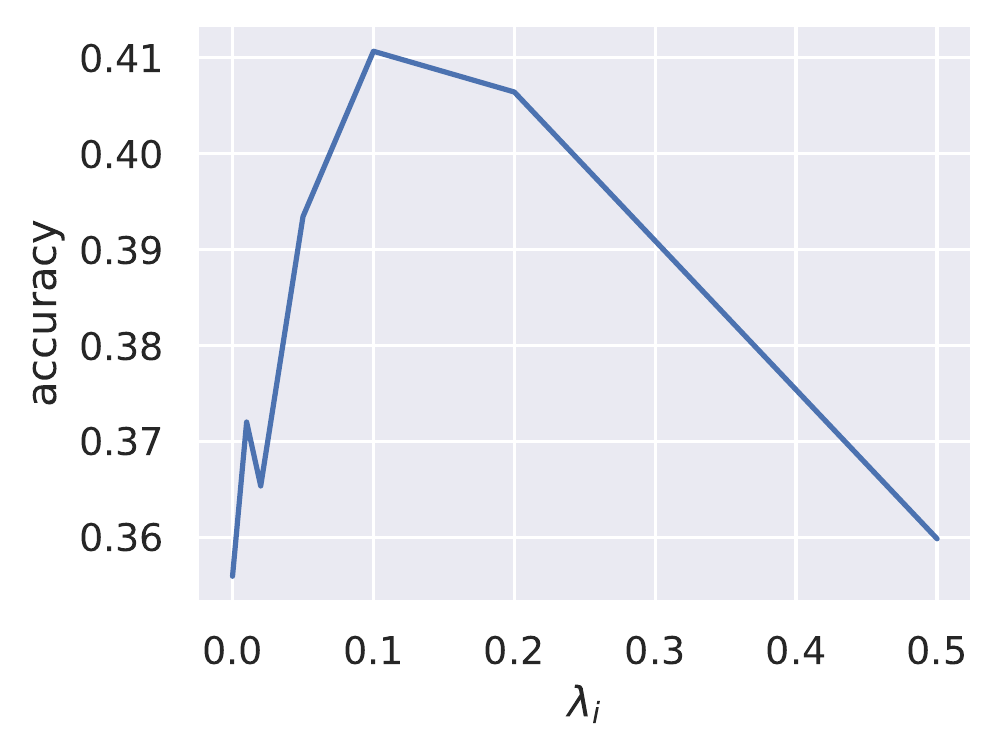}\label{fig:ood_critic}
    }
    \caption{
    Effect of the independence regularizer on compositional generalization in single-object images.
    Action representations containing less information about object classes (larger $\lambda_i$) result in smaller performance gaps between the ID and OOD sets that differ in object-action compositions.
    }
    \label{fig:comp_critic}
\end{figure}

\paragraph{Independence regularizer.}
To prevent an intervention model from absorbing the spurious correlations demonstrated above, we next consider regularizing the model by minimizing the mutual information between the action embedding and the object class.
More specifically, we use the following adversarial training regularizer as a proxy,
\begin{equation}
    \min_{f,\pi,\phi} \max_{\psi} \loss_a(f,\pi,\phi) - \lambda_i \loss_o(f,\pi,\psi)
\end{equation}
where $\loss_o$ is a cross-entropy loss for object classification, and $\lambda_i$ is the coefficient of the independence regularizer.
The adversarial regularizer shares the same spirit with the Hilbert-Schmidt Independence Criterion (HSIC) used in recent works for compositional object recognition \citep{atzmonCausalViewCompositional2020, ruisIndependentPrototypePropagation2021}. 
We opt for adversarial training as a natural extension of the post-hoc analysis in the previous section.
\cref{fig:comp_critic} shows that, while the independence regularizer constrains the performance on the ID test set, it greatly improves the OOD accuracy from $36\%$ to $41\%$, resulting in $\sim 14\%$ relative performance gain on unseen compositions.

Nevertheless, it is worth noting that the independence regularizer relies on two critical assumptions: 
(i) the attribute spuriously correlated with the action label is known a priori; 
(ii) the spurious attribute comes with detailed annotations in the training dataset.
Unfortunately, it is typically impractical to examine all possible spurious correlations and gather their corresponding annotations.
We next consider another family of regularizations that does not rely on such prior knowledge.

\begin{figure}[t]
    \vspace{-10pt}
    \centering
    \small
    \subfigure[ID]{
        \includegraphics[width=0.34\textwidth]{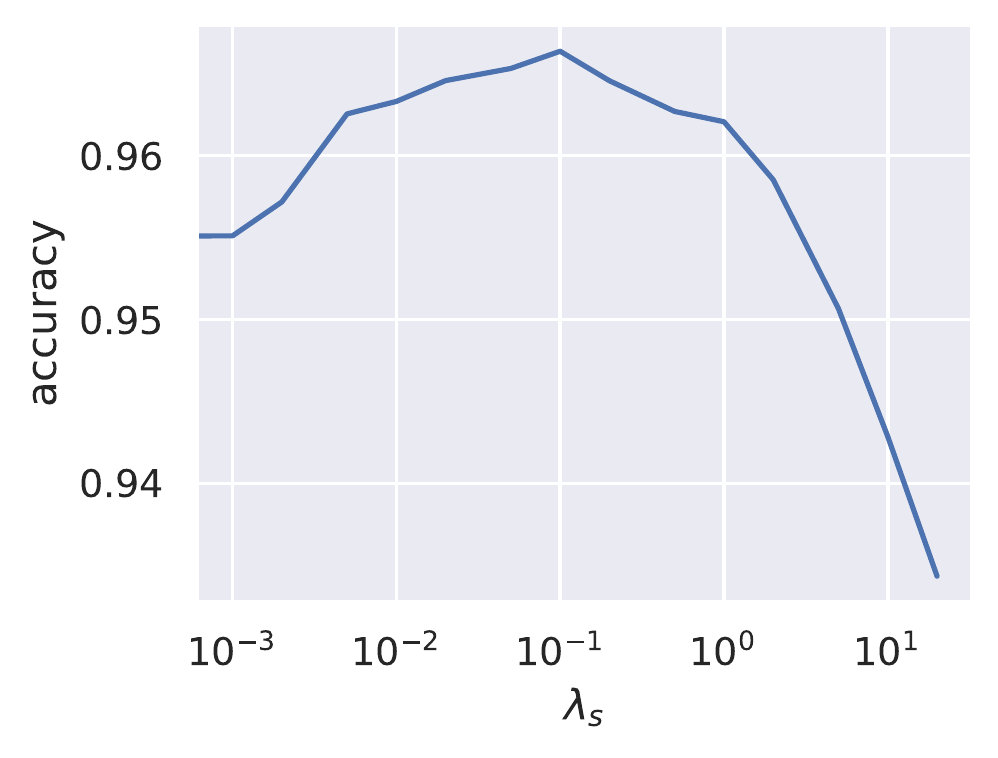}\label{fig:iid_sparsity}
    }
    \quad
    \subfigure[OOD]{
        \includegraphics[width=0.34\textwidth]{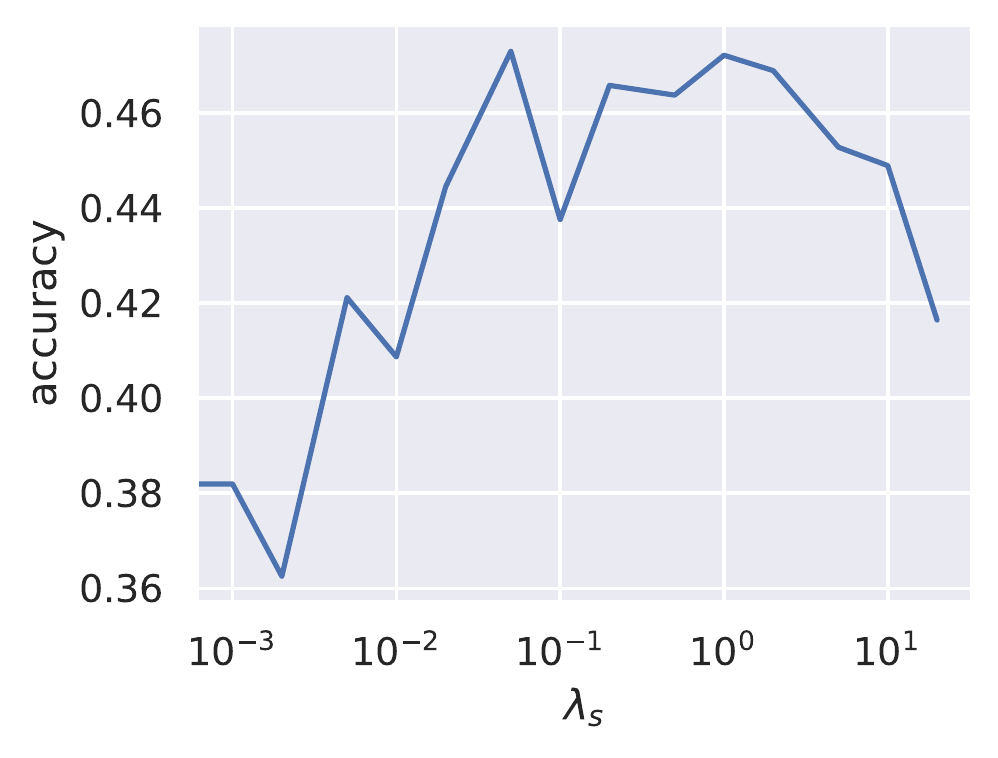}\label{fig:ood_sparsity}
    }
    \caption{
    Effect of the sparsity regularizer on compositional generalization in single-object images.
    Visual representations undergoing sparser changes between paired observations (larger $\lambda_s$) result in smaller performance gaps between the ID and OOD sets that differ in object-action compositions.
    }
    \label{fig:result_sparsity}
\end{figure}

\begin{figure}[t]
    \vspace{5pt}
    \centering
    \small
    \subfigure[$\lambda_s=0.0$]{
        \includegraphics[width=0.15\textwidth]{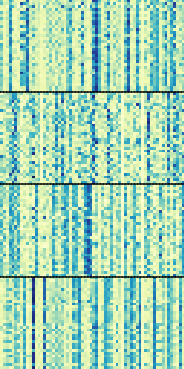}
    }
    \subfigure[$\lambda_s=0.1$]{
        \includegraphics[width=0.15\textwidth]{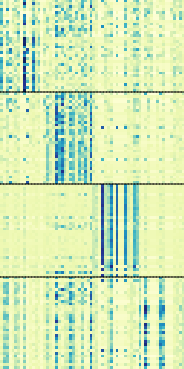}
    }
    \subfigure[$\lambda_s=1.0$]{
        \includegraphics[width=0.15\textwidth]{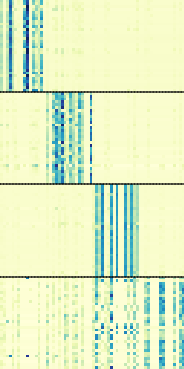}
    }
    \caption{
    Effect of the sparsity regularizer on image representation.
    We randomly sample $32$ image pairs from each symmetric group of actions (open vs.\ close, turnon vs.\ turnoff, clean vs.\ dirty, break).
    We visualize the difference of the encoded feature pairs $\norm{\eb_z - \Tilde{\eb}_z}$ in each row.
    Larger regularizer coefficients $\lambda_s$ result in sparser feature changes between paired observations and more distinct block-disentanglement across action classes.
    }
    \label{fig:feat_sparsity}
\end{figure}

\paragraph{Sparsity regularizer.}
We further investigate the impact of visual representation structures on the robustness of intervention models.
As discussed in \cref{subsec:task}, the action between paired images tends to affect only one or a few generative factors in the causal/disentangled factorization.
Motivated by this hypothesis, we introduce a regularizer that imposes sparse changes in a block-disentangled image representation before and after intervention,
\begin{equation}
    \min_{f,\pi,\phi} \loss_a(f,\pi,\phi) + \lambda_s \loss_s(f,\pi), 
    \quad \loss_s = \sum_{k \in \Omega \setminus \Omega_a} \norm{\eb_z^k - \Tilde{\eb}_z^k},
\end{equation}
where $k$ is the block index, $\Omega$ is the entire set of latent blocks, and $\Omega_a$ is the subset of latent blocks affected by the action, and $\lambda_s$ is the coefficient of the sparsity regularizer.
\cref{fig:result_sparsity} and \cref{fig:feat_sparsity} shows that the effect of the sparsity regularizer on test accuracy and visual representations.
\update{
Notably, the sparsity regularizer with a moderate coefficient ($\lambda_s \approx 0.1$) demonstrates advantages on both the ID and OOD test sets.
In particular, compared with the vanilla baseline in \cref{tab:single},
the sparsity-driven disentanglement lifts the OOD accuracy by nearly $30\%$.
}
Nevertheless, same as \citet{lippeInterventionDesignCausal2022,lippeCITRISCausalIdentifiability2022}, the sparsity regularizer requires variable-level intervention labels, \textit{i.e.}, not only is the category of each action labeled, but so is its influence on each latent factor.
While this is partially feasible in our simulated data due to the known symmetry between some actions (open vs.\ close, dirty vs.\ clean, turnon vs.\ turnoff), collecting such detailed annotations in real-world problems can be difficult.

\subsubsection{Systematic Distribution Shifts}
\label{subsec:syst}

\paragraph{Setup.}
We further consider systematic distribution shifts between training and test in simulated single-object images.
As shown in Figure~\ref{fig:split_noun}, the object class in the ID and OOD data splits are sampled from two disjoint groups.
Same as in the previous section, we compare intervention models trained with different loss functions. 

\paragraph{Results.}
\cref{tab:single} summarizes the results on the ID and OOD test sets over 10 different runs.
Similar to the observations in \cref{subsec:comp}, both the independence and sparsity regularizers improve robustness on the OOD test set without sacrificing the ID accuracy.
Nevertheless, regularizing the action representation to be independent of the object class becomes less effective for robustness under systematic shifts than it is under compositional shifts.
We conjecture this is because the major challenge here lies in visual representations of unseen objects, where disentanglement encouraged by the sparsity regularizer remains highly beneficial.

\begin{figure}[t]
    \vspace{-5pt}
    \centering
    \small
    \includegraphics[width=\textwidth]{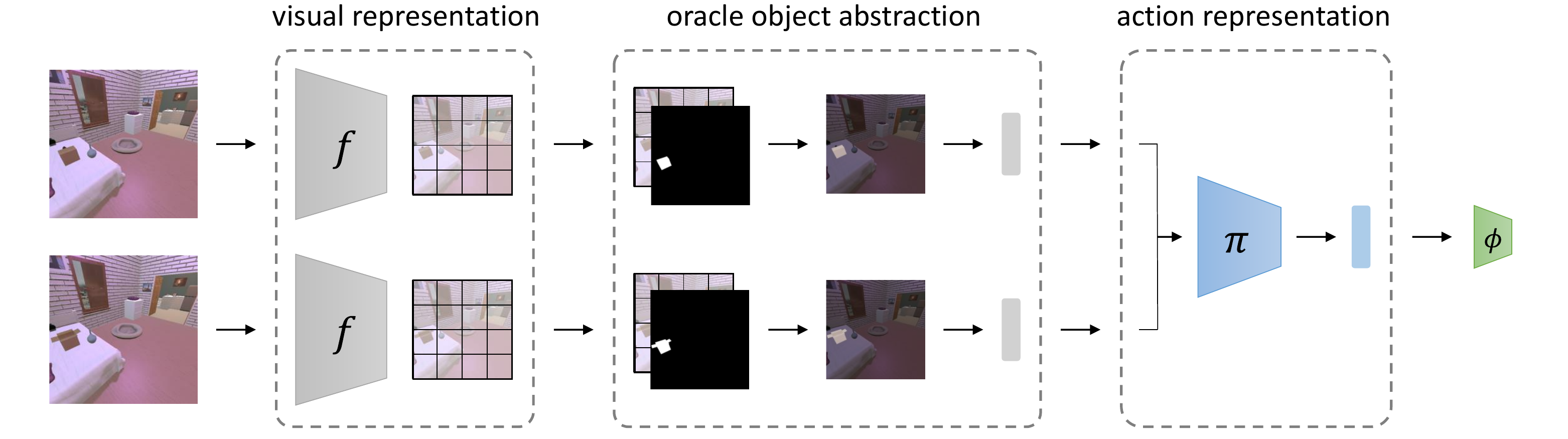}
    \caption{Overview of the intervention model built with the oracle object-centric representation.}
    \label{fig:overview_thor}
\end{figure}
\begin{figure}[t]
    \centering
    \subfigure[ID]{
        \includegraphics[width=0.34\textwidth]{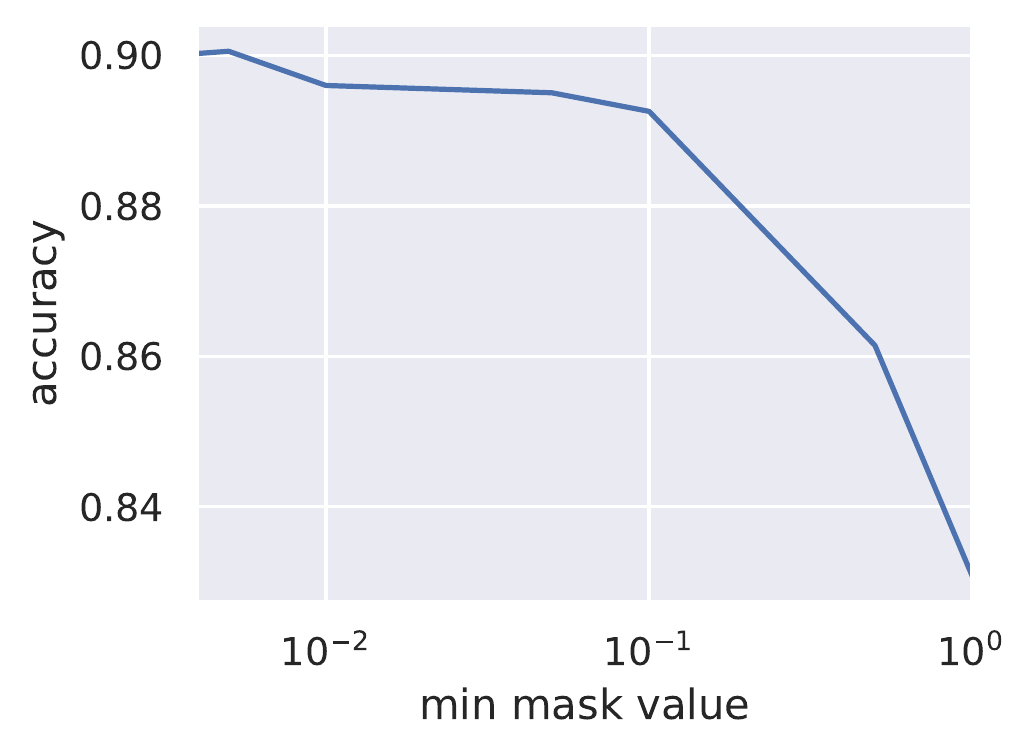}\label{fig:iid_mask}
    }
    \quad
    \subfigure[OOD]{
        \includegraphics[width=0.34\textwidth]{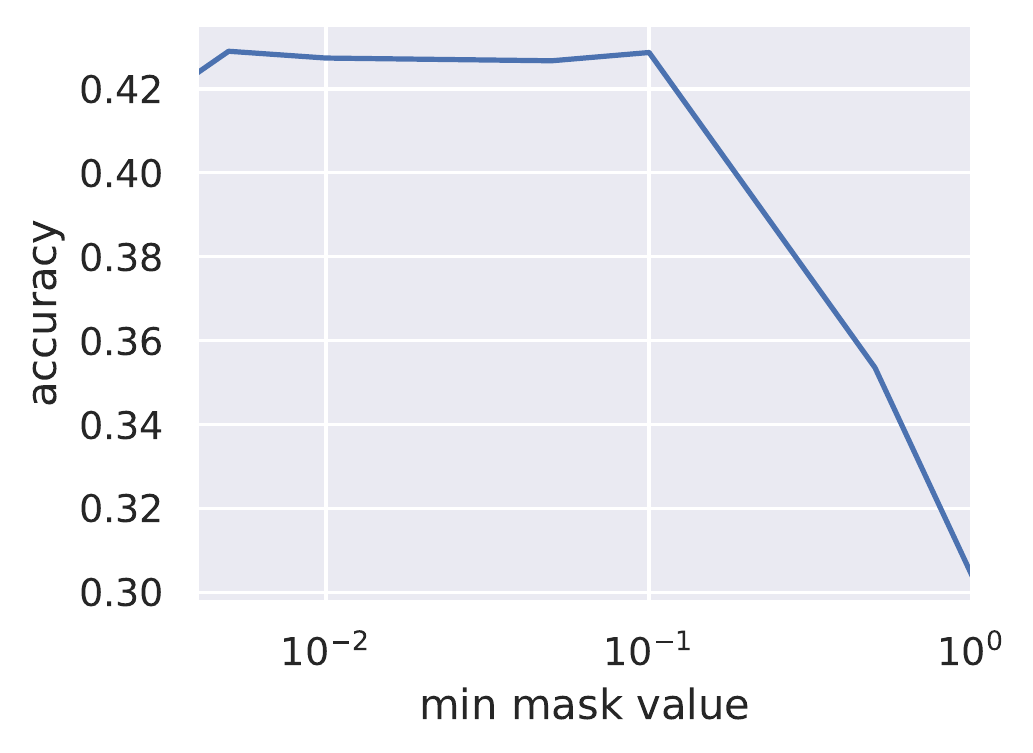}\label{fig:ood_mask}
    }
    \caption{
    Effect of the oracle object-centric representation in multi-object scenes.
    Models with clearer object-level abstractions (lower background mask value) perform better on both the ID and OOD sets that differ in object classes.}
    \label{fig:mask}
\end{figure}

\subsection{Multi-object Scenes}

\subsubsection{Simulated Multi-object Scenes}
\label{subsec:thor}

\paragraph{Setup.}
In contrast to the previous sections where each image contains only one object, we next consider more complex scenes composed of multiple objects.
Following the data split above, we train and test the model on two disjoint groups of object classes.
To represent high-level actions in multi-object scenes, we consider an extra module between the image encoder $f(\cdot)$ and action encoder $\pi(\cdot)$ for object-level abstraction.
\cref{tab:thor} summarizes the results of different design choices.

\paragraph{Distributed representation.}
Consistent with the observations in \cref{subsec:syst}, the intervention models trained on a subset of object classes generally suffer from significant performance drops on the unseen OOD objects.
Notably, while the architecture design of the distributed representation baseline is the same as that in the previous sections, the performance in the multi-object scenes (\cref{tab:thor}) is significantly worse than the counterparts in the single-object images (\cref{tab:single}).
We postulate that this is largely attributed to the lack of object-level abstractions prior to reasoning about the effect of an intervention.
To verify this, we next take a close look at intervention classifiers built with two variants of structured representations that support such high-level abstractions.

\begin{figure}[t]
    \vspace{-5pt}
    \centering
    \small
    \includegraphics[width=\textwidth]{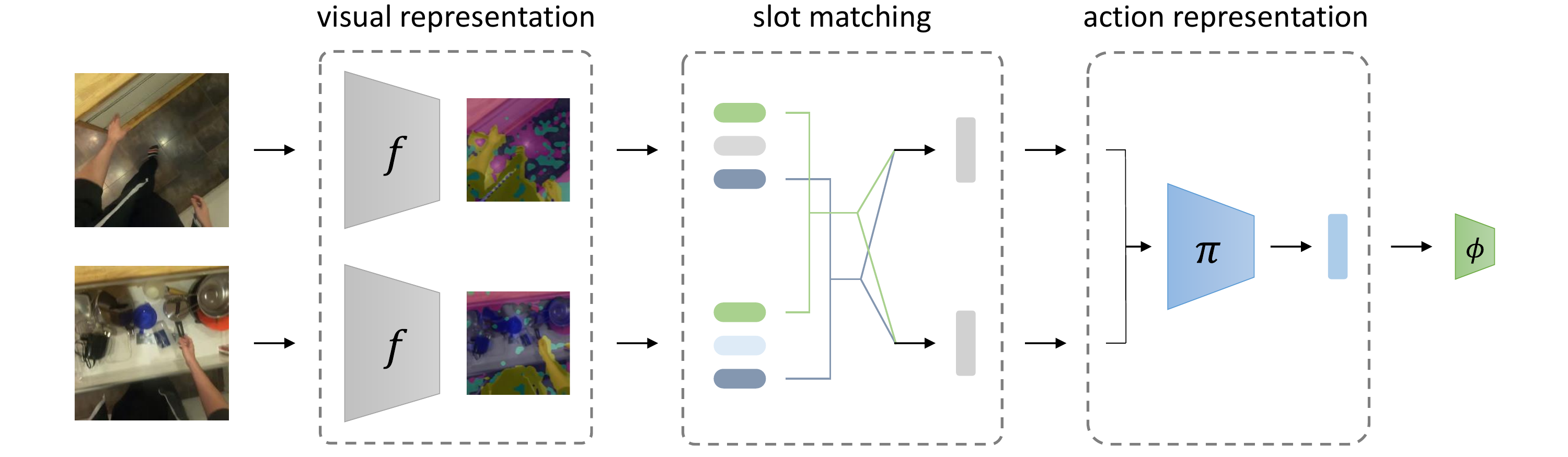}
    \caption{Overview of the intervention model built with object-centric visual representations.}
    \label{fig:overview_epic}
\end{figure}

\paragraph{Oracle object-centric representation.}
A natural way to exploit sparse mechanism shifts at the object-level is to explicitly decompose each scene into a set of constituent objects and selectively attend to the objects that undergo state changes.
Nevertheless, learning object-centric representations of complex scenes is often beyond reach in the first place~\citep{singhSimpleUnsupervisedObjectCentric2022,seitzerBridgingGapRealWorld2022a}.

To approximately estimate the potential benefits of this modeling hypothesis, we resort to an {\em oracle} object-centric representation by making use of the ground-truth instance segmentation obtained from the simulation engine.
As shown in \cref{fig:overview_thor}, each pixel in the segmentation map is a binary value, with $1$ representing the intervened object and $0$ representing the rest background.
In order to convert it to an object-level attention mask, we first resize the segmentation map to the dimensionality of the feature map, and then perform element-wise multiplication between the resized segmentation map and each channel of the feature map.
Moreover, in place of the default background value, we traverse the minimum mask value from $1$ to $10^{-3}$, which allows us to emulate object-level decomposition and attention in the representation space with varying quality.


\cref{tab:thor} and \cref{fig:mask} show the results of the intervention classifiers built with the oracle object-centric representations of different qualities.
Compared with the distributed representation baseline (the minimum mask value equals to $1$), incorporating the oracle object-level abstraction boosts performance on the ID and OOD test sets by up to $\sim8.5\%$ and $\sim40\%$, respectively.
Overall, reducing the background mask value consistently increases classification accuracy, demonstrating a clear advantage of capturing the sparse change at the object level for modeling intervention.
Meanwhile, we also notice a mild performance degradation when the min mask value approaches $0$.
We conjecture this is due to numerical instability in the case where the intervened object is very small in the scene.

\paragraph{Learned object-centric representation.}
We further investigate intervention models built with learned object-centric representations.
\update{
Specifically, we first pre-train on our dataset the implicit Slot Attention~\citep{locatelloObjectCentricLearningSlot2020a, changObjectRepresentationsFixed2022a}, a state-of-the-art unsupervised object-centric learning method, which decomposes an input scene into a number of ($N=15$) spatially and semantically related regions, each with its own feature vector.
To exploit the latent structure for downstream reasoning, we further consider three different slot matching schemes}:
\begin{itemize}[nosep]
    \item {\em slot-avg}: average-pooling over slots for each image, which degenerates to a distributed representation (only one pair of feature vectors);
    \item {\em slot-dense}: densely pair slots across the two images, pass all combinations to the action encoder, and aggregate $N \times N$ relation embeddings with average-pooling;
    \item {\em slot-match}: selectively pair slots across the two images based on the slot similarity, only pass the matched pairs to the action encoder, and aggregate $N$ relation embeddings with max-pooling, as shown in \cref{fig:overview_epic}.
\end{itemize}

\begin{figure}[t]
    \vspace{-5pt}
	\centering
    \includegraphics[height=50pt]{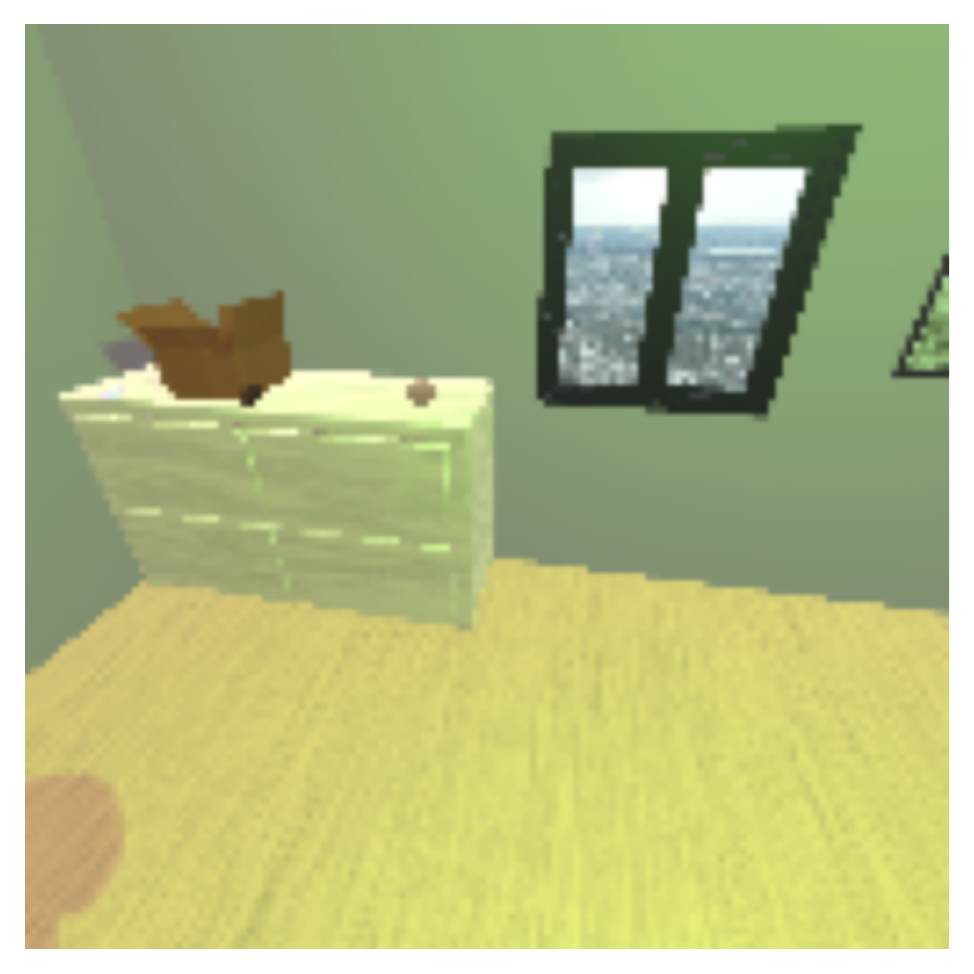}
    \includegraphics[height=50pt]{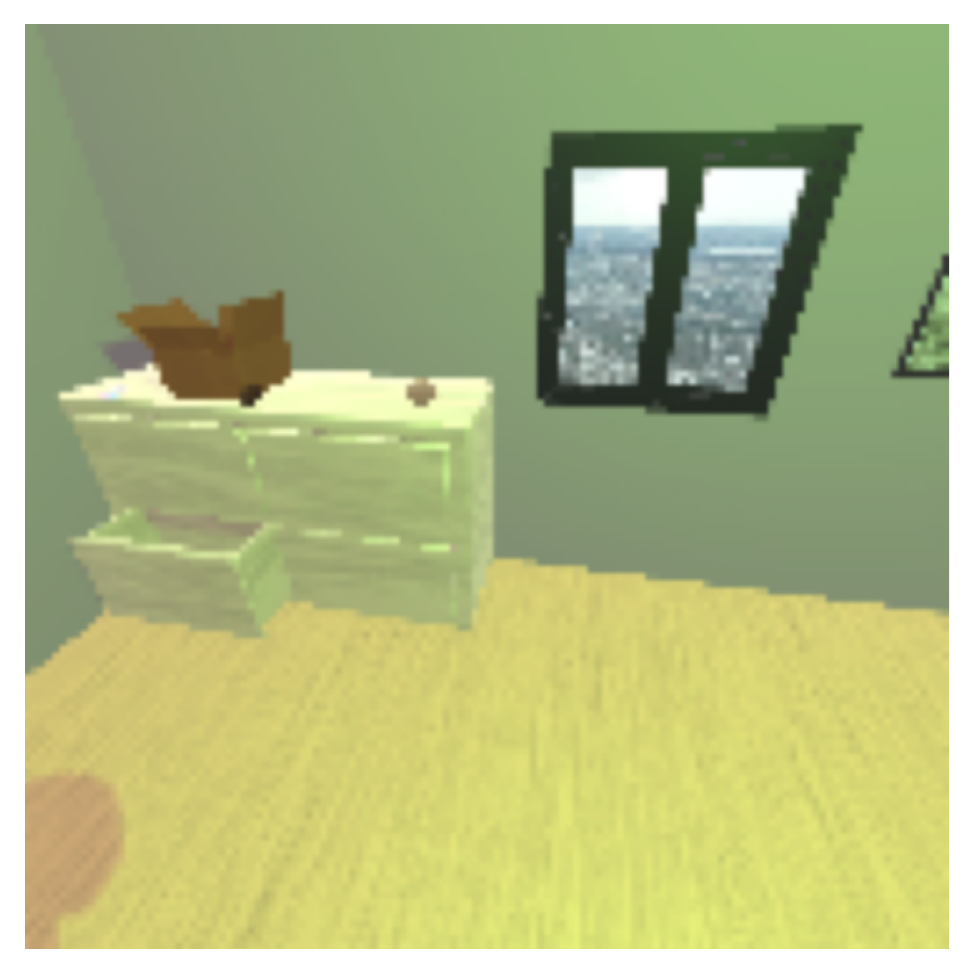}
    \includegraphics[height=50pt]{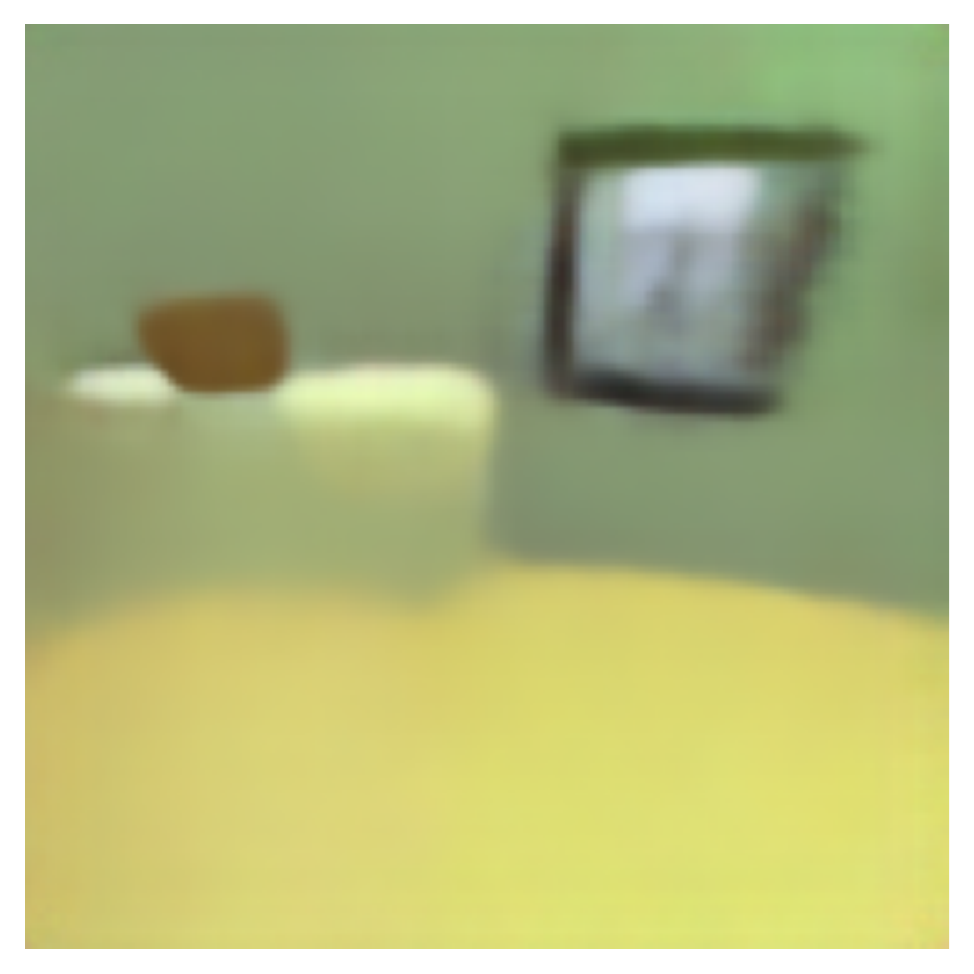}
    \includegraphics[height=50pt]{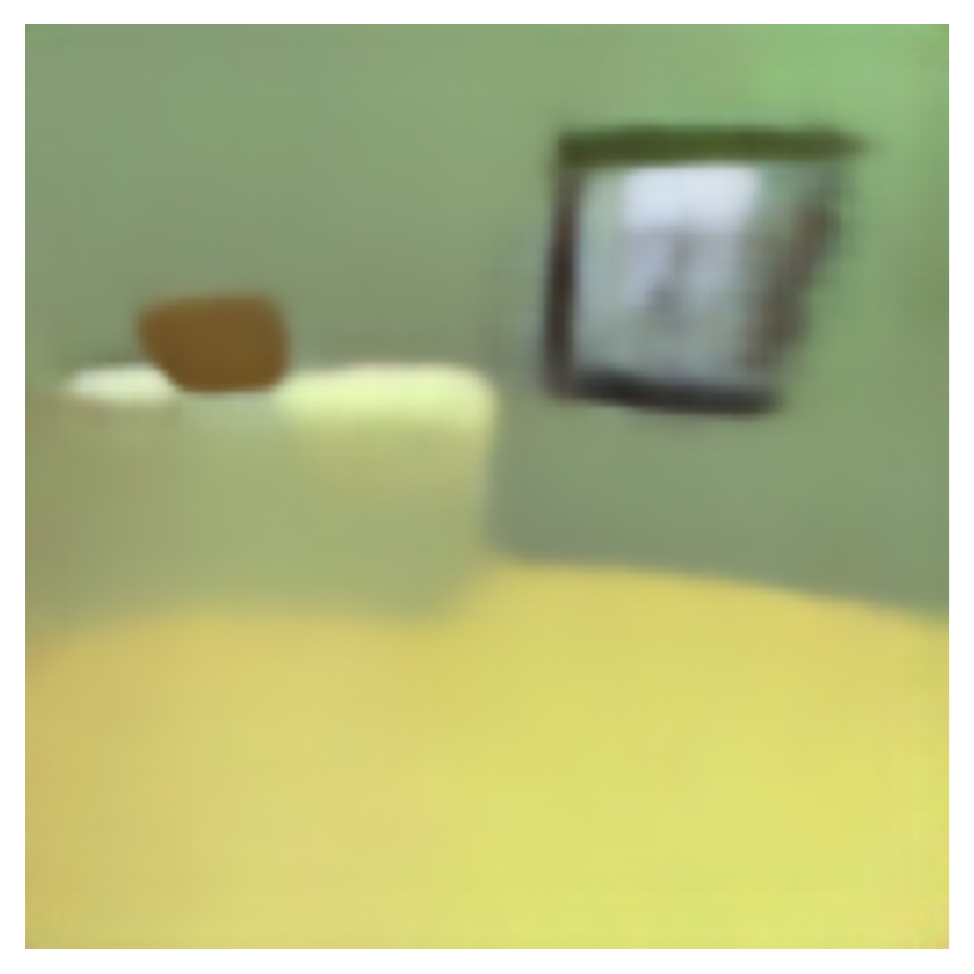}
    \includegraphics[height=50pt]{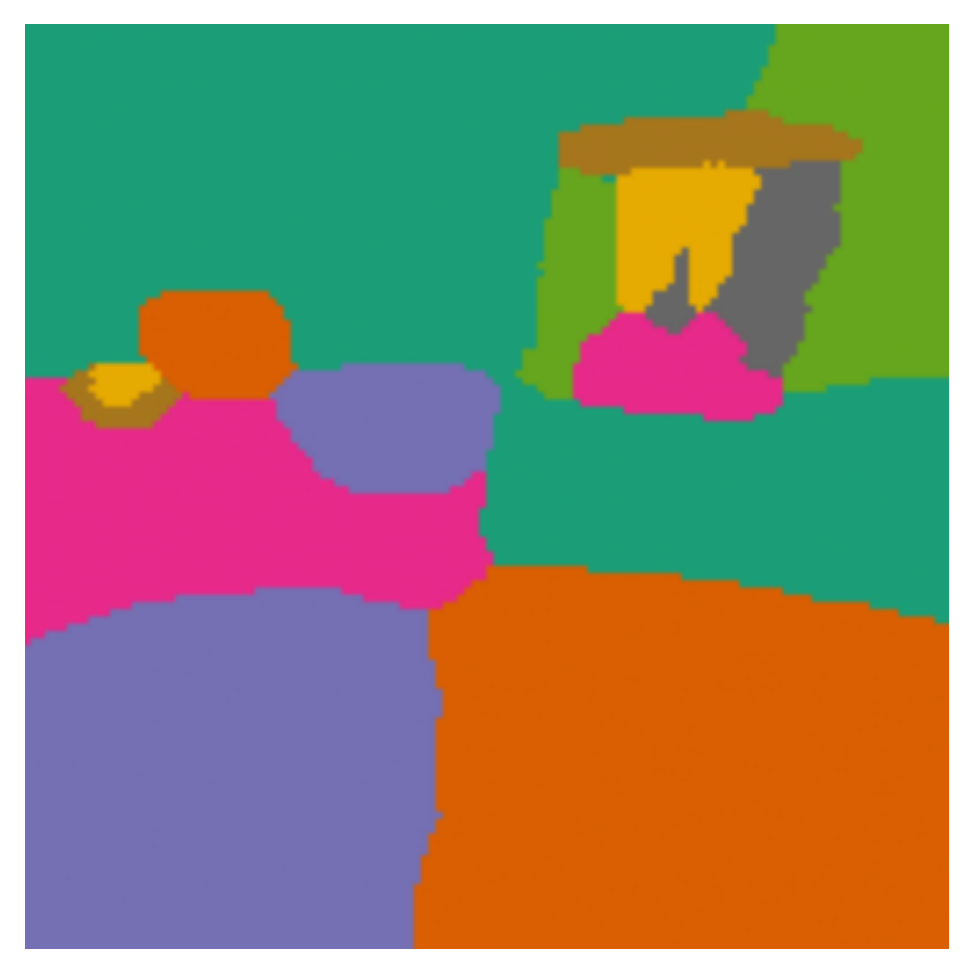}
    \includegraphics[height=50pt]{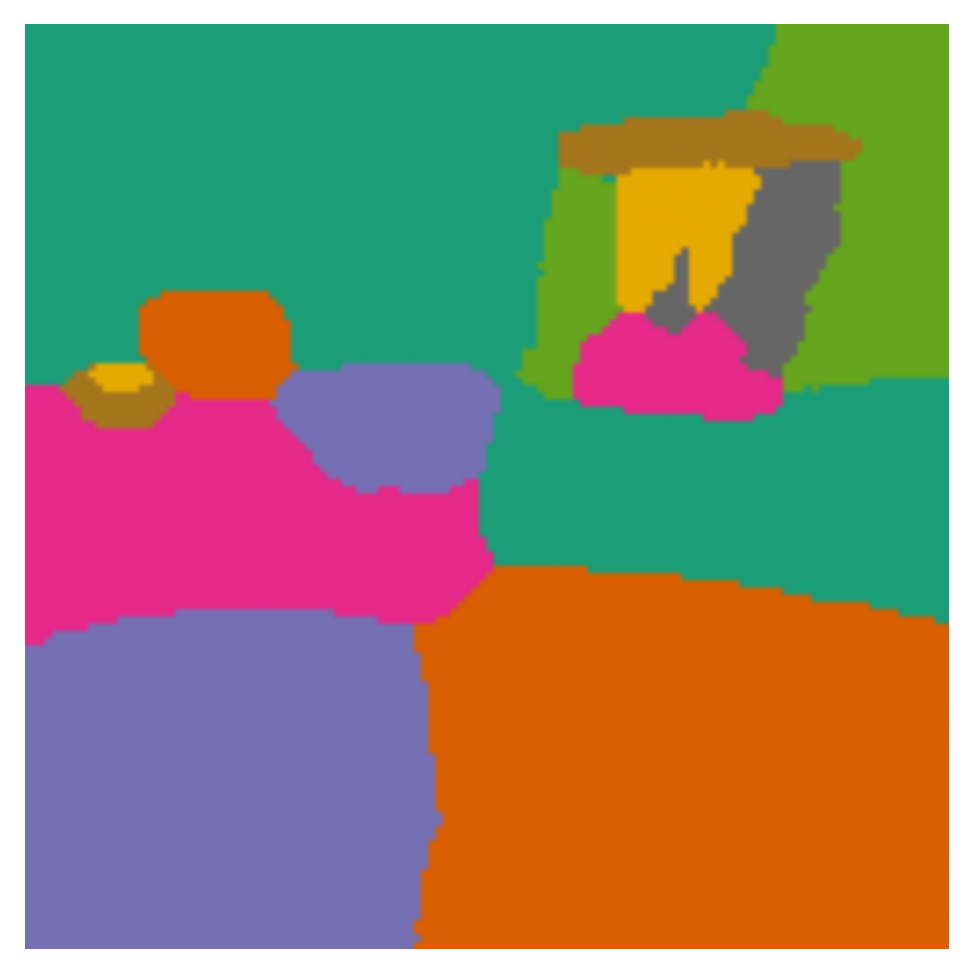}
    \caption{Results of the implicit Slot Attention \citep{locatelloObjectCentricLearningSlot2020a,changObjectRepresentationsFixed2022a} on our simulated multi-object scenes. From left to right: pair of input images, pair of reconstructed images, pair of segmentation masks. More examples are provided in \cref{fig:slot_more}. Overall, the quality of reconstruction and segmentation remains limited.}
	\label{fig:slot}
\end{figure}

\update{
\cref{tab:thor} shows the results of each design choice on the ID and OOD test sets.
Corroborating with the results from the oracle object-centric approximation, exploiting the latent slot structure leads to significantly higher accuracy for downstream reasoning.
In particular, the intervention model built with the slot matching scheme is over $34\%$ more accurate than the vanilla counterpart built with the global average-pooling operation.
Nevertheless, the overall performance from the frozen encoder remains limited, as evidenced by the inferior results to the fine-tuned ResNet model as well as the blurry image reconstructure shown in \cref{fig:slot}, indicating a large room for improvement in object-centric learning on our benchmark.
}

\begin{table}[t]
\small
\centering
\begin{minipage}[t]{0.485\linewidth} 
    \centering
    \begin{tabular}{ccc}
    \toprule
                            & ID                & OOD                         \\ \midrule
    resnet                  & $ 0.83 \pm 0.01 $ & $ 0.30 \pm 0.08 $         \\  \midrule
    oracle-mask             & $ \bm{0.90} \pm 0.01 $ & $ \bm{0.42} \pm 0.06 $     \\  \midrule
    slot-avg               & $ 0.49 \pm 0.01 $ & $ 0.15 \pm 0.01 $         \\  \midrule
    slot-dense             & $ 0.51 \pm 0.01 $ & $ 0.19 \pm 0.03 $         \\  \midrule
    slot-match             & $ 0.66 \pm 0.01 $ & $ 0.21 \pm 0.01 $    \\  \midrule
    \end{tabular}
    \caption{Intervention classification accuracy in simulated multi-object scenes. The pre-trained slot-based encoder is frozen in experiments.}
    \label{tab:thor}
\end{minipage}
\quad
\begin{minipage}[t]{0.485\linewidth}
    \centering
    \begin{tabular}{ccc}
    \toprule
                            & ID              & OOD                \\ \midrule
    resnet                  & $0.42 \pm 0.03$ & $0.17 \pm 0.03$    \\  \midrule
    clip                    & $0.45 \pm 0.02$ & $0.24 \pm 0.02$    \\  \midrule
    group-avg               & $0.47 \pm 0.03$ & $0.24 \pm 0.03$    \\  \midrule
    group-dense             & $0.50 \pm 0.04$ & $0.26 \pm 0.03$    \\  \midrule
    group-token             & $\bm{0.52} \pm 0.03$ & $\bm{0.27} \pm 0.03$    \\  \midrule
    \end{tabular}
    \caption{Intervention classification accuracy in real-world multi-object scenes. All encoders are pre-trained, and kept frozen in experiments.}
    \label{tab:epic}
\end{minipage}
\end{table}  

\subsubsection{Real-world Multi-object Scenes}
\label{subsec:epic}

\paragraph{Setup.}
\update{
We finally migrate to intervention modeling on real-world observations repurposed from the Epic-Kitchens dataset~\citep{damenRescalingEgocentricVision2022}.
Reasoning about abstract transformations between real-world image pairs is generally more challenging than in the controlled simulation environment, due to significant global changes in camera locations and perspectives, frequent local occlusions by arms and hands, as well as limited training data.
Given these challenges, we turn our attention to GroupViT~\citep{xuGroupViTSemanticSegmentation2022a}, a set-structured visual representation pre-trained on massive amounts of internet data.
}
Similar to \cref{subsec:thor}, we assess the impact of visual representations by comparing three pre-trained encoders: (i) ResNet-18 pre-trained on ImageNet~\citep{dengImageNetLargescaleHierarchical2009}, (ii) CLIP, and (iii) GroupViT pre-trained on internet data.
Furthermore, to leverage the latent structure in GroupViT, we match the slots between two images based on the group tokens, and aggregate the resulting action embeddings with two choices: average-pooling (denoted by {\em token-mean}) and max-pooling (denoted by {\em token-max}).

\begin{figure}[t]
    \vspace{-10pt}
    \centering
    \small
    \subfigure[ID]{
        \includegraphics[width=0.4\textwidth]{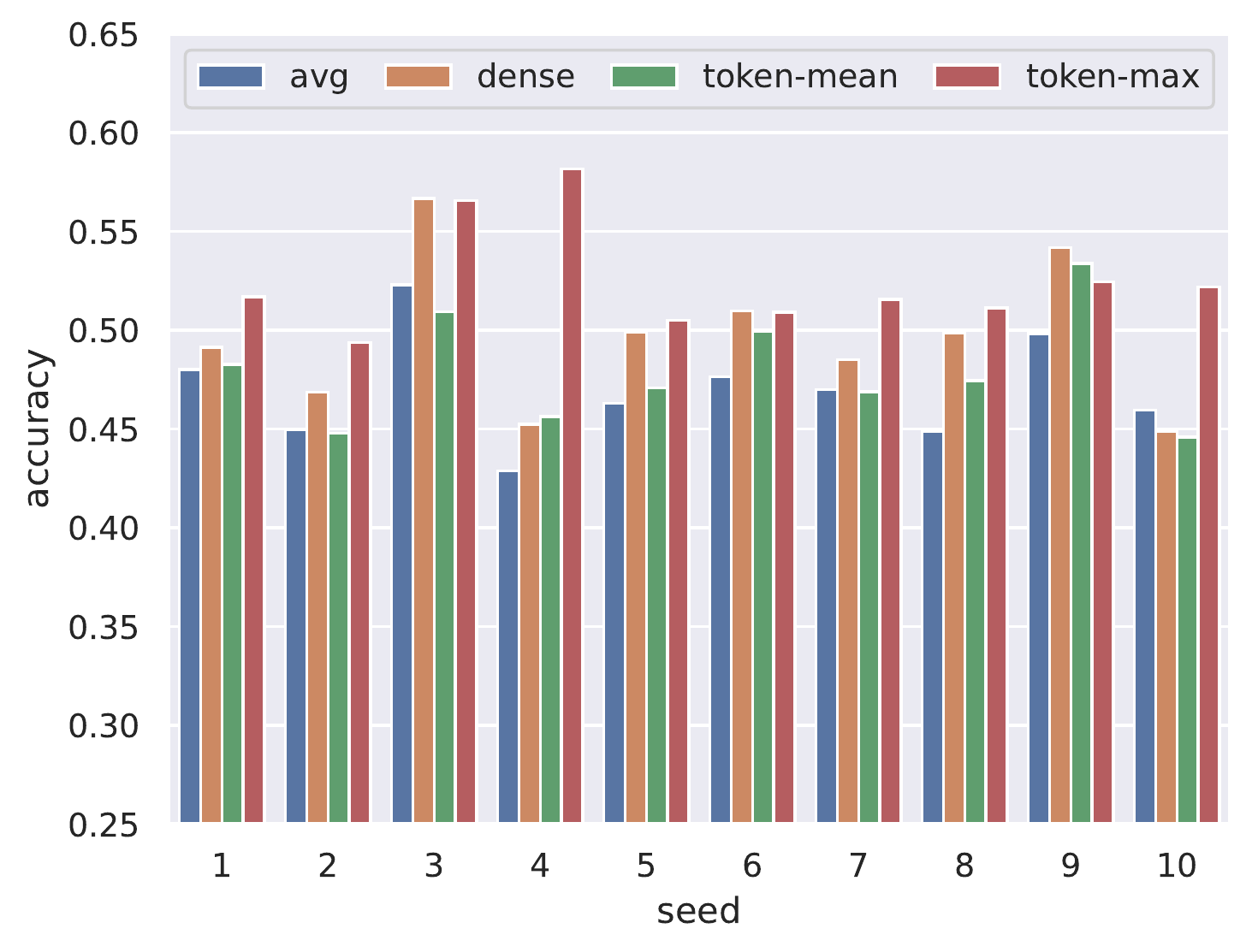}\label{fig:iid_epic}
    }
    \quad
    \subfigure[OOD]{
        \includegraphics[width=0.4\textwidth]{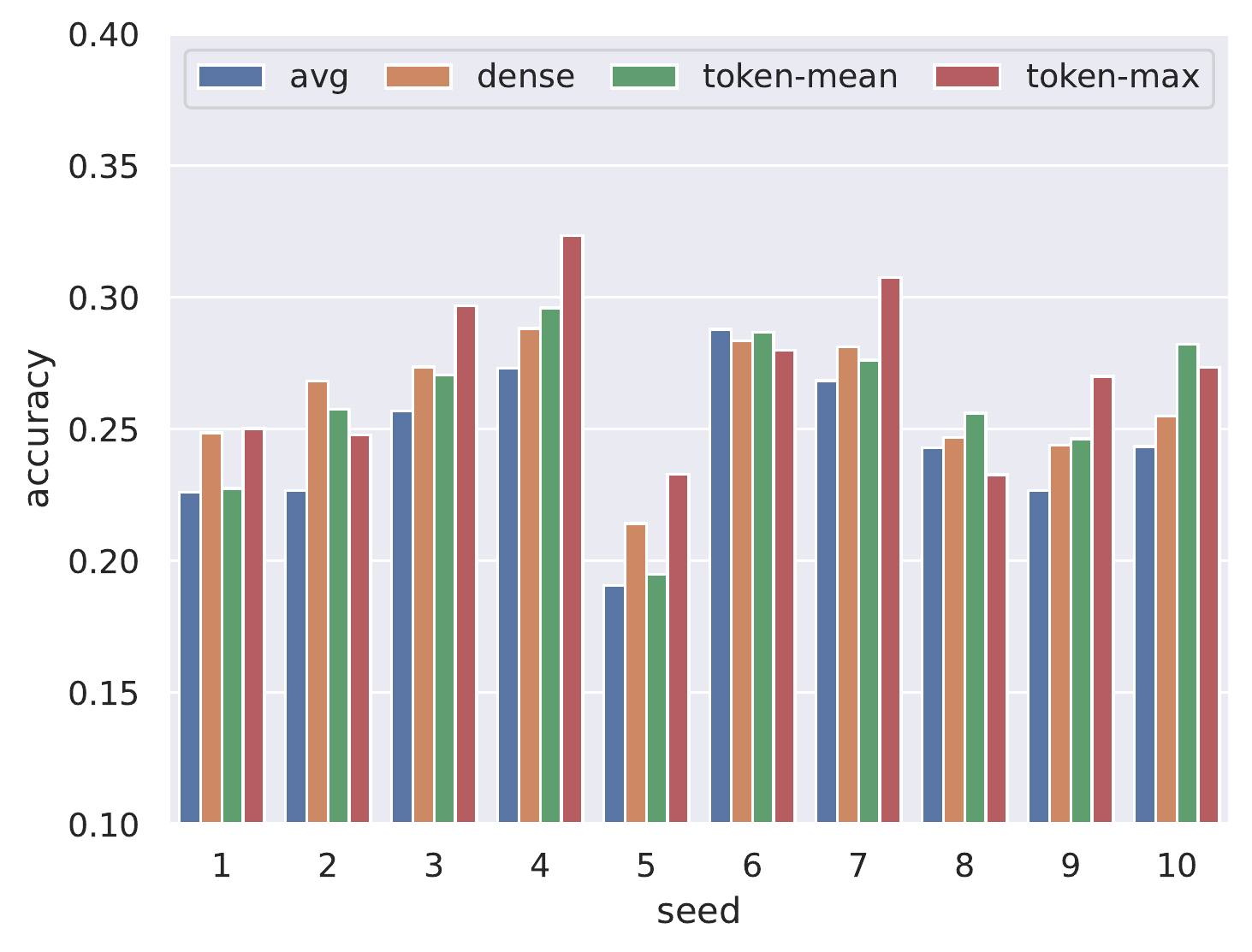}\label{fig:ood_epic}
    }
    \caption{
    Results of different group matching modules in real-world data on 10 different seeds (random splits). 
    Compared with the {\em group-avg} baseline, exploiting the learned group structure with {\em token-max} leads to improved accuracy on both the ID and OOD sets that differ in object classes.
    }
    \label{fig:epic}
\end{figure}

\paragraph{Results.}
\cref{tab:epic} shows the results of different models on the ID and OOD test sets.
While ResNet, CLIP, and {\em group-avg} all fall into the class of distributed vector representations, the intervention models built with the latter two generalize much better than the first one to unseen objects, suggesting the strength of large-scale pre-training for OOD generalization.
Among all considered models, the token-based group matching appears to be most effective for intervention modeling, outperforming the ResNet baseline on the ID and OOD test sets by $\sim24\%$ and $\sim59\%$, respectively.
In particular, we observe in \cref{fig:epic} that {\em token-max} results in better accuracy than {\em token-mean} on most seeds.
Intuitively, when semantic regions are properly matched between two images, attending to the group of the most significant changes using max-pooling can be a good inductive bias to reason about the action class.





\section{Conclusion and Discussion}
\label{sec:conclusion}
In this paper, we introduced \causaltriplet, a causal representation learning benchmark with high visual complexities, actionable counterfactuals, and an interventional downstream task.
We revisited the principles of independent causal mechanisms and sparse mechanism shifts in the context of intervention modeling, and empirically estimated their strengths for out-of-distribution robustness by making use of oracle knowledge of the underlying structures.
However, we also observed significant challenges for recent methods to discover these structures, indicating fruitful opportunities for future research.

As the first benchmark of its kind, {\tt Causal} {\tt Triplet} is still subject to two major constraints: the object states in our simulated dataset are binary variables, as opposed to continuous ones in the real world; and the scale of our repurposed real-world dataset is limited, sufficient for transfer learning but not for representation learning from scratch. 
Nevertheless, we hope our benchmark will call attention to the assumptions made in causal representation learning, serve as a public testbed in challenging yet practical settings, and propel progress towards real-world problems.

\acks{
This work is supported in part by the Swiss National Science Foundation under the Grant 2OOO21-L92326.
We thank Maximilian Seitzer, Carl-Johann Simon-Gabriel, Andrii Zadaianchuk, Yuchen Zhu, Harvineet Singh and Milton Montero for helpful discussions. We thank Parth Kothari, Riccardo Cadei, Linyan Yang and Yifan Sun for thoughtful feedback on early drafts, as well as anonymous reviewers for insightful comments.
}


\bibliography{triplet}

\begin{thebibliography}{51}
\providecommand{\natexlab}[1]{#1}
\providecommand{\url}[1]{\texttt{#1}}
\expandafter\ifx\csname urlstyle\endcsname\relax
  \providecommand{\doi}[1]{doi: #1}\else
  \providecommand{\doi}{doi: \begingroup \urlstyle{rm}\Url}\fi

\bibitem[Ahmed et~al.(2021)Ahmed, Tr{\"a}uble, Goyal, Neitz, Wuthrich, Bengio,
  Sch{\"o}lkopf, and Bauer]{ahmedCausalWorldRoboticManipulation2021}
Ossama Ahmed, Frederik Tr{\"a}uble, Anirudh Goyal, Alexander Neitz, Manuel
  Wuthrich, Yoshua Bengio, Bernhard Sch{\"o}lkopf, and Stefan Bauer.
\newblock {{CausalWorld}}: {{A Robotic Manipulation Benchmark}} for {{Causal
  Structure}} and {{Transfer Learning}}.
\newblock In \emph{International {{Conference}} on {{Learning
  Representations}}}, 2021.

\bibitem[Ahuja et~al.(2022{\natexlab{a}})Ahuja, Hartford, and
  Bengio]{ahujaWeaklySupervisedRepresentation2022}
Kartik Ahuja, Jason Hartford, and Yoshua Bengio.
\newblock Weakly {{Supervised Representation Learning}} with {{Sparse
  Perturbations}}, June 2022{\natexlab{a}}.

\bibitem[Ahuja et~al.(2022{\natexlab{b}})Ahuja, Wang, Mahajan, and
  Bengio]{ahujaInterventionalCausalRepresentation2022}
Kartik Ahuja, Yixin Wang, Divyat Mahajan, and Yoshua Bengio.
\newblock Interventional {{Causal Representation Learning}}, October
  2022{\natexlab{b}}.

\bibitem[Atzmon et~al.(2020)Atzmon, Kreuk, Shalit, and
  Chechik]{atzmonCausalViewCompositional2020}
Yuval Atzmon, Felix Kreuk, Uri Shalit, and Gal Chechik.
\newblock A causal view of compositional zero-shot recognition.
\newblock In \emph{Advances in {{Neural Information Processing Systems}}},
  volume~33, pages 1462--1473, 2020.

\bibitem[Brehmer et~al.(2022)Brehmer, {de Haan}, Lippe, and
  Cohen]{brehmerWeaklySupervisedCausal2022}
Johann Brehmer, Pim {de Haan}, Phillip Lippe, and Taco Cohen.
\newblock Weakly supervised causal representation learning.
\newblock In \emph{Advances in {{Neural Information Processing Systems}}},
  2022.

\bibitem[Burgess et~al.(2019)Burgess, Matthey, Watters, Kabra, Higgins,
  Botvinick, and Lerchner]{burgessMONetUnsupervisedScene2019}
Christopher~P. Burgess, Loic Matthey, Nicholas Watters, Rishabh Kabra, Irina
  Higgins, Matt Botvinick, and Alexander Lerchner.
\newblock {{MONet}}: {{Unsupervised Scene Decomposition}} and
  {{Representation}}, January 2019.

\bibitem[Chang et~al.(2022)Chang, Griffiths, and
  Levine]{changObjectRepresentationsFixed2022a}
Michael Chang, Thomas~L. Griffiths, and Sergey Levine.
\newblock Object {{Representations}} as {{Fixed Points}}: {{Training Iterative
  Refinement Algorithms}} with {{Implicit Differentiation}}.
\newblock In \emph{Advances in {{Neural Information Processing Systems}}},
  October 2022.

\bibitem[Chen et~al.(2016)Chen, Duan, Houthooft, Schulman, Sutskever, and
  Abbeel]{chenInfoGANInterpretableRepresentation2016}
Xi~Chen, Yan Duan, Rein Houthooft, John Schulman, Ilya Sutskever, and Pieter
  Abbeel.
\newblock {{InfoGAN}}: {{Interpretable Representation Learning}} by
  {{Information Maximizing Generative Adversarial Nets}}.
\newblock In D.~D. Lee, M.~Sugiyama, U.~V. Luxburg, I.~Guyon, and R.~Garnett,
  editors, \emph{Advances in {{Neural Information Processing Systems}} 29},
  pages 2172--2180, 2016.

\bibitem[Cohen(2022)]{cohenGroundedTheoryCausation2022}
Taco Cohen.
\newblock Towards a {{Grounded Theory}} of {{Causation}} for {{Embodied AI}},
  June 2022.

\bibitem[Damen et~al.(2022)Damen, Doughty, Farinella, Furnari, Kazakos, Ma,
  Moltisanti, Munro, Perrett, Price, and
  Wray]{damenRescalingEgocentricVision2022}
Dima Damen, Hazel Doughty, Giovanni~Maria Farinella, Antonino Furnari,
  Evangelos Kazakos, Jian Ma, Davide Moltisanti, Jonathan Munro, Toby Perrett,
  Will Price, and Michael Wray.
\newblock Rescaling {{Egocentric Vision}}: {{Collection}}, {{Pipeline}} and
  {{Challenges}} for {{EPIC-KITCHENS-100}}.
\newblock \emph{International Journal of Computer Vision}, 130:\penalty0
  33--55, January 2022.

\bibitem[Deitke et~al.(2022{\natexlab{a}})Deitke, Batra, Bisk, Campari, Chang,
  Chaplot, Chen, D'Arpino, Ehsani, Farhadi, {Fei-Fei}, Francis, Gan, Grauman,
  Hall, Han, Jain, Kembhavi, Krantz, Lee, Li, Majumder, Maksymets,
  {Mart{\'i}n-Mart{\'i}n}, Mottaghi, Raychaudhuri, Roberts, Savarese, Savva,
  Shridhar, S{\"u}nderhauf, Szot, Talbot, Tenenbaum, Thomason, Toshev, Truong,
  Weihs, and Wu]{deitkeRetrospectivesEmbodiedAI2022}
Matt Deitke, Dhruv Batra, Yonatan Bisk, Tommaso Campari, Angel~X. Chang,
  Devendra~Singh Chaplot, Changan Chen, Claudia~P{\'e}rez D'Arpino, Kiana
  Ehsani, Ali Farhadi, Li~{Fei-Fei}, Anthony Francis, Chuang Gan, Kristen
  Grauman, David Hall, Winson Han, Unnat Jain, Aniruddha Kembhavi, Jacob
  Krantz, Stefan Lee, Chengshu Li, Sagnik Majumder, Oleksandr Maksymets,
  Roberto {Mart{\'i}n-Mart{\'i}n}, Roozbeh Mottaghi, Sonia Raychaudhuri, Mike
  Roberts, Silvio Savarese, Manolis Savva, Mohit Shridhar, Niko S{\"u}nderhauf,
  Andrew Szot, Ben Talbot, Joshua~B. Tenenbaum, Jesse Thomason, Alexander
  Toshev, Joanne Truong, Luca Weihs, and Jiajun Wu.
\newblock Retrospectives on the {{Embodied AI Workshop}}, October
  2022{\natexlab{a}}.

\bibitem[Deitke et~al.(2022{\natexlab{b}})Deitke, VanderBilt, Herrasti, Weihs,
  Salvador, Ehsani, Han, Kolve, Farhadi, Kembhavi, and
  Mottaghi]{deitkeProcTHORLargeScaleEmbodied2022}
Matt Deitke, Eli VanderBilt, Alvaro Herrasti, Luca Weihs, Jordi Salvador, Kiana
  Ehsani, Winson Han, Eric Kolve, Ali Farhadi, Aniruddha Kembhavi, and Roozbeh
  Mottaghi.
\newblock {{ProcTHOR}}: {{Large-Scale Embodied AI Using Procedural
  Generation}}.
\newblock In \emph{Advances in {{Neural Information Processing Systems}}}, June
  2022{\natexlab{b}}.

\bibitem[Deng et~al.(2009)Deng, Dong, Socher, Li, and
  {and}]{dengImageNetLargescaleHierarchical2009}
J.~Deng, W.~Dong, R.~Socher, L.~Li, and {and}.
\newblock {{ImageNet}}: {{A}} large-scale hierarchical image database.
\newblock In \emph{2009 {{IEEE Conference}} on {{Computer Vision}} and
  {{Pattern Recognition}}}, pages 248--255, June 2009.

\bibitem[Dittadi et~al.(2020)Dittadi, Tr{\"a}uble, Locatello, Wuthrich,
  Agrawal, Winther, Bauer, and
  Sch{\"o}lkopf]{dittadiTransferDisentangledRepresentations2020}
Andrea Dittadi, Frederik Tr{\"a}uble, Francesco Locatello, Manuel Wuthrich,
  Vaibhav Agrawal, Ole Winther, Stefan Bauer, and Bernhard Sch{\"o}lkopf.
\newblock On the {{Transfer}} of {{Disentangled Representations}} in
  {{Realistic Settings}}.
\newblock In \emph{International {{Conference}} on {{Learning
  Representations}}}, September 2020.

\bibitem[Dittadi et~al.(2022)Dittadi, Papa, Vita, Sch{\"o}lkopf, Winther, and
  Locatello]{dittadiGeneralizationRobustnessImplications2022a}
Andrea Dittadi, Samuele~S. Papa, Michele~De Vita, Bernhard Sch{\"o}lkopf, Ole
  Winther, and Francesco Locatello.
\newblock Generalization and {{Robustness Implications}} in {{Object-Centric
  Learning}}.
\newblock In \emph{Proceedings of the 39th {{International Conference}} on
  {{Machine Learning}}}, pages 5221--5285, June 2022.

\bibitem[Fathi et~al.(2011)Fathi, Farhadi, and
  Rehg]{fathiUnderstandingEgocentricActivities2011}
Alireza Fathi, Ali Farhadi, and James~M. Rehg.
\newblock Understanding egocentric activities.
\newblock In \emph{2011 {{International Conference}} on {{Computer Vision}}},
  pages 407--414, November 2011.

\bibitem[Goyal et~al.(2021)Goyal, Lamb, Hoffmann, Sodhani, Levine, Bengio, and
  Sch{\"o}lkopf]{goyalRecurrentIndependentMechanisms2021}
Anirudh Goyal, Alex Lamb, Jordan Hoffmann, Shagun Sodhani, Sergey Levine,
  Yoshua Bengio, and Bernhard Sch{\"o}lkopf.
\newblock Recurrent {{Independent Mechanisms}}.
\newblock In \emph{International {{Conference}} on {{Learning
  Representations}}}, 2021.

\bibitem[Greff et~al.(2017)Greff, {van Steenkiste}, and
  Schmidhuber]{greffNeuralExpectationMaximization2017}
Klaus Greff, Sjoerd {van Steenkiste}, and J{\"u}rgen Schmidhuber.
\newblock Neural {{Expectation Maximization}}.
\newblock In \emph{Advances in {{Neural Information Processing Systems}}},
  volume~30, 2017.

\bibitem[Greff et~al.(2019)Greff, Kaufman, Kabra, Watters, Burgess, Zoran,
  Matthey, Botvinick, and Lerchner]{greffMultiObjectRepresentationLearning2019}
Klaus Greff, Rapha{\"e}l~Lopez Kaufman, Rishabh Kabra, Nick Watters,
  Christopher Burgess, Daniel Zoran, Loic Matthey, Matthew Botvinick, and
  Alexander Lerchner.
\newblock Multi-{{Object Representation Learning}} with {{Iterative Variational
  Inference}}.
\newblock In \emph{Proceedings of the 36th {{International Conference}} on
  {{Machine Learning}}}, pages 2424--2433, May 2019.

\bibitem[Ha and Schmidhuber(2018)]{haWorldModels2018}
David Ha and J{\"u}rgen Schmidhuber.
\newblock World {{Models}}.
\newblock \emph{arXiv:1803.10122 [cs, stat]}, March 2018.

\bibitem[He et~al.(2016)He, Zhang, Ren, and Sun]{heDeepResidualLearning2016}
Kaiming He, Xiangyu Zhang, Shaoqing Ren, and Jian Sun.
\newblock Deep {{Residual Learning}} for {{Image Recognition}}.
\newblock In \emph{Proceedings of the {{IEEE Conference}} on {{Computer
  Vision}} and {{Pattern Recognition}}}, pages 770--778, 2016.

\bibitem[Higgins et~al.(2017)Higgins, Matthey, Pal, Burgess, Glorot, Botvinick,
  Mohamed, and Lerchner]{higginsBetaVAELearningBasic2017}
Irina Higgins, Loic Matthey, Arka Pal, Christopher Burgess, Xavier Glorot,
  Matthew Botvinick, Shakir Mohamed, and Alexander Lerchner.
\newblock Beta-{{VAE}}: {{Learning Basic Visual Concepts}} with a {{Constrained
  Variational Framework}}.
\newblock In \emph{International {{Conference}} on {{Learning
  Representations}}}, 2017.

\bibitem[Holland(1986)]{hollandStatisticsCausalInference1986}
Paul~W. Holland.
\newblock Statistics and {{Causal Inference}}.
\newblock \emph{Journal of the American Statistical Association}, 81:\penalty0
  945--960, 1986.

\bibitem[Klindt et~al.(2021)Klindt, Schott, Sharma, Ustyuzhaninov, Brendel,
  Bethge, and Paiton]{klindtNonlinearDisentanglementNatural2021}
David~A. Klindt, Lukas Schott, Yash Sharma, Ivan Ustyuzhaninov, Wieland
  Brendel, Matthias Bethge, and Dylan Paiton.
\newblock Towards {{Nonlinear Disentanglement}} in {{Natural Data}} with
  {{Temporal Sparse Coding}}.
\newblock In \emph{International {{Conference}} on {{Learning
  Representations}}}, 2021.

\bibitem[Lachapelle et~al.(2022)Lachapelle, Rodriguez, Sharma, Everett, Priol,
  Lacoste, and
  {Lacoste-Julien}]{lachapelleDisentanglementMechanismSparsity2022}
Sebastien Lachapelle, Pau Rodriguez, Yash Sharma, Katie~E. Everett, R{\'e}mi~LE
  Priol, Alexandre Lacoste, and Simon {Lacoste-Julien}.
\newblock Disentanglement via {{Mechanism Sparsity Regularization}}: {{A New
  Principle}} for {{Nonlinear ICA}}.
\newblock In \emph{Proceedings of the {{First Conference}} on {{Causal
  Learning}} and {{Reasoning}}}, pages 428--484, June 2022.

\bibitem[Lei et~al.(2022)Lei, Sch{\"o}lkopf, and
  Posner]{leiVariationalCausalDynamics2022}
Anson Lei, Bernhard Sch{\"o}lkopf, and Ingmar Posner.
\newblock Variational {{Causal Dynamics}}: {{Discovering Modular World Models}}
  from {{Interventions}}, June 2022.

\bibitem[Lippe et~al.(2022{\natexlab{a}})Lippe, Magliacane, L{\"o}we, Asano,
  Cohen, and Gavves]{lippeICITRISCausalRepresentation2022}
Phillip Lippe, Sara Magliacane, Sindy L{\"o}we, Yuki~M. Asano, Taco Cohen, and
  Efstratios Gavves.
\newblock {{iCITRIS}}: {{Causal Representation Learning}} for {{Instantaneous
  Temporal Effects}}.
\newblock In \emph{{{UAI}} 2022 {{Workshop}} on {{Causal Representation
  Learning}}}, July 2022{\natexlab{a}}.

\bibitem[Lippe et~al.(2022{\natexlab{b}})Lippe, Magliacane, L{\"o}we, Asano,
  Cohen, and Gavves]{lippeInterventionDesignCausal2022}
Phillip Lippe, Sara Magliacane, Sindy L{\"o}we, Yuki~M. Asano, Taco Cohen, and
  Efstratios Gavves.
\newblock Intervention {{Design}} for {{Causal Representation Learning}}.
\newblock In \emph{{{UAI}} 2022 {{Workshop}} on {{Causal Representation
  Learning}}}, July 2022{\natexlab{b}}.

\bibitem[Lippe et~al.(2022{\natexlab{c}})Lippe, Magliacane, L{\"o}we, Asano,
  Cohen, and Gavves]{lippeCITRISCausalIdentifiability2022}
Phillip Lippe, Sara Magliacane, Sindy L{\"o}we, Yuki~M. Asano, Taco Cohen, and
  Stratis Gavves.
\newblock {{CITRIS}}: {{Causal Identifiability}} from {{Temporal Intervened
  Sequences}}.
\newblock In \emph{Proceedings of the 39th {{International Conference}} on
  {{Machine Learning}}}, pages 13557--13603, June 2022{\natexlab{c}}.

\bibitem[Liu et~al.(2022)Liu, Cadei, Schweizer, Bahmani, and
  Alahi]{liuRobustAdaptiveMotion2022a}
Yuejiang Liu, Riccardo Cadei, Jonas Schweizer, Sherwin Bahmani, and Alexandre
  Alahi.
\newblock Towards {{Robust}} and {{Adaptive Motion Forecasting}}: {{A Causal
  Representation Perspective}}.
\newblock In \emph{Proceedings of the {{IEEE}}/{{CVF Conference}} on {{Computer
  Vision}} and {{Pattern Recognition}}}, pages 17081--17092, 2022.

\bibitem[Locatello et~al.(2019)Locatello, Bauer, Lucic, Raetsch, Gelly,
  Sch{\"o}lkopf, and Bachem]{locatelloChallengingCommonAssumptions2019}
Francesco Locatello, Stefan Bauer, Mario Lucic, Gunnar Raetsch, Sylvain Gelly,
  Bernhard Sch{\"o}lkopf, and Olivier Bachem.
\newblock Challenging {{Common Assumptions}} in the {{Unsupervised Learning}}
  of {{Disentangled Representations}}.
\newblock In \emph{Proceedings of the 36th {{International Conference}} on
  {{Machine Learning}}}, pages 4114--4124, May 2019.

\bibitem[Locatello et~al.(2020{\natexlab{a}})Locatello, Poole, Raetsch,
  Sch{\"o}lkopf, Bachem, and
  Tschannen]{locatelloWeaklySupervisedDisentanglementCompromises2020a}
Francesco Locatello, Ben Poole, Gunnar Raetsch, Bernhard Sch{\"o}lkopf, Olivier
  Bachem, and Michael Tschannen.
\newblock Weakly-{{Supervised Disentanglement Without Compromises}}.
\newblock In \emph{Proceedings of the 37th {{International Conference}} on
  {{Machine Learning}}}, pages 6348--6359, November 2020{\natexlab{a}}.

\bibitem[Locatello et~al.(2020{\natexlab{b}})Locatello, Weissenborn,
  Unterthiner, Mahendran, Heigold, Uszkoreit, Dosovitskiy, and
  Kipf]{locatelloObjectCentricLearningSlot2020a}
Francesco Locatello, Dirk Weissenborn, Thomas Unterthiner, Aravindh Mahendran,
  Georg Heigold, Jakob Uszkoreit, Alexey Dosovitskiy, and Thomas Kipf.
\newblock Object-{{Centric Learning}} with {{Slot Attention}}.
\newblock In \emph{Advances in {{Neural Information Processing Systems}}},
  volume~33, pages 11525--11538, 2020{\natexlab{b}}.

\bibitem[Montero et~al.(2021)Montero, Ludwig, Costa, Malhotra, and
  Bowers]{monteroRoleDisentanglementGeneralisation2021}
Milton~Llera Montero, Casimir~JH Ludwig, Rui~Ponte Costa, Gaurav Malhotra, and
  Jeffrey Bowers.
\newblock The role of {{Disentanglement}} in {{Generalisation}}.
\newblock In \emph{International {{Conference}} on {{Learning
  Representations}}}, 2021.

\bibitem[Parascandolo et~al.(2018)Parascandolo, Kilbertus, {Rojas-Carulla}, and
  Sch{\"o}lkopf]{parascandoloLearningIndependentCausal2018a}
Giambattista Parascandolo, Niki Kilbertus, Mateo {Rojas-Carulla}, and Bernhard
  Sch{\"o}lkopf.
\newblock Learning {{Independent Causal Mechanisms}}.
\newblock In \emph{International {{Conference}} on {{Machine Learning}}}, pages
  4036--4044, July 2018.

\bibitem[Peters et~al.(2017)Peters, Janzing, and
  Sch{\"o}lkopf]{petersElementsCausalInference2017}
Jonas Peters, Dominik Janzing, and Bernhard Sch{\"o}lkopf.
\newblock \emph{Elements of {{Causal Inference}}: {{Foundations}} and
  {{Learning Algorithms}}}.
\newblock Adaptive {{Computation}} and {{Machine Learning}} Series. {Cambridge,
  MA, USA}, November 2017.
\newblock ISBN 978-0-262-03731-0.

\bibitem[Radford et~al.(2021)Radford, Kim, Hallacy, Ramesh, Goh, Agarwal,
  Sastry, Askell, Mishkin, Clark, Krueger, and
  Sutskever]{radfordLearningTransferableVisual2021a}
Alec Radford, Jong~Wook Kim, Chris Hallacy, Aditya Ramesh, Gabriel Goh,
  Sandhini Agarwal, Girish Sastry, Amanda Askell, Pamela Mishkin, Jack Clark,
  Gretchen Krueger, and Ilya Sutskever.
\newblock Learning {{Transferable Visual Models From Natural Language
  Supervision}}.
\newblock In \emph{Proceedings of the 38th {{International Conference}} on
  {{Machine Learning}}}, pages 8748--8763, July 2021.

\bibitem[Rolinek et~al.(2019)Rolinek, Zietlow, and
  Martius]{rolinekVariationalAutoencodersPursue2019}
Michal Rolinek, Dominik Zietlow, and Georg Martius.
\newblock Variational {{Autoencoders Pursue PCA Directions}} (by {{Accident}}).
\newblock In \emph{Proceedings of the {{IEEE}}/{{CVF Conference}} on {{Computer
  Vision}} and {{Pattern Recognition}}}, pages 12406--12415, 2019.

\bibitem[Ruis et~al.(2021)Ruis, Burghouts, and
  Bucur]{ruisIndependentPrototypePropagation2021}
Frank Ruis, Gertjan Burghouts, and Doina Bucur.
\newblock Independent {{Prototype Propagation}} for {{Zero-Shot
  Compositionality}}.
\newblock In \emph{Advances in {{Neural Information Processing Systems}}},
  volume~34, pages 10641--10653, 2021.

\bibitem[Sch{\"o}lkopf et~al.(2012)Sch{\"o}lkopf, Janzing, Peters, Sgouritsa,
  Zhang, and Mooij]{scholkopfCausalAnticausalLearning2012}
Bernhard Sch{\"o}lkopf, Dominik Janzing, Jonas Peters, Eleni Sgouritsa, Kun
  Zhang, and Joris Mooij.
\newblock On causal and anticausal learning.
\newblock In \emph{Proceedings of the 29th {{International Coference}} on
  {{International Conference}} on {{Machine Learning}}}, {{ICML}}'12, pages
  459--466, {Madison, WI, USA}, June 2012.
\newblock ISBN 978-1-4503-1285-1.

\bibitem[Sch{\"o}lkopf et~al.(2021)Sch{\"o}lkopf, Locatello, Bauer, Ke,
  Kalchbrenner, Goyal, and Bengio]{scholkopfCausalRepresentationLearning2021a}
Bernhard Sch{\"o}lkopf, Francesco Locatello, Stefan Bauer, Nan~Rosemary Ke, Nal
  Kalchbrenner, Anirudh Goyal, and Yoshua Bengio.
\newblock Towards {{Causal Representation Learning}}, February 2021.

\bibitem[Seitzer et~al.(2022)Seitzer, Horn, Zadaianchuk, Zietlow, Xiao,
  {Simon-Gabriel}, He, Zhang, Sch{\"o}lkopf, Brox, and
  Locatello]{seitzerBridgingGapRealWorld2022a}
Maximilian Seitzer, Max Horn, Andrii Zadaianchuk, Dominik Zietlow, Tianjun
  Xiao, Carl-Johann {Simon-Gabriel}, Tong He, Zheng Zhang, Bernhard
  Sch{\"o}lkopf, Thomas Brox, and Francesco Locatello.
\newblock Bridging the {{Gap}} to {{Real-World Object-Centric Learning}},
  September 2022.

\bibitem[Singh et~al.(2022)Singh, Wu, and
  Ahn]{singhSimpleUnsupervisedObjectCentric2022}
Gautam Singh, Yi-Fu Wu, and Sungjin Ahn.
\newblock Simple {{Unsupervised Object-Centric Learning}} for {{Complex}} and
  {{Naturalistic Videos}}.
\newblock In \emph{Advances in {{Neural Information Processing Systems}}},
  October 2022.

\bibitem[Tr{\"a}uble et~al.(2021)Tr{\"a}uble, Creager, Kilbertus, Locatello,
  Dittadi, Goyal, Sch{\"o}lkopf, and
  Bauer]{traubleDisentangledRepresentationsLearned2021}
Frederik Tr{\"a}uble, Elliot Creager, Niki Kilbertus, Francesco Locatello,
  Andrea Dittadi, Anirudh Goyal, Bernhard Sch{\"o}lkopf, and Stefan Bauer.
\newblock On {{Disentangled Representations Learned}} from {{Correlated Data}}.
\newblock In \emph{Proceedings of the 38th {{International Conference}} on
  {{Machine Learning}}}, pages 10401--10412, July 2021.

\bibitem[{van Steenkiste} et~al.(2019){van Steenkiste}, Locatello, Schmidhuber,
  and Bachem]{vansteenkisteAreDisentangledRepresentations2019}
Sjoerd {van Steenkiste}, Francesco Locatello, J{\"u}rgen Schmidhuber, and
  Olivier Bachem.
\newblock Are {{Disentangled Representations Helpful}} for {{Abstract Visual
  Reasoning}}?
\newblock In \emph{Advances in {{Neural Information Processing Systems}}},
  volume~32, 2019.

\bibitem[{von K{\"u}gelgen} et~al.(2021){von K{\"u}gelgen}, Sharma, Gresele,
  Brendel, Sch{\"o}lkopf, Besserve, and
  Locatello]{vonkugelgenSelfSupervisedLearningData2021}
Julius {von K{\"u}gelgen}, Yash Sharma, Luigi Gresele, Wieland Brendel,
  Bernhard Sch{\"o}lkopf, Michel Besserve, and Francesco Locatello.
\newblock Self-{{Supervised Learning}} with {{Data Augmentations Provably
  Isolates Content}} from {{Style}}.
\newblock In \emph{Advances in {{Neural Information Processing Systems}}},
  volume~34, pages 16451--16467, 2021.

\bibitem[Xu et~al.(2022)Xu, De~Mello, Liu, Byeon, Breuel, Kautz, and
  Wang]{xuGroupViTSemanticSegmentation2022a}
Jiarui Xu, Shalini De~Mello, Sifei Liu, Wonmin Byeon, Thomas Breuel, Jan Kautz,
  and Xiaolong Wang.
\newblock {{GroupViT}}: {{Semantic Segmentation Emerges From Text
  Supervision}}.
\newblock In \emph{Proceedings of the {{IEEE}}/{{CVF Conference}} on {{Computer
  Vision}} and {{Pattern Recognition}}}, pages 18134--18144, 2022.

\bibitem[Yang et~al.(2020)Yang, Chen, and
  Soatto]{yangLearningManipulateIndividual2020}
Yanchao Yang, Yutong Chen, and Stefano Soatto.
\newblock Learning to {{Manipulate Individual Objects}} in an {{Image}}.
\newblock In \emph{Proceedings of the {{IEEE}}/{{CVF Conference}} on {{Computer
  Vision}} and {{Pattern Recognition}}}, pages 6558--6567, 2020.

\bibitem[Yang et~al.(2021)Yang, Lai, and
  Soatto]{yangDyStaBUnsupervisedObject2021}
Yanchao Yang, Brian Lai, and Stefano Soatto.
\newblock {{DyStaB}}: {{Unsupervised Object Segmentation}} via {{Dynamic-Static
  Bootstrapping}}.
\newblock In \emph{Proceedings of the {{IEEE}}/{{CVF Conference}} on {{Computer
  Vision}} and {{Pattern Recognition}}}, pages 2826--2836, 2021.

\bibitem[Zhang et~al.(2021)Zhang, Jia, Edmonds, Zhu, and
  Zhu]{zhangACREAbstractCausal2021}
Chi Zhang, Baoxiong Jia, Mark Edmonds, Song-Chun Zhu, and Yixin Zhu.
\newblock {{ACRE}}: {{Abstract Causal REasoning Beyond Covariation}}.
\newblock In \emph{Proceedings of the {{IEEE}}/{{CVF Conference}} on {{Computer
  Vision}} and {{Pattern Recognition}}}, pages 10643--10653, 2021.

\bibitem[Zimmermann et~al.(2021)Zimmermann, Sharma, Schneider, Bethge, and
  Brendel]{zimmermannContrastiveLearningInverts2021}
Roland~S. Zimmermann, Yash Sharma, Steffen Schneider, Matthias Bethge, and
  Wieland Brendel.
\newblock Contrastive {{Learning Inverts}} the {{Data Generating Process}}.
\newblock In \emph{Proceedings of the 38th {{International Conference}} on
  {{Machine Learning}}}, pages 12979--12990, July 2021.

\end{thebibliography}

\newpage

\appendix

\section{Benchmark Details}
\label{sec:appendix}

\paragraph{Dataset Collection.} Our simulated data is collected from ProcTHOR, the largest simulation environment for embodied agents.
In order to repurpose it to our proposed actionable counterfactual setting, we shortlist the objects that can be independently manipulated, and skip the objects that are physically coupled with others, \textit{e.g.}, tables below cups, bowls, etc.
To enlarge the diversity of the collected examples, we randomly sample the position of the embodied agent in the environment and save no more than $20$ examples from each distinct room.


\begin{figure}[t]
    \centering
    \small
    \includegraphics[width=1.0\textwidth]{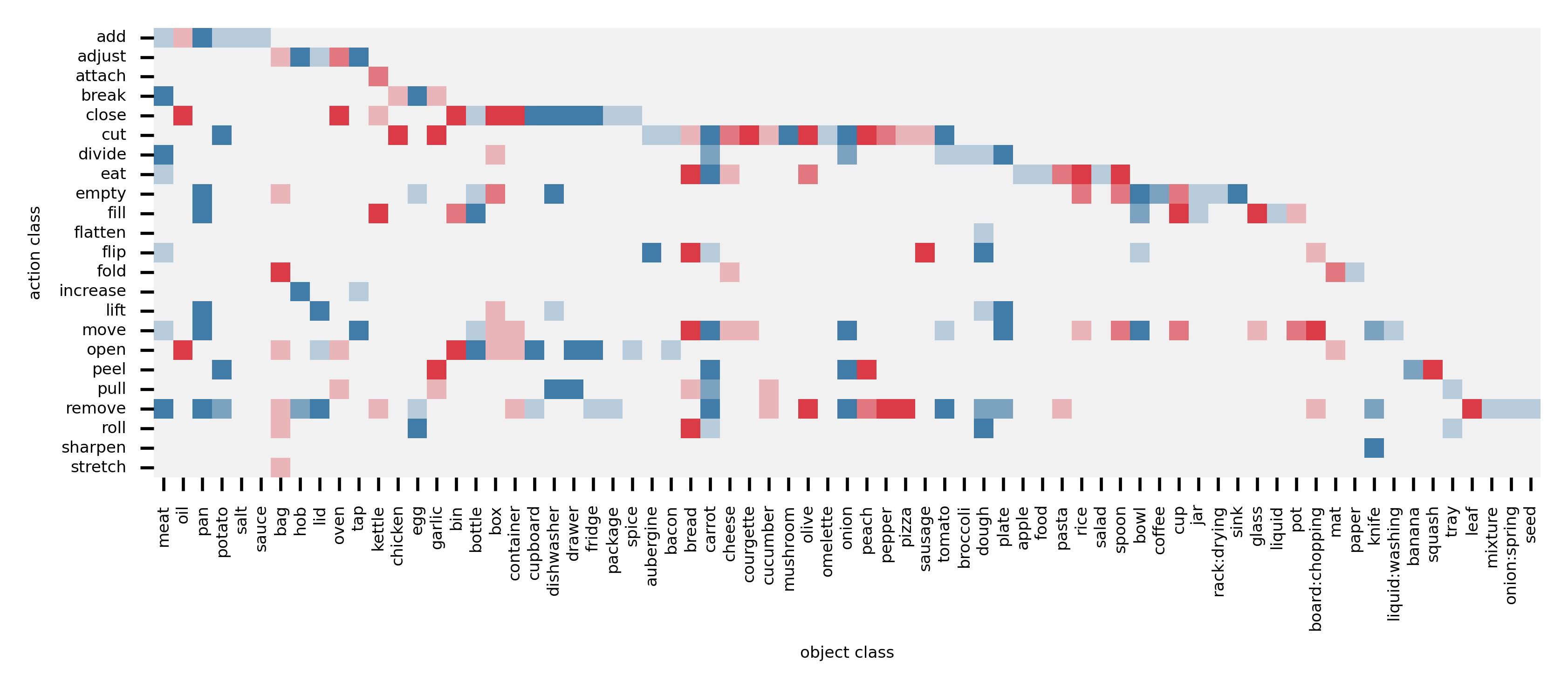}
    \caption{
    Data split in {\tt Causal} {\tt Triplet} collected from real-world observations.
    }
    \label{fig:split_epic}
\end{figure}

Our real-world data is sourced from the Epic-Kitchens-100 dataset, one of the largest ego-centric video datasets known for its strong human-object interactions. In order to ensure the image quality, we removed the examples in which the intervened object is not visually clear.
More specifically, we utilized an off-the-shelf object detector to estimate the image quality, filtering out those with a detection confidence score lower than 0.4. Additionally, we remove the action classes that do not result in noticeable visual variations before and after the action, such as ``feel" and ``spray."
\cref{fig:split_epic} shows the data split used in our experiments in \cref{{subsec:epic}}.

\paragraph{Dataset Properties.} 
\update{
As summarized in \cref{tab:benchmark}, our gathered datasets possess several crucial properties that the preceding datasets do not.
For instance, the variability of local object-level variables, \textit{e.g.}, shape and scale, is highly limited in the Causal3DIdent and CausalWorld datasets. In contrast, the diversity of object shape and scale within each object class is substantial in both our simulated and real-world datasets.
Similarly, scene-level variables like the illumination intensity are often fixed in the previous ones, but vary widely in ours, \textit{e.g.}, we randomized the illumination elements (artificial lights and skyboxes) in ProcTHOR~\citep{deitkeProcTHORLargeScaleEmbodied2022}, rendering simulated scenes at different times of the day.
}

\paragraph{Experiment Details.}
\update{
Our implementations are built upon the public libraries of the baseline models. We applied their default settings for image processing and fine-tuning the representations on our datasets. Specifically, we resized images to 128 x 128 for Slot Attention, and 224 x 224 for the other baselines. For pre-trained foundation models (CLIP and GroupViT), we used their official checkpoints. In our experiments, each random seed leads to a unique data split. Common hyperparameters specific to our benchmark are summarized in \cref{tab:hyper}.
In addition to the quantitative results in \cref{tab:thor}, visualizations of simulated image pairs, along with their corresponding reconstructions and segmentation masks from implicit Slot Attention, are provided in \cref{fig:slot}.
}

\begin{table}[t]
    \small
    \centering
    \begin{tabular}{wl{6cm}|wl{2.2cm}}
        config & value \\
    \specialrule{.1em}{.05em}{.05em}
        batch size with frozen encoder      & 128 \\
        batch size with fine-tuned encoder  & 32 \\
        learning rate                       & 0.001 \\
        training epochs in simulated data            & 50 \\
        training epochs in real-world data           & 200 \\
    \end{tabular}
    \caption{Default hyper-parameters in {\tt Causal} {\tt Triplet} experiments.
    }
    \label{tab:hyper}
\end{table}

\begin{figure}[t]
	\centering
    \includegraphics[height=60pt]{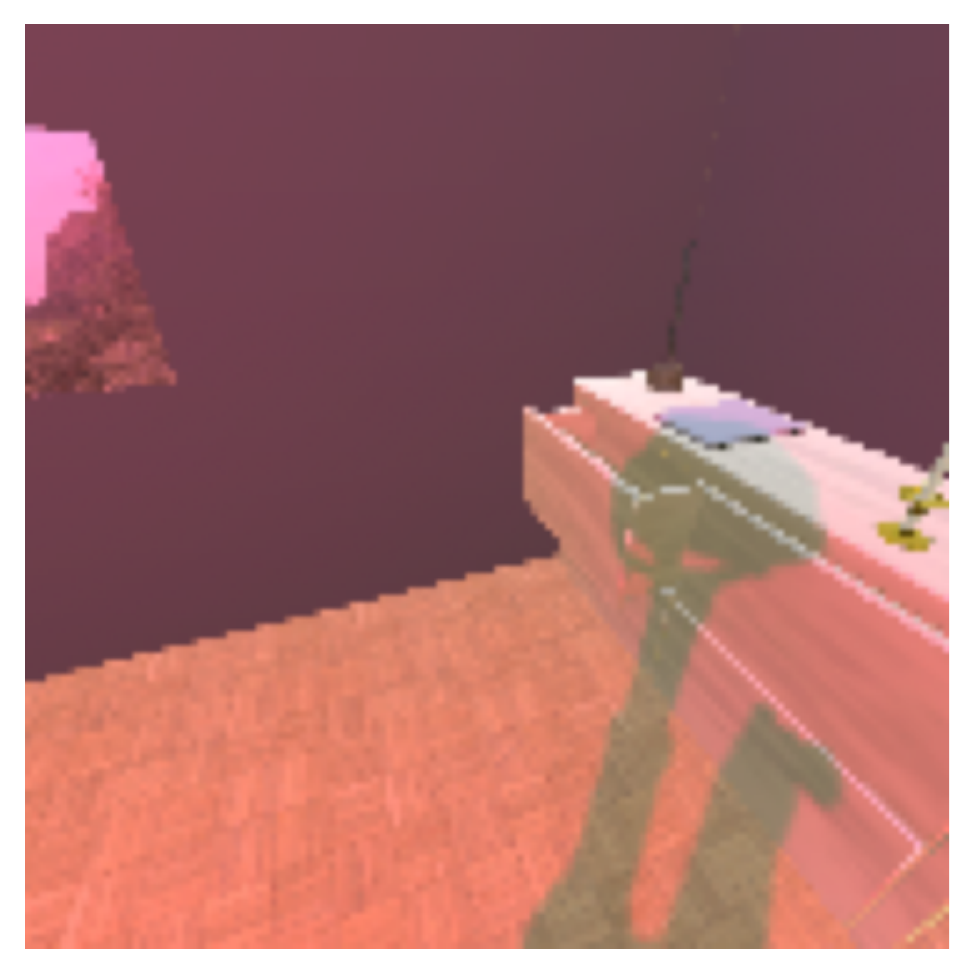}
    \includegraphics[height=60pt]{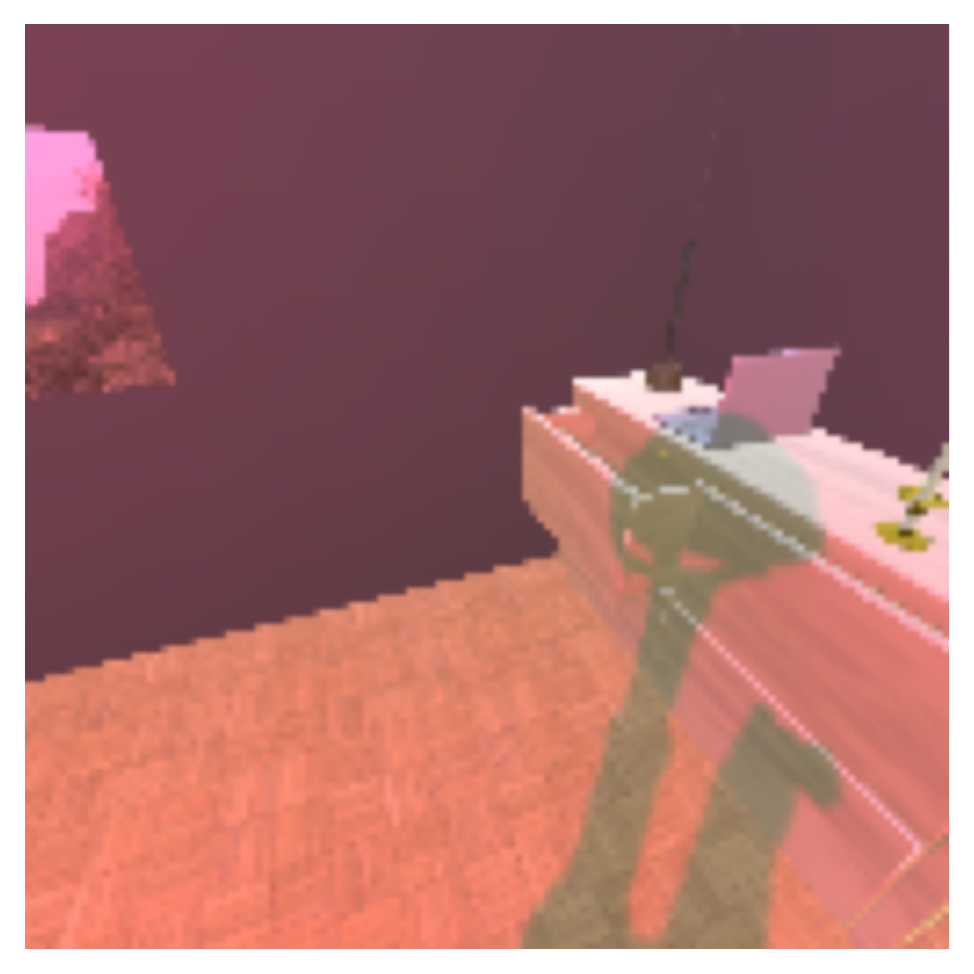}
    \includegraphics[height=60pt]{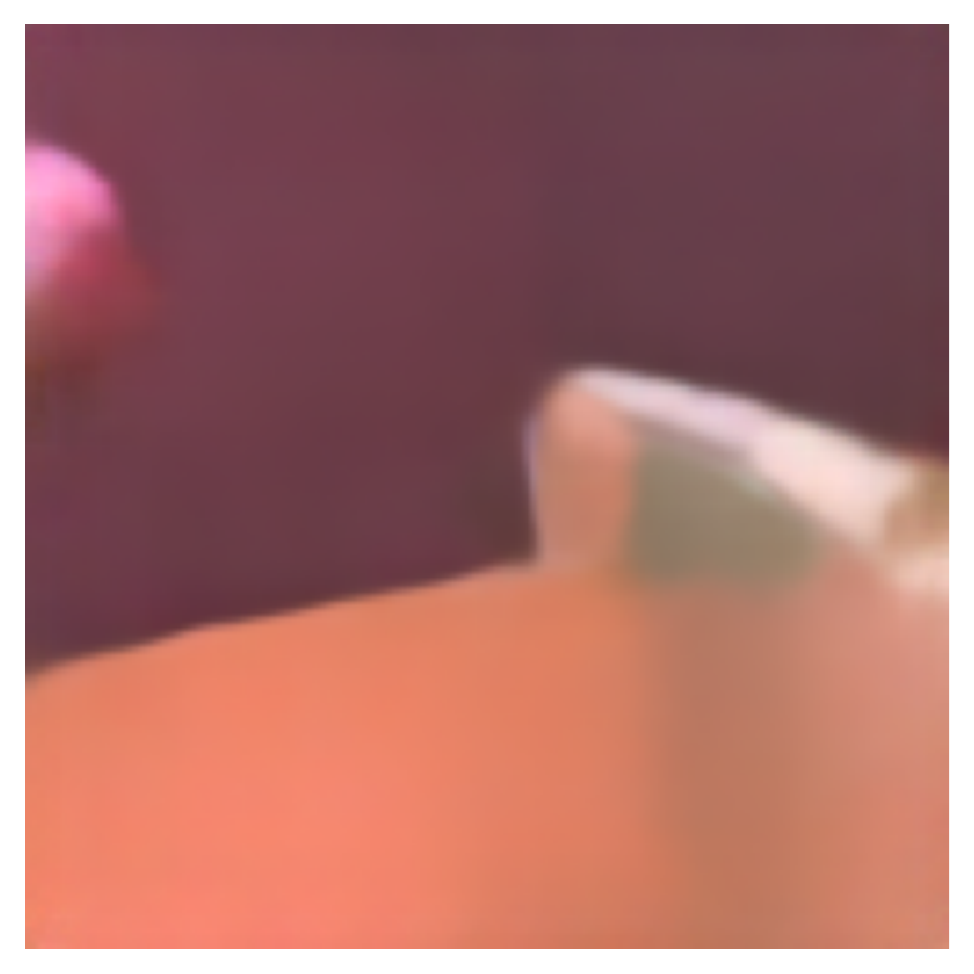}
    \includegraphics[height=60pt]{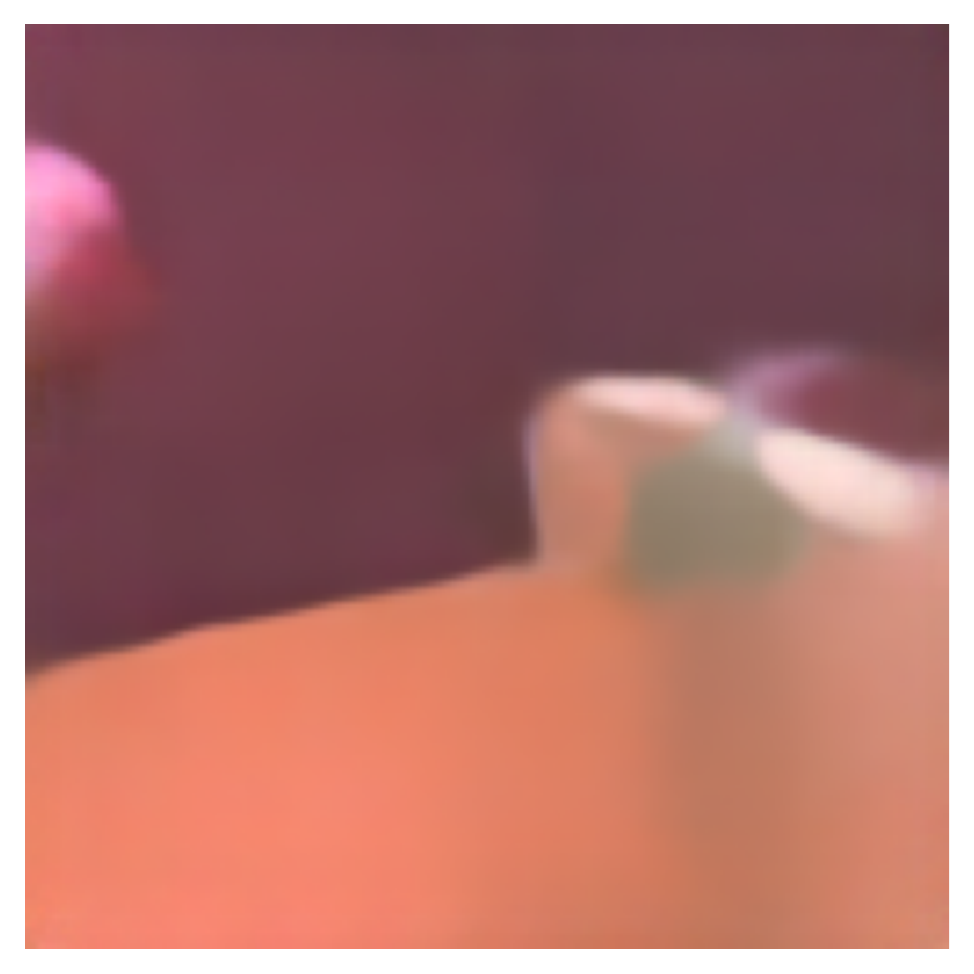}
    \includegraphics[height=60pt]{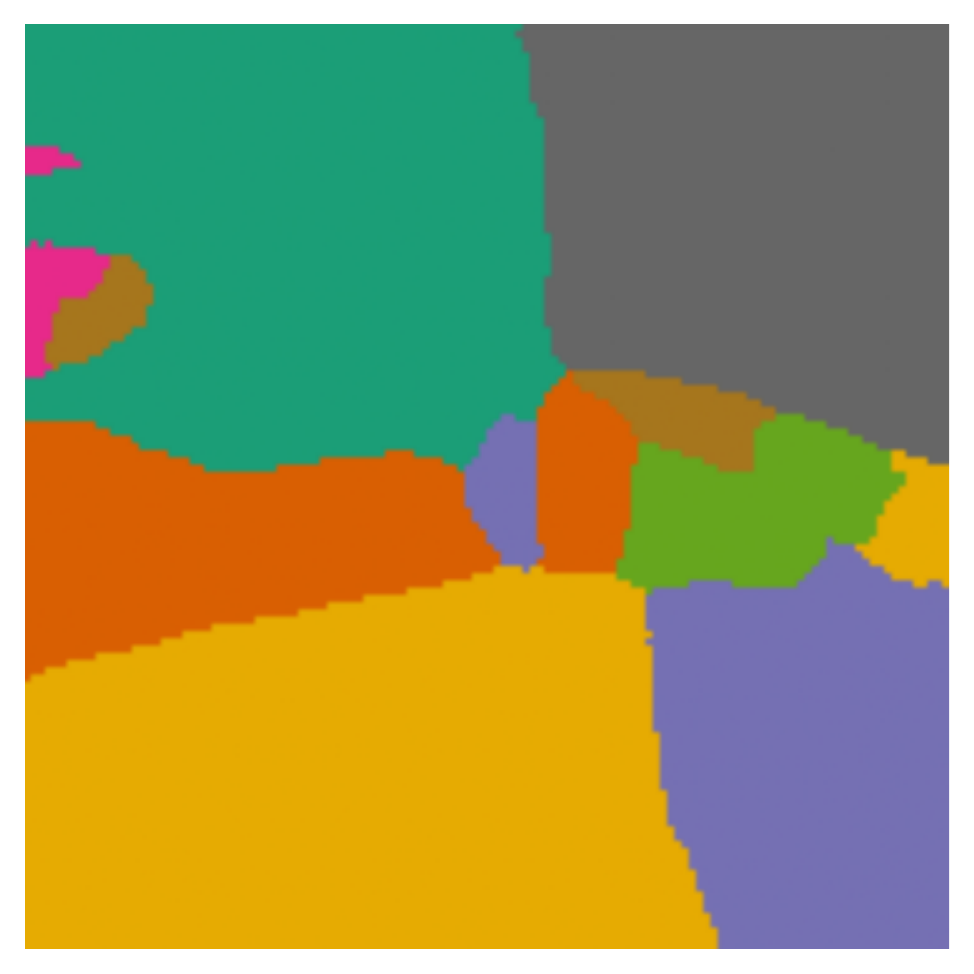}
    \includegraphics[height=60pt]{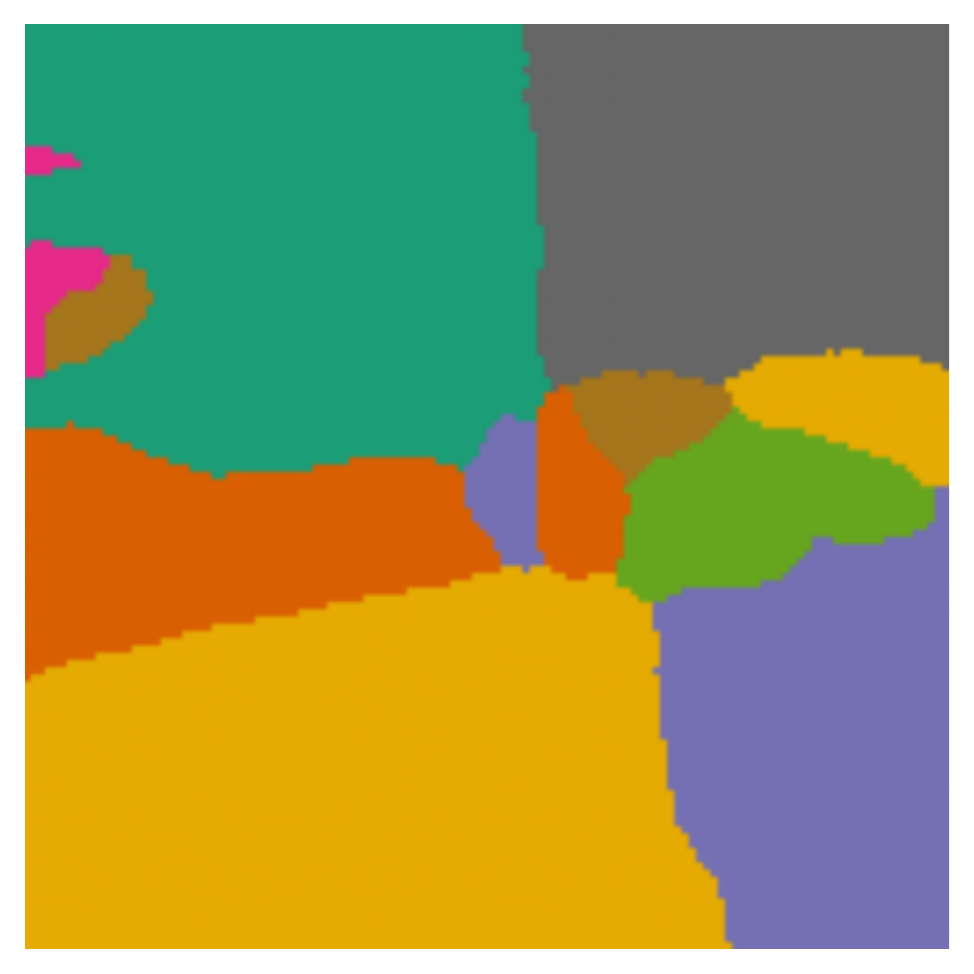}
    \includegraphics[height=60pt]{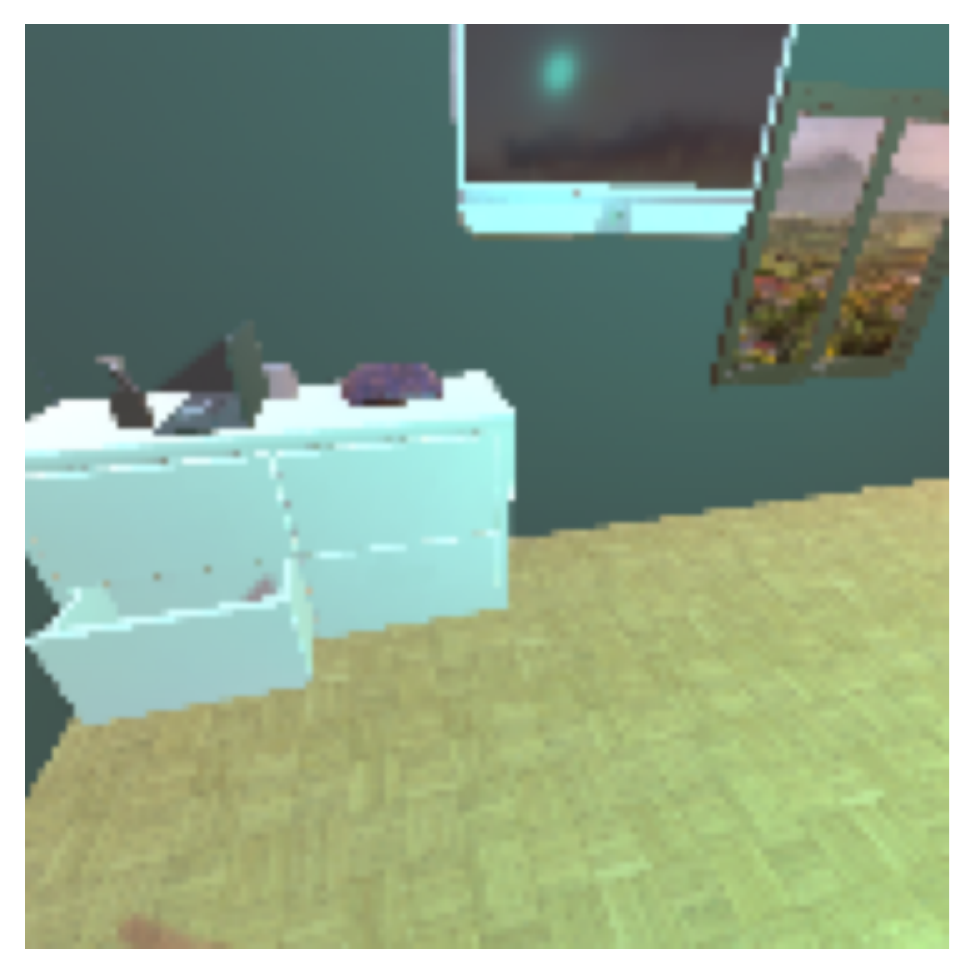}
    \includegraphics[height=60pt]{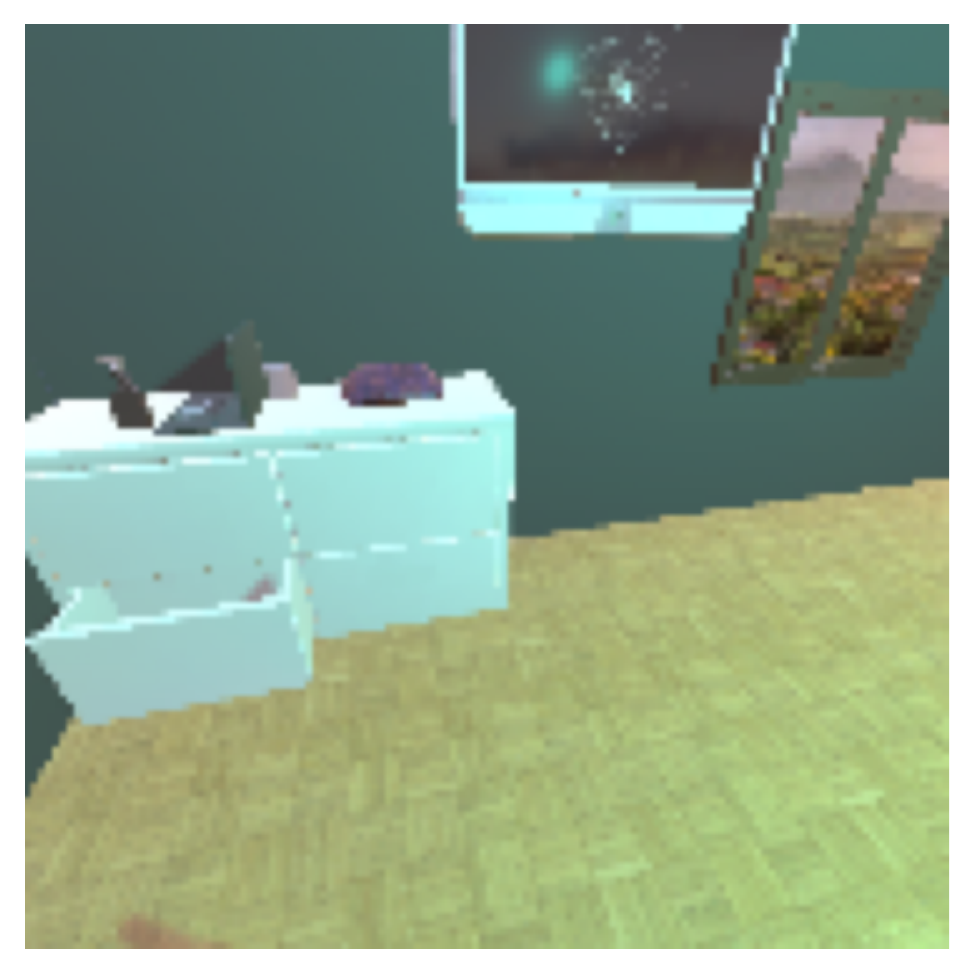}
    \includegraphics[height=60pt]{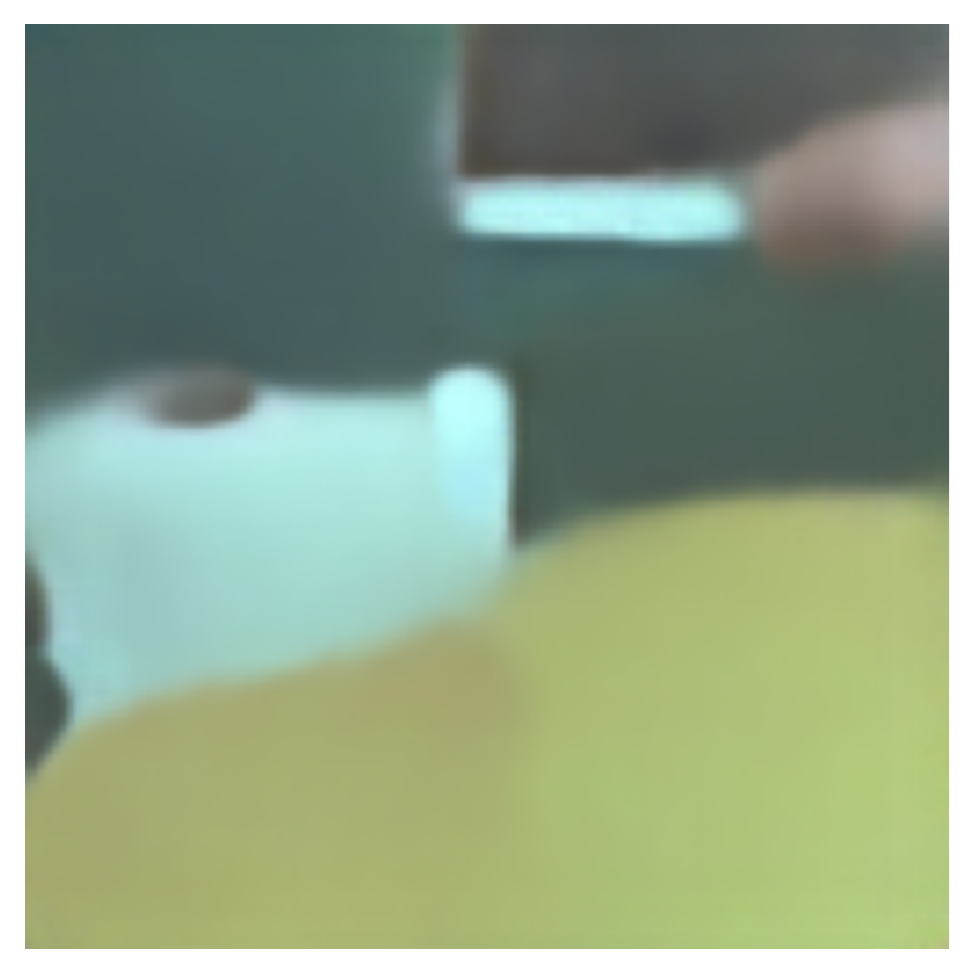}
    \includegraphics[height=60pt]{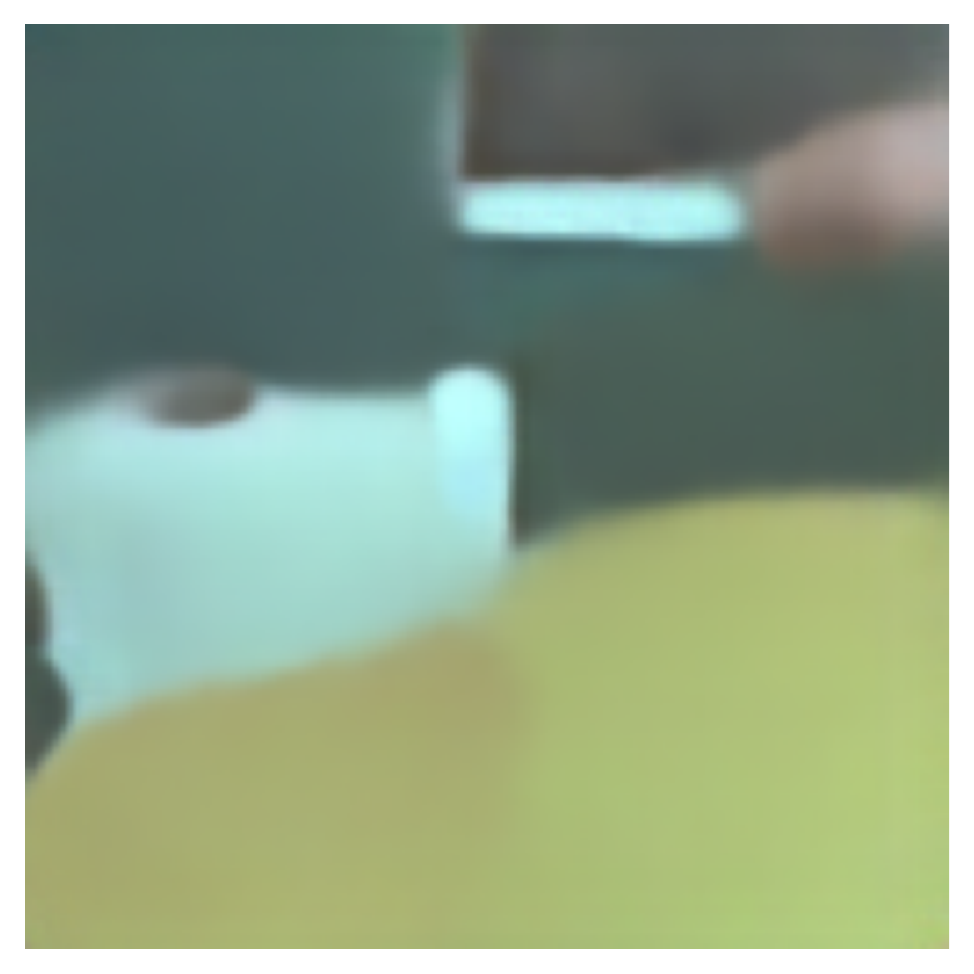}
    \includegraphics[height=60pt]{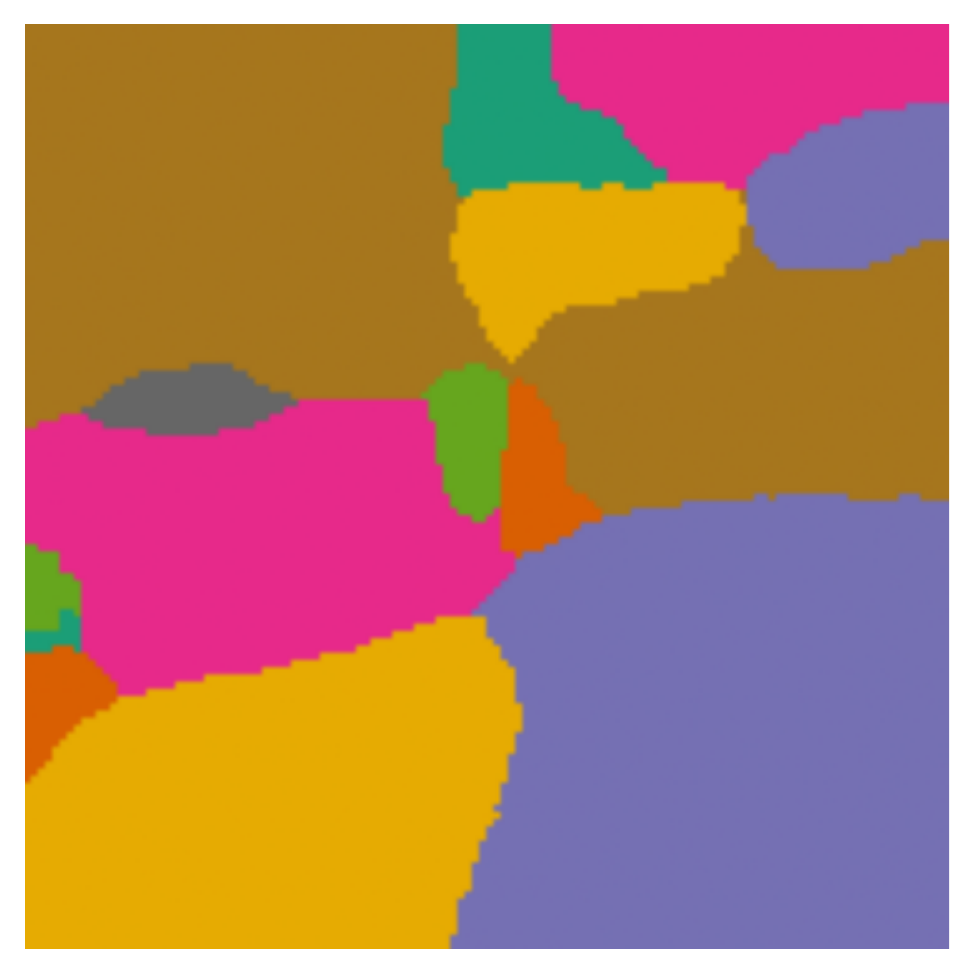}
    \includegraphics[height=60pt]{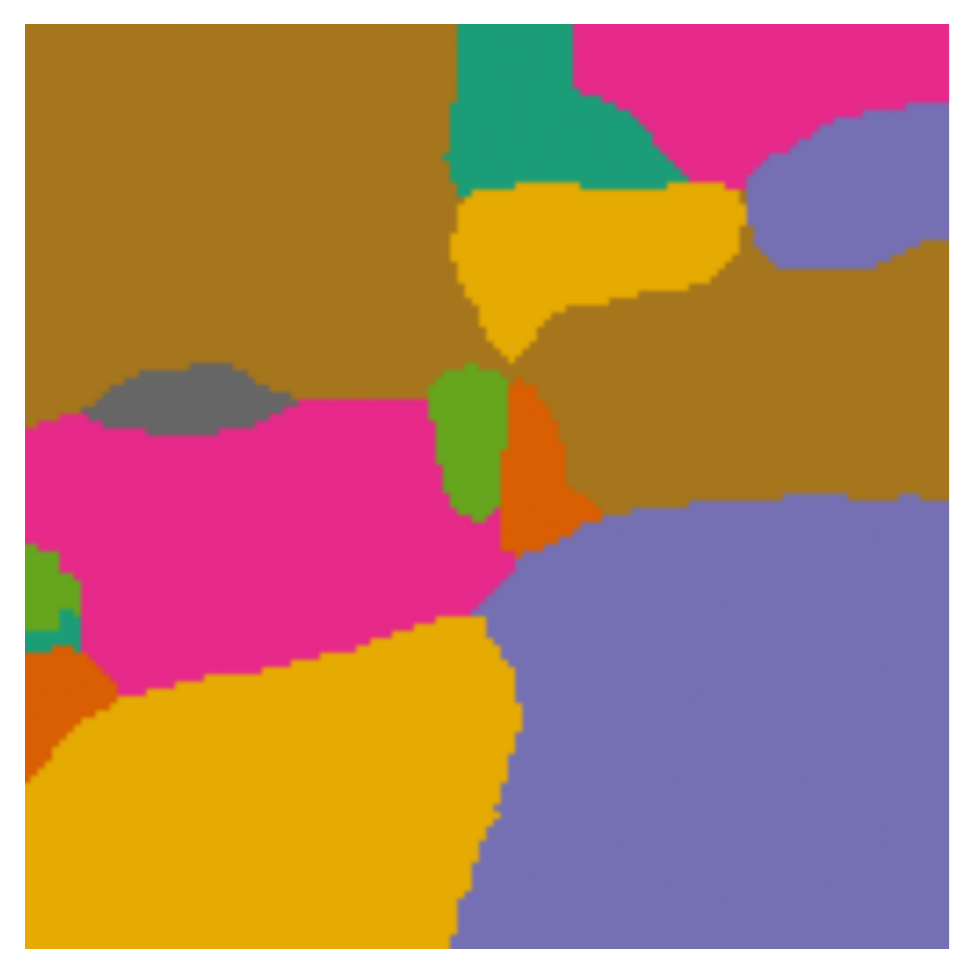}
    \includegraphics[height=60pt]{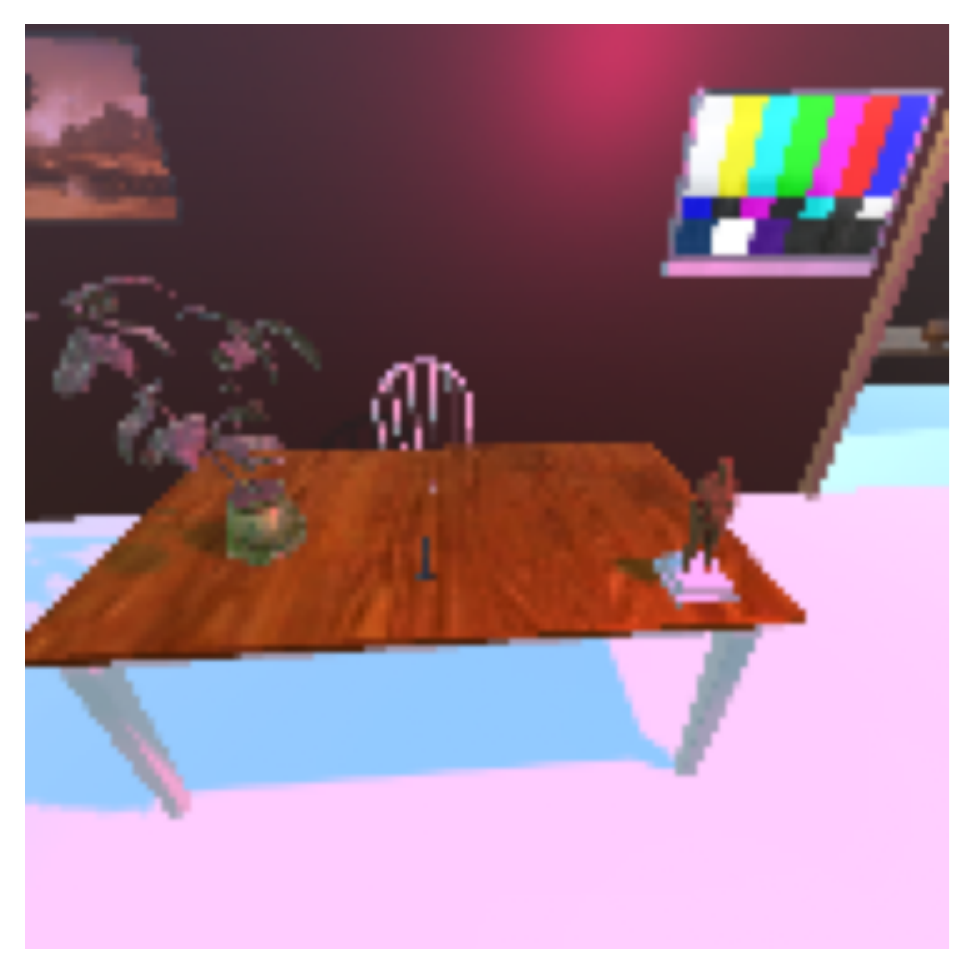}
    \includegraphics[height=60pt]{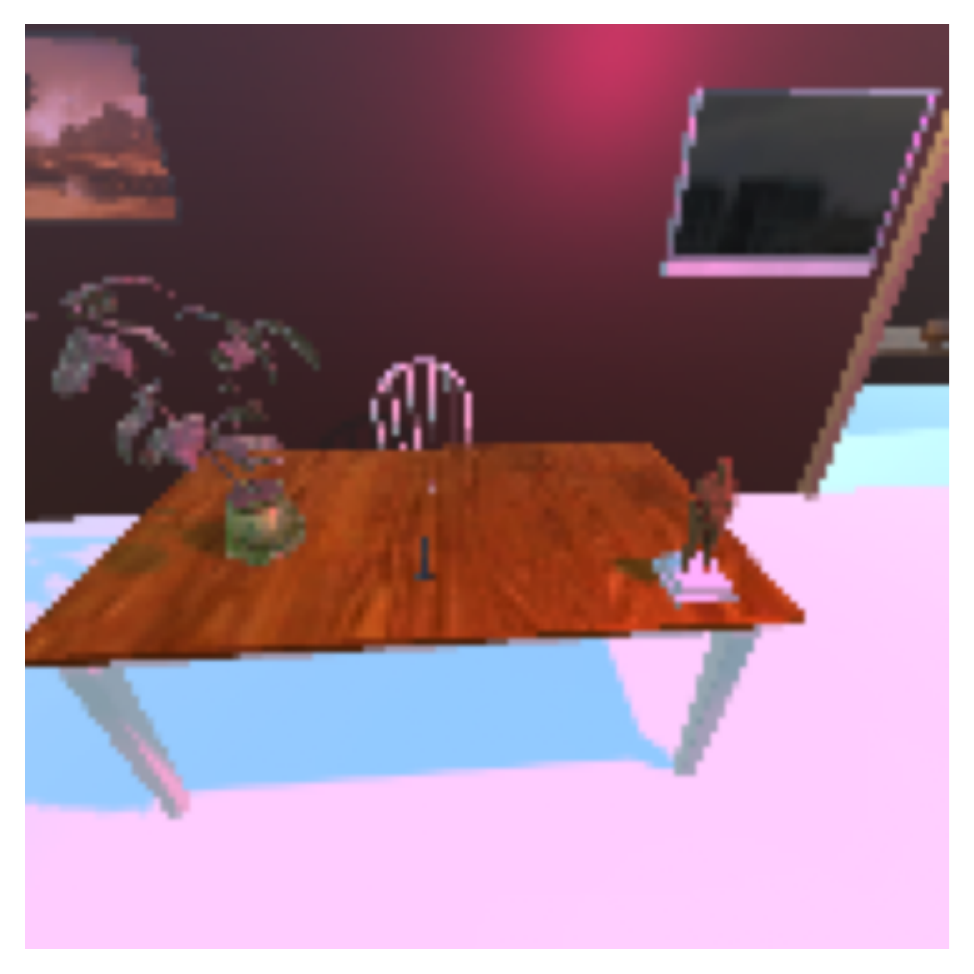}
    \includegraphics[height=60pt]{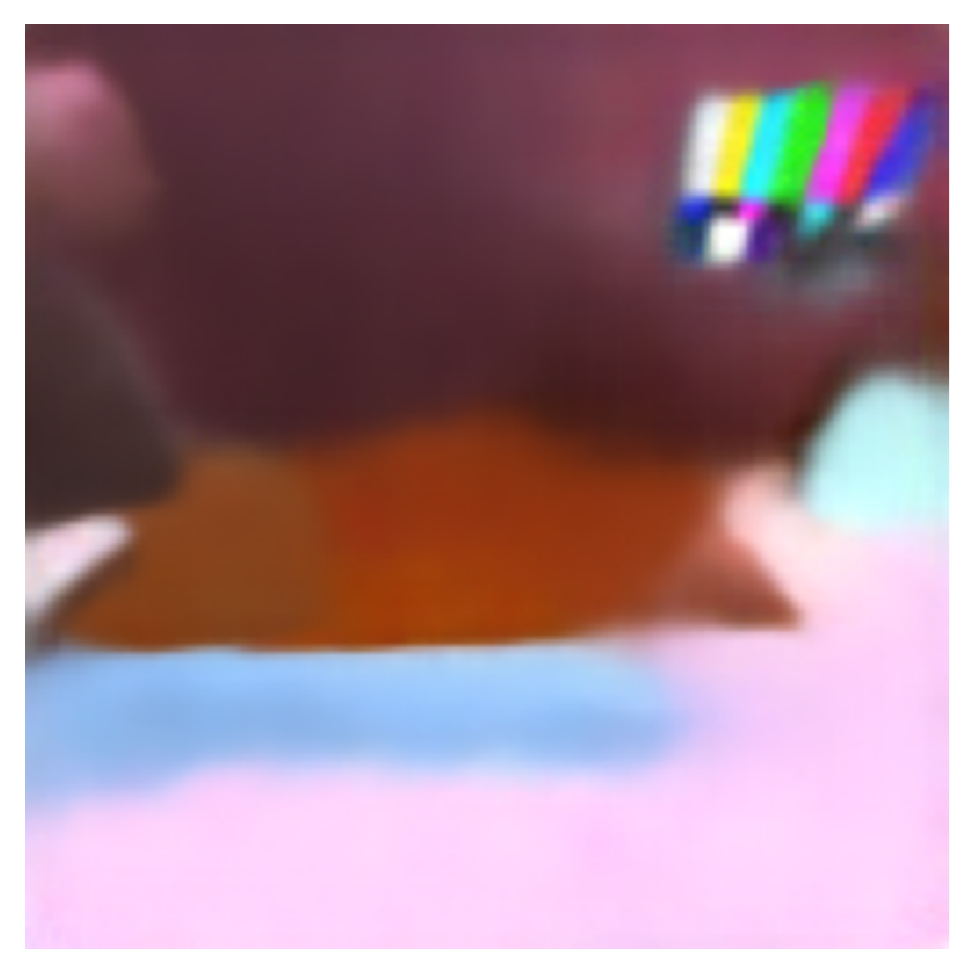}
    \includegraphics[height=60pt]{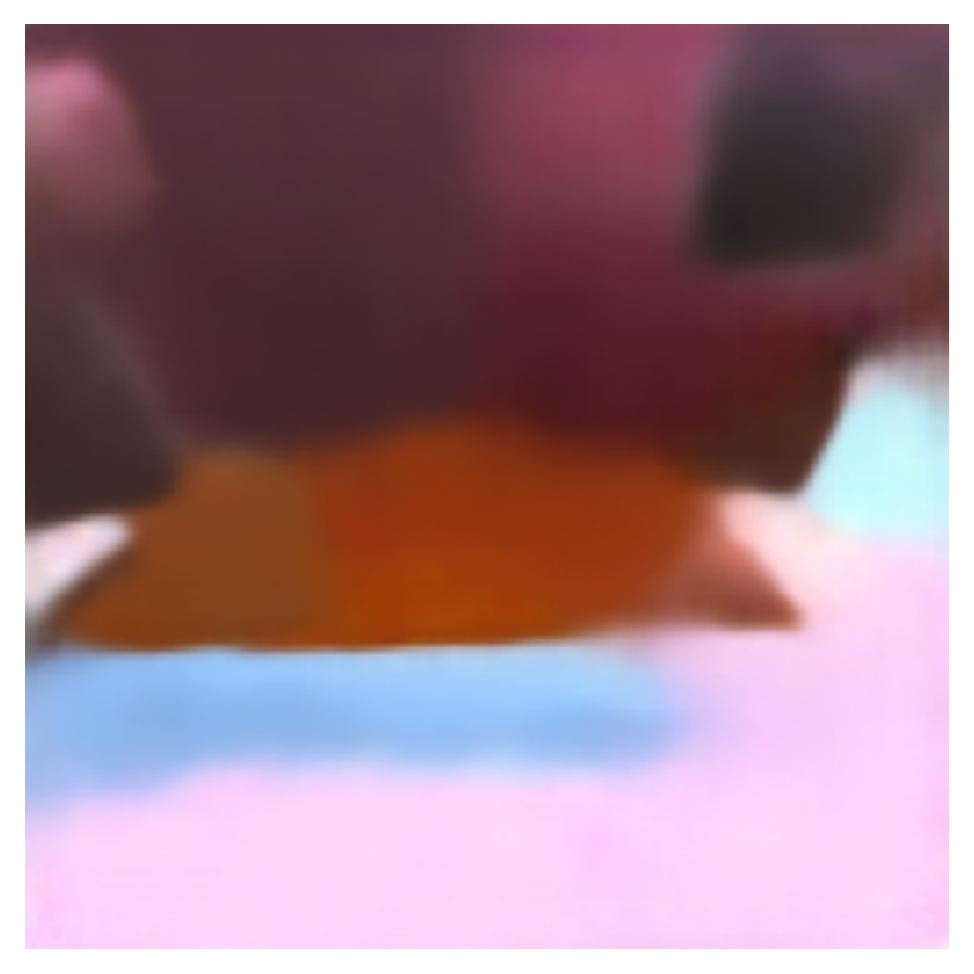}
    \includegraphics[height=60pt]{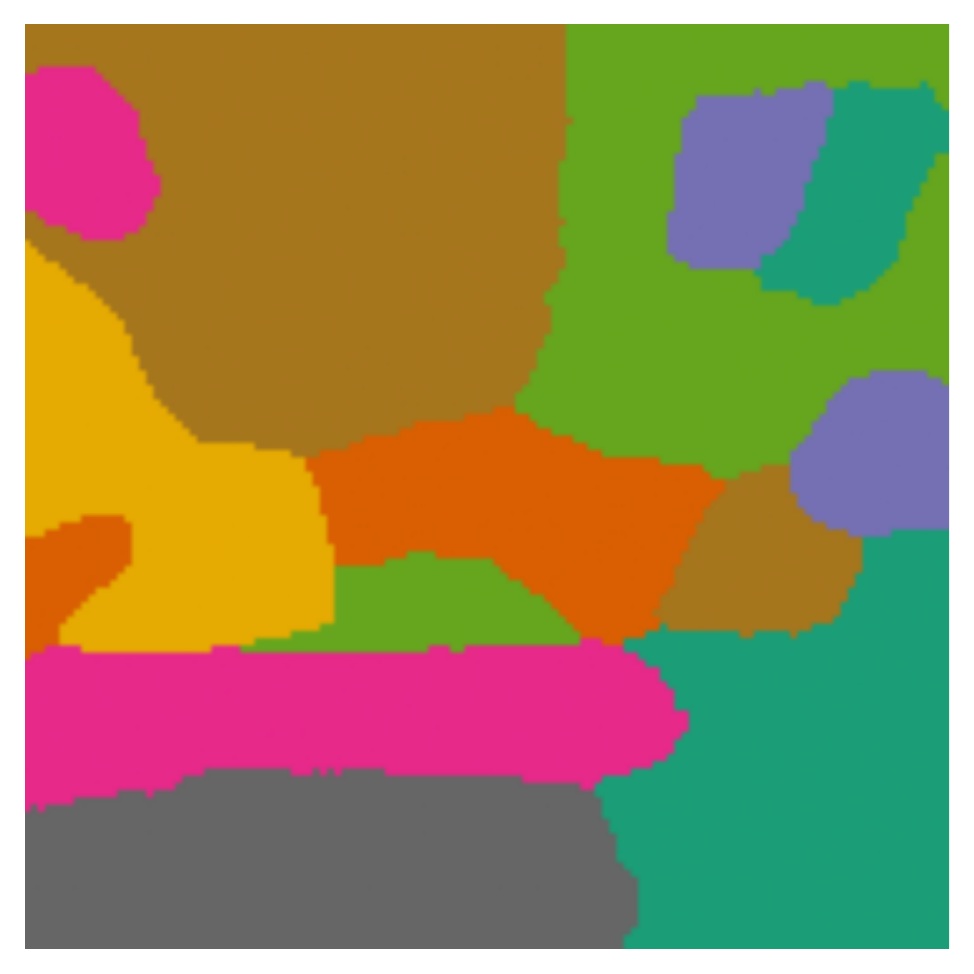}
    \includegraphics[height=60pt]{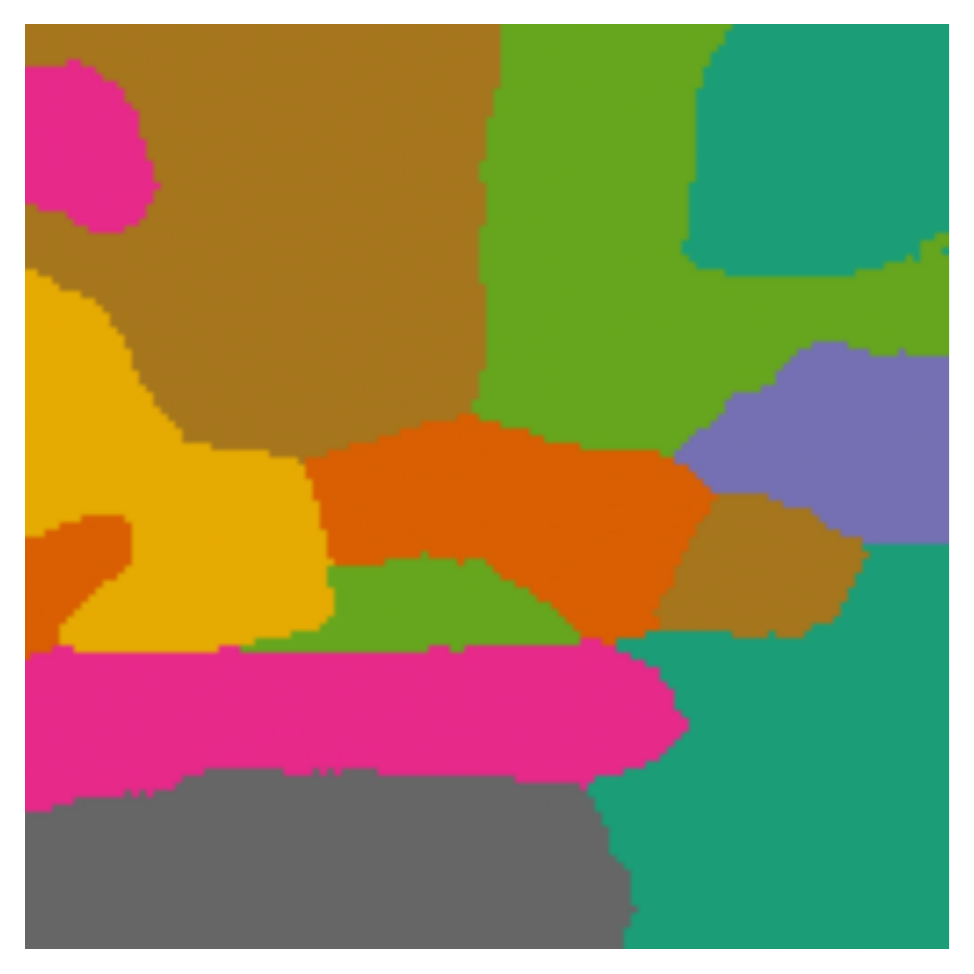}
    \includegraphics[height=60pt]{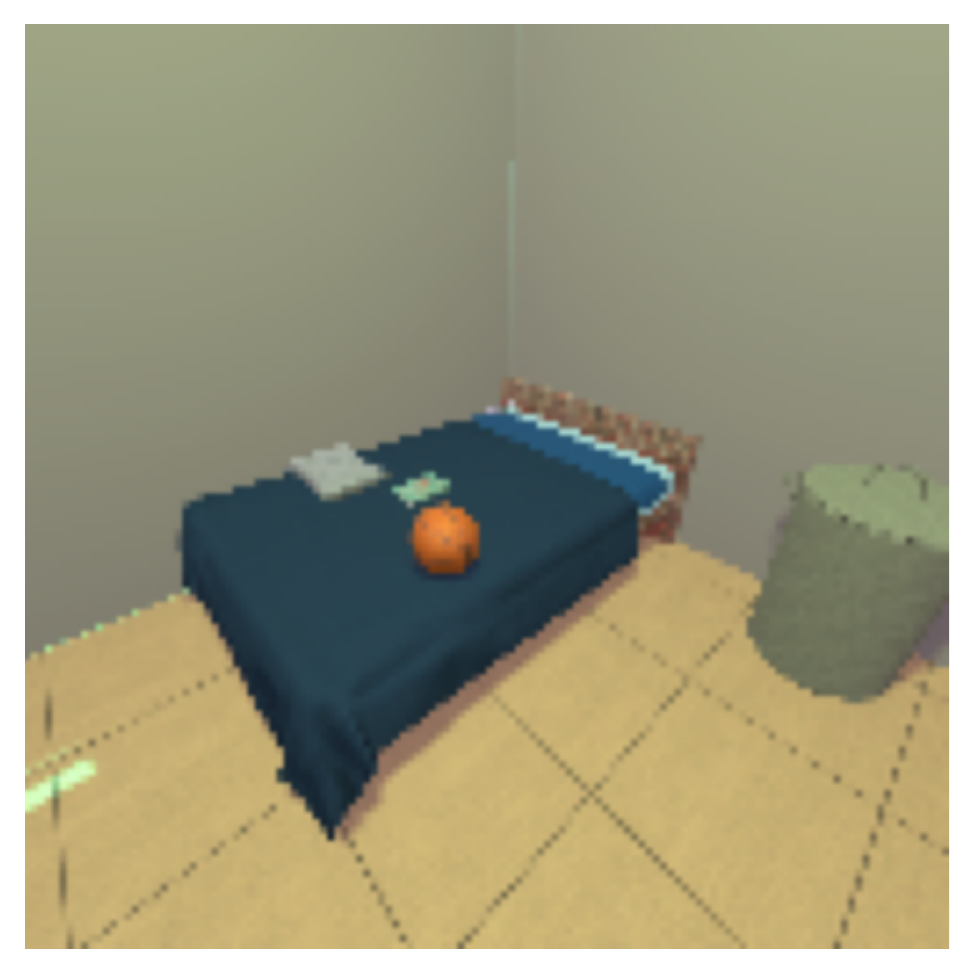}
    \includegraphics[height=60pt]{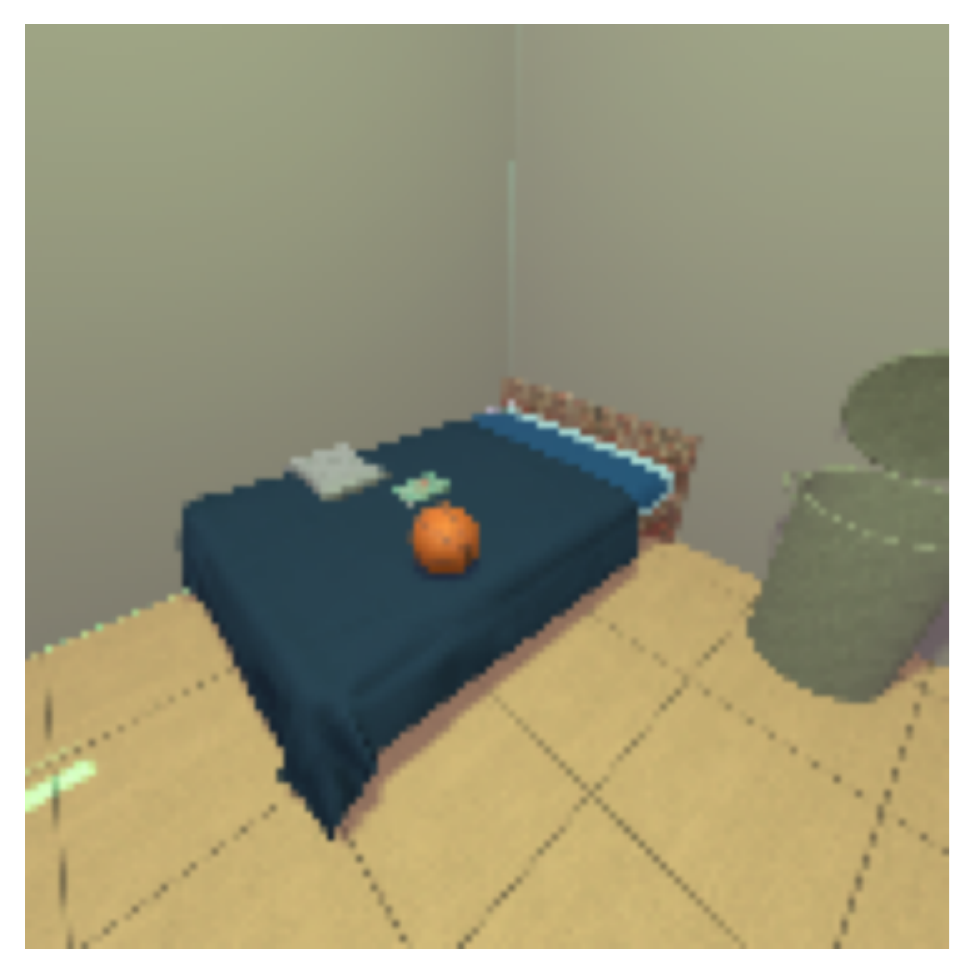}
    \includegraphics[height=60pt]{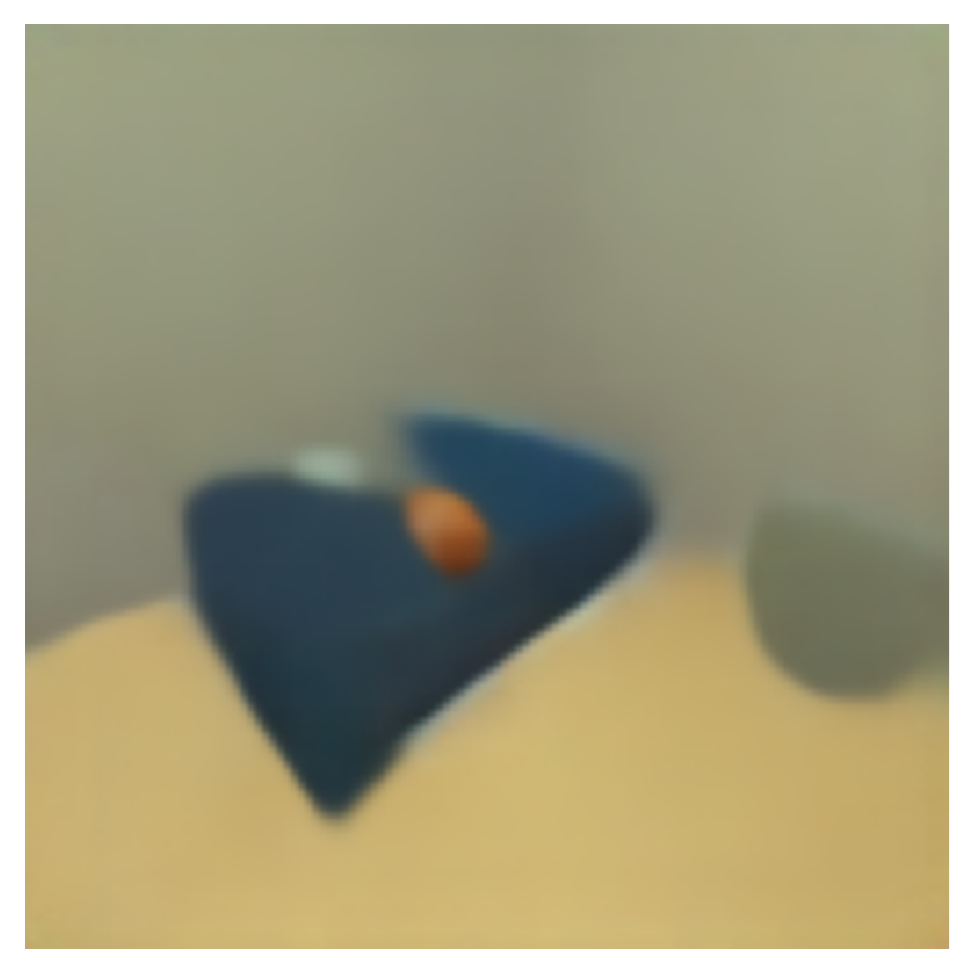}
    \includegraphics[height=60pt]{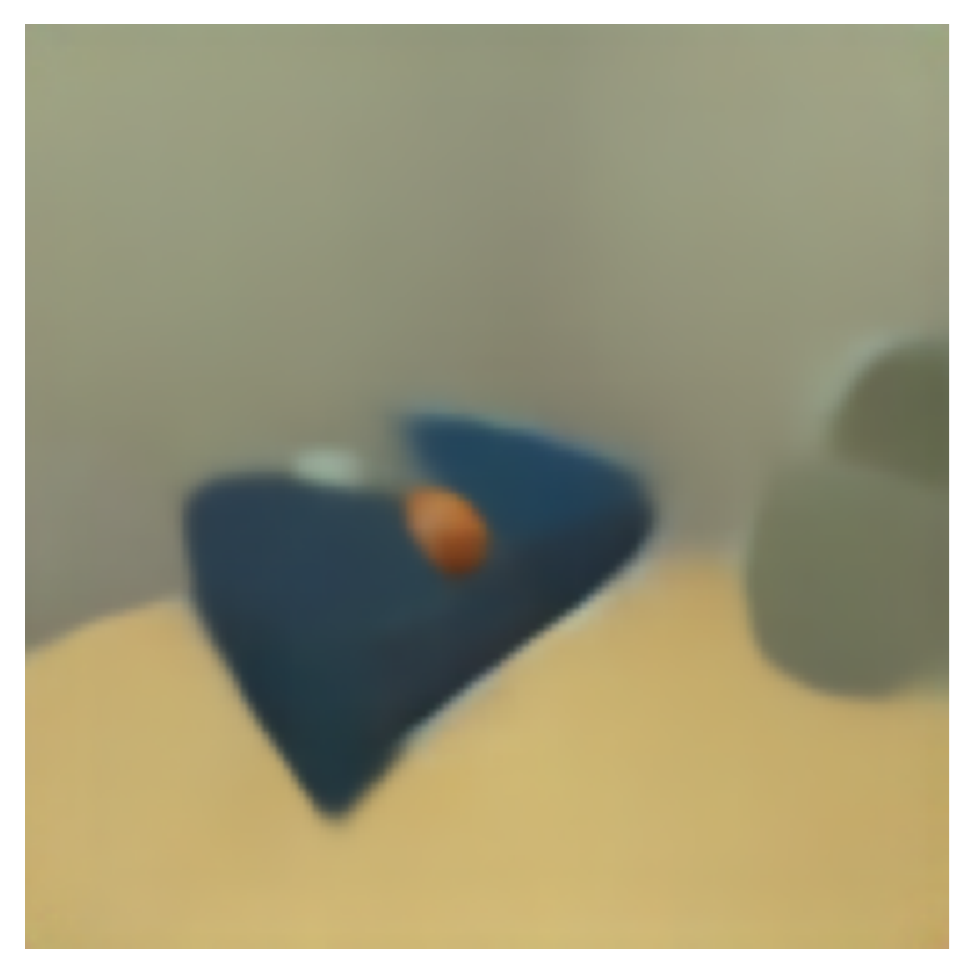}
    \includegraphics[height=60pt]{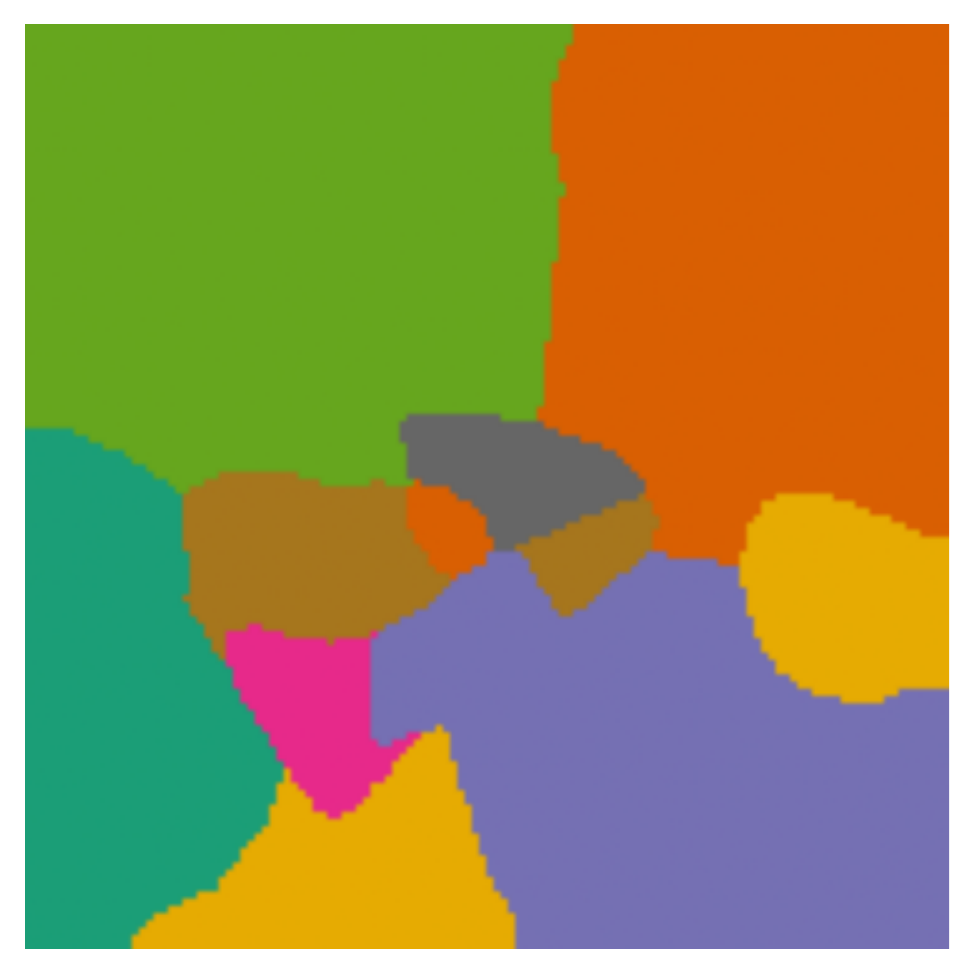}
    \includegraphics[height=60pt]{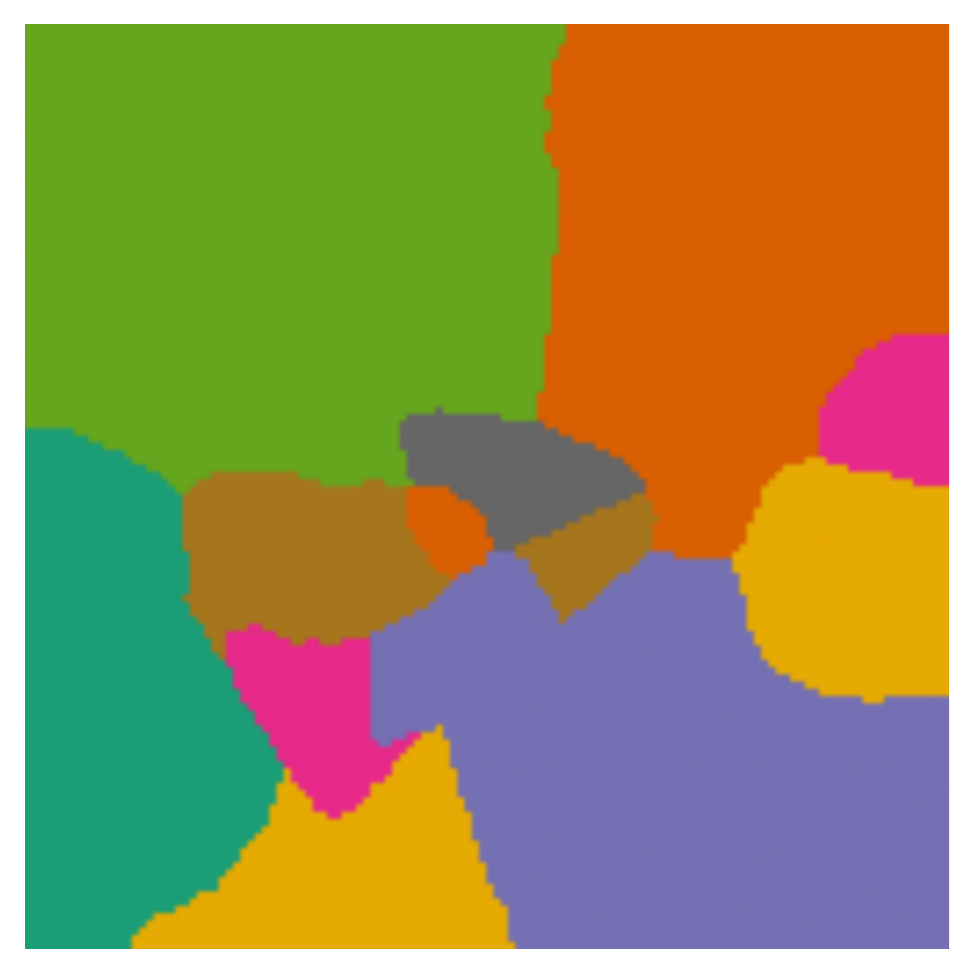}
    \caption{Additional results of the implicit Slot Attention \citep{locatelloObjectCentricLearningSlot2020a,changObjectRepresentationsFixed2022a} on our simulated multi-object scenes. The number of slots is set to $N=15$ based on the maximum number of individual objects. From left to right: pair of input images, pair of reconstructed images, pair of segmentation masks.}
	\label{fig:slot_more}
\end{figure}

\section{Additional Discussions}

\update{
\paragraph{Downstream task.}
The downstream task considered in our benchmark is designed to reason about high-level actions rather than low-level actions for two reasons: (i) from causal standpoint, high-level actions correspond to interventions in the world and induce clear changes on the underlying causal variables, which low-level actions often do not, (ii) from computer vision standpoint, despite the long history of action recognition, the task remains less saturated than object classification. We add to this challenge with an egocentric camera view, which while corresponds to real-world data capture, hides much of the actor and requires direct reasoning about the effects of actions on objects.
\paragraph{Confounder examples.}
Unobserved confounders $\cbb$ between latent variables $\zb$ and action classes $\ab$ can be ubiquitous in practical problems. One common source is the set of constraints under which an agent can interact with the surrounding objects. For instance, whether or not remote control of electronic devices is feasible jointly influences the size/distance of the intervened object (\textit{e.g.}, TV, light) and the action class (\textit{e.g.}, turnon, turnoff).
\paragraph{Additional limitations.}
Aside from the major limitations discussed in \cref{sec:conclusion}, our benchmark is subject to a few other practical shortcomings, including
\begin{itemize}[nosep]
    \item the numbers of object and action classes are limited in our simulated dataset;
    \item ground truth annotations of causal factors are generally not available for complex scenes;
    \item actions of the same semantic class sometimes induce varying effects, \textit{e.g.}, the effect of `open' on a fridge/cupboard may look different from that on a book/laptop.
\end{itemize}
While our benchmark is significantly closer to the real-world problems than previous ones of its kind, these issues need to be addressed in future research.
}

\end{document}